%% file: paper.tex
\pdfoutput=1
\documentclass{article}

\input{preamble}
\input{metadata}

\begin{document}
\raggedbottom
\setlength{\textfloatsep}{7pt plus 2pt minus 2pt}
\setlength{\intextsep}{6pt plus 2pt minus 2pt}
\setlength{\floatsep}{7pt plus 2pt minus 2pt}
\setlength{\abovecaptionskip}{3pt}
\maketitle
\ifarxivversion\lhead{Preprint}%
{\let\thefootnote\relax\footnotetext{$^{*}$First author. \quad $^{\ddag}$Corresponding author.}}\fi

\input{sections/main/00_abstract}
\input{sections/main/01_introduction}
\input{sections/main/02_background_related_work}

\input{sections/main/03_method}
\input{sections/main/04_experiments}
\input{sections/main/06_discussion_conclusion}
\clearpage
\input{sections/appendix/c_ethics_llm_disclosure}

\bibliography{references}
\bibliographystyle{iclr2027_conference}

\clearpage
\appendix
\section*{Appendix}
Appendix~\ref{app:implementation} gives experimental details.
Appendices~\ref{app:wan5b-matched-budget} to~\ref{app:ewmbench} report additional
benchmark results: separating \rev{fine-tuning on the Enlarge dataset} from the registry
interface, Physics-IQ components, WorldModelBench references, and EWMBench.
Appendix~\ref{app:selector-detail} details the selector,
Appendix~\ref{app:dynamics-transfer} the \rev{source}-transfer probe, and
Appendix~\ref{app:mechanistic-diagnostics} the mechanistic diagnostics.
Appendix~\ref{app:extended-experiments} collects qualitative comparisons.

\input{sections/appendix/a_technical_appendix}
\input{sections/appendix/k_wan5b_matched_budget}
\input{sections/appendix/c_physicsiq_components}

\input{sections/appendix/d_worldmodelbench_leaderboard}

\input{sections/appendix/m_ewmbench}
\input{sections/appendix/e_selector_detail}

\input{sections/appendix/l_dynamics_transfer}
\input{sections/appendix/f_mechanistic_diagnostics}
\input{sections/appendix/e_extended_experiments}

\end{document}

%% file: preamble.tex
\usepackage{iclr2027_conference}
\usepackage{iftex}
\ifPDFTeX
    \usepackage{times}
\else
    \usepackage{fontspec}
\fi
\usepackage{amsmath,amsfonts,amssymb}
\usepackage{array}
\usepackage{booktabs}
\usepackage{capt-of}
\usepackage{float}
\usepackage{graphicx}
\usepackage{longtable}
\usepackage{microtype}
\usepackage{tabularx}
\usepackage{xcolor}
\usepackage{url}
\usepackage{hyperref}
\usepackage[capitalise]{cleveref}
\crefname{section}{Sec.}{Secs.}
\Crefname{section}{Section}{Sections}
\usepackage{tikz}
\usepackage{comment}
\usetikzlibrary{arrows.meta,positioning,calc,patterns,shadows}

\newif\ifshowrevisions
\showrevisionsfalse
\newcommand{\rev}[1]{\ifshowrevisions\ifvmode\leavevmode\fi\begingroup\color{red}#1\endgroup\else#1\fi}
\newcommand{\postsub}[1]{#1}%

\newcolumntype{Y}{>{\raggedright\arraybackslash}X}
\newcolumntype{P}[1]{>{\raggedright\arraybackslash}p{#1}}

\newcommand{\method}{Physical Registry}
\newcommand{\methodfull}{Physical Registry}
\newcommand{\scorestd}[2]{\ensuremath{#1\pm#2}}
\newcommand{\prsscore}[2]{#1\,\textcolor{green!45!black}{(#2)}}
\newcommand{\scorechange}[2]{\,{\fontsize{6.3}{7.2}\selectfont\textcolor{#1}{(#2)}}}
\newcommand{\DRSUnavailable}{%
  \begingroup
  \setlength{\fboxsep}{0pt}%
  \colorbox{black!12}{\strut\phantom{\textbf{00.00}}}%
  \endgroup}
\newcommand{\TaskUnavailable}{\DRSUnavailable}

\AtBeginDocument{%
  \setlength{\abovedisplayskip}{3pt plus 1pt minus 2pt}%
  \setlength{\belowdisplayskip}{3pt plus 1pt minus 2pt}%
  \setlength{\abovedisplayshortskip}{1pt plus 1pt minus 1pt}%
  \setlength{\belowdisplayshortskip}{2pt plus 1pt minus 1pt}%
}


\definecolor{tabours}{RGB}{234,242,251}
\newcommand{\prrow}[1][\linewidth]{\makebox[0pt][l]{\textcolor{tabours}{\rule[-\arraystretch\dp\strutbox]{#1}{\dimexpr\arraystretch\ht\strutbox+\arraystretch\dp\strutbox\relax}}}}
\newsavebox{\prtabbox}

%% file: metadata.tex
\title{World Models Need a Physical Registry:\\
Learning Internal Predictive States\\
for Physically Plausible World Models}
\newif\ifarxivversion
\arxivversiontrue
\ifarxivversion
\author{\hspace*{-\tabcolsep}\begin{tabular}[t]{@{}l@{\hspace{1.1em}}l@{\hspace{1.1em}}l@{\hspace{1.1em}}l@{}}
Shinseong Kim\textsuperscript{1,2,*} & Jaeseok Jeong\textsuperscript{1} & Jianlong Fu\textsuperscript{2} & Youngjung Uh\textsuperscript{1,\ddag} \\
{\normalfont\footnotesize tltydl2@yonsei.ac.kr} & {\normalfont\footnotesize jete{\ttfamily\char95}jeong@yonsei.ac.kr} & {\normalfont\footnotesize jianf@microsoft.com} & {\normalfont\footnotesize yj.uh@yonsei.ac.kr} \\[4pt]
\multicolumn{4}{@{}l}{\normalfont\textsuperscript{1}Yonsei University \quad \textsuperscript{2}Microsoft Research Asia}
\end{tabular}}
\iclrfinalcopy
\else
\author{Anonymous authors}
\fi

%% file: sections/main/00_abstract.tex
\begin{abstract}
Video world models must generate physically plausible futures, not just
realistic frames. Existing approaches typically use semantic, physical, or
predictive information to guide the visual generation stream, leaving a
fundamental question:
\textbf{where inside the generator should predictions of scene evolution live?}
Standard video tokens jointly encode appearance and dynamics, but provide no
distinct state for learning and reading predictive dynamics.
\rev{We answer this question with the \textbf{\method{}}:}
temporally grounded registry slots inside a video DiT that are initialized
from an observation-conditioned forecast, co-evolve with video tokens, and
receive future-representation supervision\rev{, making} predictive dynamics an explicit,
readable state of the generator.
In a controlled \rev{comparison of
predictive-state interfaces} on Physics-IQ Verified, \method{} improves vanilla
fine-tuning by \(9.9/8.7\) points on I2V/V2V, the largest gain among the
interfaces we compare. \rev{Across all Cosmos and WAN backbones, \method{} improves Physics-IQ
Verified scores. With only \(0.45\%\) additional parameters, it enables Cosmos 3
Nano 16B to surpass the published I2V score of Cosmos 3 Super 64B (\(43.56\)
vs.\ \(42.7\)). On WorldModelBench, it also improves physical adherence and the
total score for all four evaluated backbones.}
Because the registry becomes candidate-specific while remaining readable,
\rev{the \emph{Physical Registry Selector (PRS)} ranks emerging futures by how
far their intermediate registry readouts depart from the shared forecast
before full denoising is complete. Across four backbones, this selection improves
I2V scores by \(5.6\)--\(15.9\%\) without completed-video external scoring.}
Together, these results support a broader principle:
\textbf{dynamics prediction should live as an explicit model state that
co-evolves with the video representation.}
Project page: \url{https://grainsack.github.io/physical-registry/}.
\end{abstract}

%% file: sections/main/01_introduction.tex
\section{Introduction}
\label{sec:introduction}

Video generators are increasingly used as world models, i.e., environments for training agents
\citep{ha2018worldmodels,bruce2024genie,nvidia2025cosmospredict25}.
From this perspective, a generated video should contain not only high-quality visuals, but also physically plausible phenomena such as solid mechanics, fluid dynamics, optics, thermodynamics, and magnetism~\citep{motamed2026physicsiq}.
However, recent video models still often fail to follow them~\citep{li2025worldmodelbench,motamed2026physicsiq}.

To address this, exploiting physical priors has made some progress by supplying explicit cues about scene evolution, such as motion, force, or physical control signals ~\citep{yang2025vlipp,gillman2025forceprompting,wang2025physctrl}. However, it depends on availability of the controls and corresponding supervision.
On the other hand, supervision for aligning the intermediate features in the video models to pre-trained representations is readily available~\citep{yu2025repa,zhang2025videorepa,hwang2025crepa,bhowmik2026moalign}. However, their appearance and dynamics can remain entangled~\citep{bhowmik2026moalign}.
This context motivates us to ask a more basic question: \textbf{How and where can we internalize scene evolution using representation alignment?}

We introduce the \emph{\method{}}, dedicated registry slots containing scene evolution inside video DiTs. They are encouraged to be aligned with representations of future frames to be generated, which are predicted from a source frame. \postsub{\rev{We use JEPA representations for the \method{} because JEPA's predictive representation learning aligns with our goal of modeling scene evolution: a frozen JEPA encoder--predictor~\citep{bardes2025vjepa,assran2025vjepa2} produces forecast features for future temporal positions in the same space as future-video targets, compactly representing how the scene is expected to evolve. Prior works have also shown that JEPA representations capture physical properties~\citep{garrido2025intuitivephysics,joseph2026interpreting}.}} In detail, the \method{} of the first DiT layer starts from JEPA features of the source and future frames via the JEPA encoder and encoder--predictor, respectively. Then, a readout layer aligns the registries for future frames to the original representations of the future frames from the JEPA encoder. As a whole, the registries and video tokens are designed to co-evolve through the transformer blocks such that the registries can estimate the original representations from the predicted representations.

We \postsub{show} that \postsub{a dedicated registry combined with predictive supervision} yields the largest gain \postsub{among the compared interfaces. Notably,} this gain generalizes to Cosmos and WAN backbones. Furthermore, the readouts can be used to choose the best sample among denoising processes from different initial noises, namely \emph{Physical Registry Selector (PRS)}. It works in the middle of denoising processes instead of completely generated videos~\citep{yuan2026wmreward,interno2026geophys}.

Our contributions are threefold:
\begin{itemize}
    \item The \method{} conveys future dynamics from the initial frames in a video generator.
    \item The \method{} provides an internal selection signal across different initial noises before fully generating videos.
    \item The \method{} outperforms other interfaces for predictive representations.
\end{itemize}

%% file: sections/main/02_background_related_work.tex
\section{Related Work}
\label{sec:related-work}

\paragraph{\rev{Physical plausibility in video generation.}}
Prior work improves physical plausibility or controllability by conditioning
video generation on motion, force, trajectory, or language-derived physical
signals
\citep{li2025magicmotion,gillman2025forceprompting,wang2025physctrl,
yang2025vlipp,pathak2026physvid}.
\rev{These approaches express the intended evolution through an external
control signal, whereas} \method{} represents predicted future dynamics as an
internal state of the video generator.

\paragraph{\rev{Predictive representations inside video generators.}}
I-JEPA learns predictive representations from visible context, with V-JEPA and
V-JEPA~2 extending this principle to video
\citep{assran2023ijepa,bardes2025vjepa,assran2025vjepa2}.
JEPA representations capture dynamics-relevant structure, including sensitivity
to intuitive-physics violations and information about motion and physical
plausibility
\citep{garrido2025intuitivephysics,joseph2026interpreting}.
\rev{Generative models typically use such pretrained representations in one of
two ways.} REPA and its video extensions \rev{align them} with intermediate
generative features
\citep{yu2025repa,zhang2025videorepa,hwang2025crepa,
bhowmik2026moalign,lian2026sara},
while ReDi \citep{kouzelis2025redi} and PHANTOM \citep{shen2026phantom}
jointly generate visual and representation states.
\rev{Register tokens offer a different primitive:} Vision Transformer registers
provide additional tokens for internal computation
\citep{darcet2024registers}, and HyperDiT \citep{he2026hyperdit} uses
representation-aligned semantic registers for image generation.
\rev{\method{} brings these lines together by holding forecast-initialized
predictive representations in dedicated, temporally grounded registry slots
within the video backbone and supervising them toward future representations.}

\paragraph{Inference-time guidance and candidate selection.}
WMReward uses V-JEPA~2 as a physics reward for inference-time guidance and
best-of-\(N\) selection \citep{yuan2026wmreward}, while GeoPhys uses an
external physical-plausibility verifier to rerank generated candidates
\citep{interno2026geophys}. PRS instead ranks candidates from the generator's own intermediate registry states, without an external reward model or verifier.

%% file: sections/main/03_method.tex
\section{Method}
\label{sec:method}
\input{figures/main/predictive_registry}

We augment flow-matching video DiTs with \method{} \rev{(\cref{sec:flow-matching})}
with forecast-initialized registry states \rev{(\cref{sec:registry-construction})}
that evolve with the video representations \rev{\postsub{(\cref{sec:registry-construction})}}, and align them
\rev{toward the target future (\cref{sec:registry-alignment}).} Then we introduce our training objective (\cref{sec:training-objective}) and our inference technique (\cref{sec:selection-logic}).

\subsection{Conditional flow matching}
\label{sec:flow-matching}

Let \(X_0\) denote a clean video latent,
\(\epsilon\sim\mathcal N(0,I)\), and \(t\sim\mathcal U(0,1)\).
With condition \(c\), rectified flow \citep{liu2023rectifiedflow} uses
linear interpolation path, velocity target, and objective:
\begin{align}
X_t&=(1-t)X_0+t\epsilon,\qquad V^\star=\epsilon-X_0,
\label{eq:flow-path}\\
\mathcal L_{\mathrm{FM}}^{\mathrm{base}}
&=
\mathbb E_{X_0,\epsilon,t}
\left[\left\|V_\theta(X_t,t\mid c)-V^\star\right\|_2^2\right], \text{respectively}.
\label{eq:background-flow}
\end{align}
Below, we augment DiTs with the \method{} using the same formulation.

\subsection{Physical Registry}
\label{sec:registry-construction}

\method{} \rev{represents the forecast in a fixed set of dedicated registry slots
that carry predictive features alongside the video latent rather than within it.}
Let \(X_t\in\mathbb R^{B\times C\times T_v\times H\times W}\) denote the noisy
video latent. 
We use the frozen JEPA encoder $E$ and predictor $P$. The encoder receives a full video and produces JEPA features for individual frames. The predictor receives the JEPA feature of the first frame and produces future representations in the shared representation space\citep{bardes2025vjepa,assran2025vjepa2}.
Given 
an initial observed frame \(x\),
we retain the first encoder-derived feature as
an observed context and use the predictor to forecast the remaining
features:
\begin{equation}
\begin{aligned}
H^{\mathrm{enc}} &= E(x),\\
J^{\mathrm{cond}} &=
\left[
H^{\mathrm{enc}};
P\!\left(H^{\mathrm{enc}};n_{\mathrm{context}}=1\right)
\right]
\in\mathcal J^{T_r}.
\end{aligned}
\label{eq:forecast}
\end{equation}
We define \(J^{\mathrm{cond}}\) to be the \emph{forecast}, with \(T_r=8\): one
encoder-derived feature for the observed context and seven predictor-generated
features for the subsequent future temporal positions. The forecast occupies \(T_r\) positions, called \emph{registry slots}, placed after the video tokens. Each slot is a spatial grid of
\emph{registry tokens}
which are the units for attention, whereas slots correspond to frames.

A learned spatio-temporal adapter \(G_\phi\)
(Appendix~\ref{sec:projection-architecture}) maps the forecast into
the video latent's shape and coordinates. At each denoising step \(t\), the same forecast initializes
the registry, which is concatenated with the noisy video latent along the
temporal dimension:
\begin{equation}
R_t^{(0)}=G_\phi\!\left(J^{\mathrm{cond}}\right)
\in
\mathbb R^{B\times C\times T_r\times H\times W},
\qquad
Z_t^{(0)}=[X_t;R_t^{(0)}]
\in
\mathbb R^{B\times C\times(T_v+T_r)\times H\times W}.
\label{eq:registry-projection}
\end{equation}

\postsub{To place} the registry on the \postsub{same temporal} axis as the video\postsub{, each} registry slot
\postsub{is assigned} the RoPE coordinate of its \postsub{corresponding video} time. Let
\(u_s\in[0,1)\) denote the normalized clip time \postsub{of} registry slot \(s\), and let
\(\rho_v\) map normalized time to the corresponding video-latent RoPE
coordinate. \postsub{We use}
\begin{equation}
\tau^{\mathrm{reg}}_s=\rho_v(u_s),
\qquad
s\in\{1,\ldots,T_r\}\postsub{,}
\label{eq:registry-rope}
\end{equation}
\postsub{with \(u_s=(s-1)/T_r\) for} uniformly spaced registry slots\postsub{.}

\postsub{After patchification,} video and registry \postsub{tokens remain distinct but are jointly
processed by} the \postsub{same bidirectional self-attention blocks. Although the} registry
\postsub{is} initialized from the \postsub{same observation-conditioned} forecast \postsub{at each denoising
step, interaction with the evolving video tokens makes its hidden state
candidate-specific as} it \postsub{propagates through} the \postsub{DiT.}

\subsection{Predictive supervision of the registry}
\label{sec:registry-alignment}

We supervise the registry toward representations of the realized future.
Conditioning and target features play complementary roles:
\(J^{\mathrm{cond}}\) is an observation-conditioned forecast that initializes
\(R_t^{(0)}\), while \(Z^{\mathrm{tgt}}\) is the target encoding of the corresponding full training video \(v\):
\begin{equation}
Z^{\mathrm{tgt}}
=
E^{\mathrm{tgt}}(v)
=
\left(
Z^{\mathrm{tgt}}_1,\ldots,Z^{\mathrm{tgt}}_{T_r}
\right),
\qquad
Z^{\mathrm{tgt}}_s\in\mathcal J.
\label{eq:jepa-target}
\end{equation}
\rev{Because \(J^{\mathrm{cond}}\) depends only on the observation, the registry
starts each denoising step from an observation-only prediction. As
the registry exchanges information with the video tokens, predictive
supervision encourages this state to align with the representation of the future
realized in the training video.}
We apply the supervision at an intermediate DiT block so that \postsub{the registry is
shaped toward the realized future and} can \postsub{subsequently interact with the video
states throughout the remaining DiT blocks.} Let \(H_{\mathrm{reg},t,s}\) denote the hidden \postsub{state} of registry slot
\(s\) at the supervised layer and denoising step \(t\).
A readout \(q_\psi\) maps each registry state to a slot-level vector in the
JEPA feature space:
\begin{equation}
\widehat Z_{t,s}
=
q_\psi\!\left(H_{\mathrm{reg},t,s}\right).
\label{eq:registry-readout}
\end{equation}
Writing \(\overline Z_s^{\mathrm{tgt}}\) for the corresponding slot-level
target, we optimize
\begin{equation}
\mathcal L_{\mathrm{align}}
=
-\frac{1}{T_r}
\sum_s
\cos\!\left(
\widehat Z_{t,s},
\overline Z_s^{\mathrm{tgt}}
\right).
\label{eq:registry-alignment}
\end{equation}

Appendix~\ref{app:implementation} specifies the readout and target construction.
\rev{After training, the same readout \(q_\psi\) is frozen and reused for selection
in \cref{sec:selection-logic}.}

\subsection{Training objective}
\label{sec:training-objective}

Given the augmented sequence
\(Z_t^{(0)}=[X_t;R_t^{(0)}]\), the DiT jointly processes the video and
registry tokens. We denote by
\(V_\theta^X(Z_t^{(0)},t,c_{\mathrm{txt}})\) the velocity prediction from
the video tokens. The flow-matching objective is
\begin{equation}
\mathcal L_{\mathrm{FM}}
=
\mathbb E_{X_0,\epsilon,t}
\left[
\left\|
V_\theta^X\!\left(Z_t^{(0)},t,c_{\mathrm{txt}}\right)
-V^\star
\right\|_2^2
\right].
\label{eq:flow-loss}
\end{equation}
Both terms optimize the shared DiT:
\begin{equation}
\mathcal L
=
\mathcal L_{\mathrm{FM}}
+\lambda_{\mathrm{align}}\mathcal L_{\mathrm{align}}.
\label{eq:total-loss}
\end{equation}
The video tokens receive the generation objective and the registry slots
receive predictive supervision through the same DiT blocks, so the registry
carries future-referenced information while the backbone continues to learn
generation.

\input{sections/main/04_selection_logic}

%% file: figures/main/predictive_registry.tex
\begin{figure}[t]
\centering
\caption{\textbf{\rev{Physical Registry during training and inference.}}
\rev{The Physical Registry gives predicted future dynamics a dedicated, readable
state inside the video DiT. (a) Training: forecast-initialized registry tokens
co-evolve with video tokens and receive predictive supervision. (b) Inference:
the same registry provides readouts for PRS (Section~\ref{sec:selection-logic}).}}
\label{fig:physical-registry}
\includegraphics[width=\linewidth]{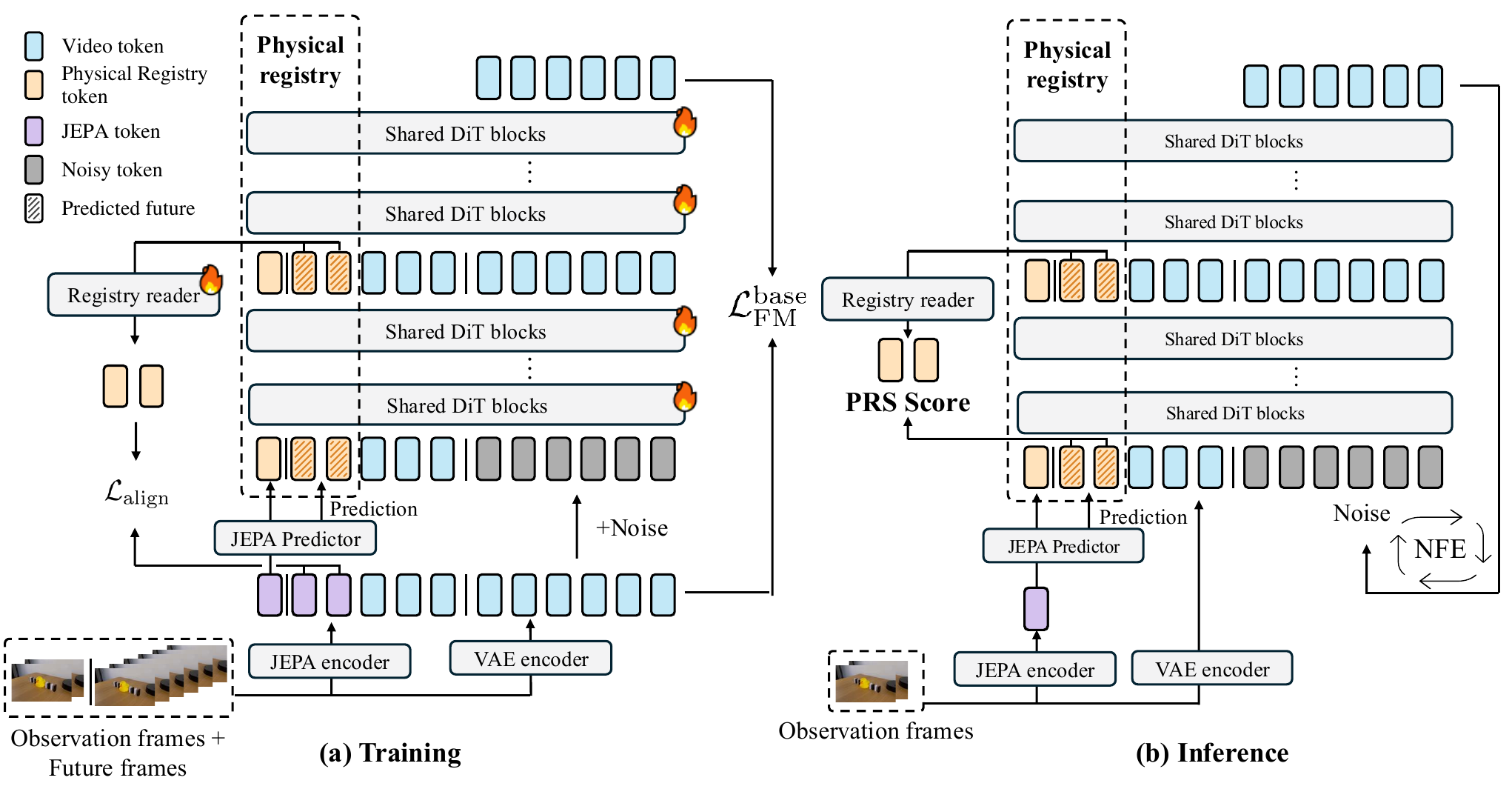}
\end{figure}

%% file: sections/main/04_selection_logic.tex
\subsection{The Physical Registry Selector}
\label{sec:selection-logic}

\postsub{Through joint interaction with} the \postsub{video tokens and predictive supervision,
the Physical Registry becomes a} readable \postsub{internal state that reflects the}
future dynamics \postsub{represented by each generation trajectory.
This makes it possible to compare candidate trajectories} before generation is
complete \postsub{using the registry states themselves.}

For a fixed observation, all candidates \postsub{start from} the same
observation-conditioned forecast, \postsub{but} their registry states \postsub{become}
candidate-specific \postsub{as they interact with different video trajectories during
denoising. We therefore interpret the departure of each registry} from \postsub{the}
shared forecast \postsub{as} the trajectory-specific update \postsub{to the predicted future.}
The \emph{Physical Registry Selector (PRS)} ranks candidates by this departure
and continues denoising only the selected trajectory for the remaining steps. For input \(x\), let the \(N\) candidate trajectories be
\(\mathcal C(x)=\{Y^{(n)}\}_{n=1}^{N}\), \postsub{all initialized from} the \postsub{same}
forecast \(J^{\mathrm{cond}}\)\postsub{, which we denote by
\(\overline J^{\mathrm{cond}}\) when used as their shared reference.}
After \(K\) denoising steps, PRS compares each candidate's registry readouts
with \postsub{\(\overline J^{\mathrm{cond}}\), selects the trajectory whose registry
reflects the largest candidate-specific update to the} shared forecast\postsub{,} and
continues \postsub{denoising it} from its current latent for the remaining \(D-K\) steps,
where \(D\) is the total number of denoising steps
\postsub{(Figure~\ref{fig:selector-procedure}).}

\input{figures/main/selector_procedure}

The readout \(q_\psi\) maps the registry hidden state
\(H_{k,s}^{(n)}\) into the JEPA feature space.
The \emph{forecast-departure score} for candidate \(n\), accumulated over
\(K\) denoising steps and a registry subset \(\mathcal S\), is
\begin{equation}
\begin{aligned}
r^{(n)}
&=
1-
\frac{1}{K|\mathcal S|}
\sum_{k=1}^{K}\sum_{s\in\mathcal S}
\cos\!\left(
q_\psi\!\left(H_{k,s}^{(n)}\right),
\overline{J}^{\mathrm{cond}}_s
\right),
\\
\widehat n
&=
\arg\max_{n\in\{1,\ldots,N\}}
r^{(n)}.
\end{aligned}
\label{eq:selector-main}
\end{equation}

The score measures angular departure from the shared forecast \postsub{and is invariant to positive rescaling of the registry readouts.} Appendix~\ref{app:prs-direction} empirically supports this ranking direction and its robustness.

%% file: figures/main/selector_procedure.tex
\begingroup
\definecolor{drForecast}{HTML}{78549C}
\definecolor{drForecastFill}{HTML}{D4C1EE}
\definecolor{drModule}{HTML}{042433}
\definecolor{drModuleFill}{HTML}{F3F3F4}
\definecolor{drVideo}{HTML}{3880B8}
\definecolor{drVideoFill}{HTML}{C2E8F7}
\definecolor{drRegistry}{HTML}{DB8529}
\definecolor{drRegistryFill}{HTML}{FFE2BB}
\definecolor{drSelect}{HTML}{042433}
\definecolor{drSelectFill}{HTML}{F3F3F4}
\begin{figure}[!htbp]
\centering
\caption{\textbf{Three-stage prefix selection with PRS.}
Candidates share the same forecast initialization, are ranked from registry
readouts over a \(K\)-step prefix, and only the selected trajectory continues
for the remaining \(D-K\) steps.}
\label{fig:selector-procedure}
\expandafter\def\csname selprc@noise@1\endcsname{\includegraphics[width=6.6mm]{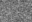}}
\expandafter\def\csname selprc@noise@2\endcsname{\includegraphics[width=6.6mm]{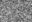}}
\expandafter\def\csname selprc@noise@n\endcsname{\includegraphics[width=6.6mm]{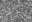}}
\expandafter\def\csname selprc@prefix@1\endcsname{\includegraphics[width=6.4mm]{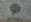}}
\expandafter\def\csname selprc@prefix@2\endcsname{\includegraphics[width=6.4mm]{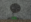}}
\expandafter\def\csname selprc@prefix@n\endcsname{\includegraphics[width=6.4mm]{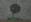}}
\expandafter\def\csname selprc@xk@1\endcsname{\includegraphics[width=9.6mm]{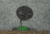}}
\expandafter\def\csname selprc@xk@2\endcsname{\includegraphics[width=9.6mm]{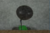}}
\expandafter\def\csname selprc@xk@n\endcsname{\includegraphics[width=9.6mm]{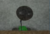}}
\resizebox{\linewidth}{!}{%
\begin{tikzpicture}[
  x=1cm,y=1cm,font=\scriptsize,
  box/.style={draw=black!40,rounded corners=1pt,align=center,inner sep=1.6pt},
  reg/.style={box,draw=drRegistry,fill=drRegistryFill},
  vid/.style={box,draw=drVideo,fill=drVideoFill},
  forecaststate/.style={box,draw=drForecast,fill=drForecastFill},
  module/.style={box,draw=drModule,fill=drModuleFill},
  eq/.style={draw=drModule!45,fill=white,rounded corners=2pt,align=left,inner sep=2.4pt,
    line width=.4pt,drop shadow={shadow xshift=.5pt,shadow yshift=-.6pt,opacity=.18,fill=black}},
  eqlab/.style={font=\fontsize{3.6}{4.3}\selectfont\sffamily\bfseries,text=drModule!70,inner sep=0pt},
  token/.style={draw=drRegistry,fill=drRegistryFill,rounded corners=.4pt},
  arrow/.style={-{Latex[length=1.3mm,width=.9mm]},line width=.45pt,draw=black!55},
  regarrow/.style={arrow,draw=drRegistry},
  forecastarrow/.style={arrow,draw=drForecast},
  keep/.style={arrow,draw=drSelect,line width=.8pt},
  frame/.style={inner sep=0pt,draw=black!30,line width=.3pt},
]

\def\rowA{4.70}\def\rowB{3.55}\def\rowN{2.40}\def\fcy{1.05}
\def\sA{4.40}\def\sB{2.95}\def\sN{1.50}
\foreach \xa/\xb in {0/6.75, 6.95/11.35, 11.55/17.45}{
  \path[draw=black!12,fill=black!2,rounded corners=3pt,line width=.4pt]
    (\xa,.45) rectangle (\xb,6.78);
  \begin{scope}
    \clip[rounded corners=3pt] (\xa,.45) rectangle (\xb,6.78);
    \path[fill=drModule!9] (\xa,6.22) rectangle (\xb,6.78);
  \end{scope}
  \draw[draw=drModule!18,line width=.4pt] (\xa,6.22) -- (\xb,6.22);
}
\foreach \x/\n/\t in {3.375/1/{Generate candidate prefixes}, 9.15/2/{Read out and compare}, 14.50/3/{Pick one, finish it}}{
  \node[font=\scriptsize\bfseries,text=drModule,anchor=base] (hd\n) at ({\x+.16},6.43) {\t};
  \node[circle,fill=drModule,text=white,font=\fontsize{5.5}{6}\selectfont\bfseries,
        inner sep=0pt,minimum size=3.1mm,anchor=east] at ([xshift=-3pt,yshift=2.3pt]hd\n.base west) {\n};
}

\node[font=\tiny,anchor=base] at (1.02,5.30) {Denoising \(K\) steps};
\node[font=\tiny,anchor=base] at (4.05,5.36) {Video DiT};
\foreach \f/\tag/\y in {1/1/\rowA, 2/2/\rowB, n/N/\rowN}{
  \node[frame] (noise\f) at (.42,\y)
    {\csname selprc@noise@\f\endcsname};
  \node[font=\tiny,anchor=north,inner sep=1pt] at (noise\f.south)
    {\(\varepsilon^{(\tag)}\)};
  \node[font=\tiny] at (1.03,\y) {\(\cdots\)};
  \node[frame] (pre\f) at (1.62,\y)
    {\csname selprc@prefix@\f\endcsname};
  \node[draw=black!30,rounded corners=1.6pt,fill=white,
        minimum width=30.5mm,minimum height=10.0mm,inner sep=0pt] (trunk\f)
    at (4.05,\y) {};
  \node[box,fill=white,font=\tiny,minimum height=7.4mm,minimum width=6.4mm]
    at (3.05,\y) {DiT\\blocks};
  \node[box,fill=white,font=\tiny,minimum height=7.4mm,minimum width=6.4mm]
    at (5.05,\y) {DiT\\blocks};
  \path[reg] (3.45,{\y+.19}) rectangle ++(1.20,.155);
  \path[vid] (3.45,{\y-.08}) rectangle ++(1.20,.155);
  \path[reg] (3.45,{\y-.35}) rectangle ++(1.20,.155);
  \draw[arrow] (noise\f.east) -- (.87,\y);
  \draw[arrow] (pre\f.east) -- (trunk\f.west);
}
\node[font=\tiny,anchor=base west] at (.10,1.40) {JEPA forecast \(J^{\mathrm{cond}}\)};
\node[draw=drForecast,rounded corners=.7pt,inner sep=0pt,
      minimum width=8.6mm,minimum height=4.2mm] (jcond) at (.55,\fcy) {};
\foreach \x/\h in {.18/0,.43/1,.68/1}{\path[draw=drForecast,fill=drForecastFill,rounded corners=.4pt] (\x,{\fcy-.145}) rectangle ++(.2,.29);
  \ifnum\h=1 \path[pattern=north east lines,pattern color=drForecast] (\x,{\fcy-.145}) rectangle ++(.2,.29);\fi}
\node[module,font=\tiny,minimum width=5.4mm,minimum height=3.4mm]
  (gphi) at (1.62,\fcy) {\(G_\phi\)};
\node[draw=drRegistry,rounded corners=.7pt,inner sep=0pt,
      minimum width=15.6mm,minimum height=4.2mm] (tokens) at (2.85,\fcy) {};
\foreach \x/\h in {2.17/0,2.55/1,3.27/1}{\path[token] (\x,{\fcy-.145}) rectangle ++(.26,.29);
  \ifnum\h=1 \path[pattern=north east lines,pattern color=drRegistry] (\x,{\fcy-.145}) rectangle ++(.26,.29);\fi}
\node[font=\tiny,text=drRegistry] at (3.04,\fcy) {\(\cdots\)};
\node[font=\tiny,anchor=west,inner sep=2pt] at (3.67,\fcy) {\(R_t^{(0)}\)};
\draw[forecastarrow] (jcond.east) -- (gphi.west);
\draw[regarrow] (gphi.east) -- (tokens.west);
\path[draw=drRegistry,dashed,line width=.45pt,rounded corners=2pt]
  (2.42,{\rowN-.58}) rectangle (5.68,{\rowA+.58});
\draw[regarrow] (tokens.north) -- (2.85,{\rowN-.58});
\foreach \dy/\sty/\lab in {.40/reg/{Physical Registry token}, .15/vid/{Video token}, -.10/forecaststate/{JEPA token}}{
  \path[\sty] (4.24,{\fcy+\dy-.09}) rectangle ++(.30,.18);
  \node[font=\tiny,anchor=west,inner sep=1.5pt] at (4.56,{\fcy+\dy}) {\lab};}
\path[draw=black!60,fill=white,rounded corners=.4pt] (4.24,{\fcy-.44}) rectangle ++(.30,.18);
\path[pattern=north east lines,pattern color=black!60] (4.24,{\fcy-.44}) rectangle ++(.30,.18);
\node[font=\tiny,anchor=west,inner sep=1.5pt] at (4.56,{\fcy-.35}) {Predicted future};

\node[eq,font=\tiny] (eq) at (9.15,5.60)
  {\begin{tabular}{@{}r@{\hspace{2.5pt}}l@{}}
   \tikz[baseline]\node[eqlab,anchor=base]{COSINE}; & \(a_{k,s}^{(n)}=\cos\!\bigl(q_\psi(H_{k,s}^{(n)}),\overline J_s^{\mathrm{cond}}\bigr)\)\\[3pt]
   \tikz[baseline]\node[eqlab,anchor=base]{DEPARTURE}; & \(r_n=1-\operatorname{mean}_{k,s}\,a_{k,s}^{(n)}\)
   \end{tabular}};
\foreach \f/\tag/\y in {1/1/\rowA, 2/2/\rowB, n/N/\rowN}{
  \node[reg,minimum width=12mm,minimum height=4.6mm] (readout\f) at (7.75,\y)
    {\(H_{k,s}^{(\tag)}\)};
  \draw[regarrow] (trunk\f.east) -- (readout\f.west);
}
\node[module,minimum width=6.4mm,minimum height=5.6mm]
  (qpsi) at (9.15,\rowB) {\(q_\psi\)};
\node[circle,draw=drRegistry,fill=white,inner sep=1.7pt] (cosine) at (10.45,\rowB) {cos};
\draw[regarrow] (readout1.east) -- (8.55,\rowA) -- (8.55,\rowB) -- (qpsi.west);
\draw[regarrow] (readoutn.east) -- (8.55,\rowN) -- (8.55,\rowB) -- (qpsi.west);
\draw[regarrow] (readout2.east) -- (qpsi.west);
\draw[regarrow] (qpsi.east) -- (cosine.west);
\node[forecaststate,font=\tiny,minimum width=17mm,minimum height=6.6mm]
  (shared) at (10.45,\fcy) {Shared forecast\\[1pt]\(\overline J_s^{\mathrm{cond}}\)};
\draw[forecastarrow] (shared.north) -- (cosine.south);
\draw[forecastarrow] (jcond.south) -- (.55,.51) -- (10.45,.51) -- (shared.south);
\draw[arrow] (cosine.north) -- (cosine.north |- eq.south)
  node[pos=.45,right,font=\tiny,inner sep=1.5pt] {\(a_{k,s}^{(n)}\)};

\node[eq,font=\tiny] (pick) at (14.50,5.60)
  {\tikz[baseline]\node[eqlab,anchor=base]{SELECT};\hspace{4pt}\(\widehat n=\arg\max_n r_n\)};
\draw[keep] (eq.east) -- (pick.west);
\foreach \f/\tag/\y/\val/\len/\col in {%
  1/1/\sA/{0.123}/0.185/{black!45},
  2/2/\sB/{0.950}/1.425/{drSelect},
  n/N/\sN/{0.235}/0.353/{black!45}}{
  \node[font=\tiny,anchor=east,inner sep=1pt] at (12.25,{\y+.30}) {\(r_{\tag}\)};
  \node[font=\tiny,anchor=west,inner sep=1pt] at (12.33,{\y+.30}) {\val};
  \path[draw=black!30,fill=black!8,rounded corners=.4pt]
    (11.85,{\y-.13}) rectangle ++(1.30,.26);
  \path[draw=none,fill=\col,rounded corners=.4pt]
    (11.85,{\y-.13}) rectangle ++({\len*1.3/1.5},.26);
  \node[frame] (xk\f) at (13.95,\y)
    {\csname selprc@xk@\f\endcsname};
  \node[font=\tiny,anchor=north,inner sep=1pt] at (xk\f.south) {\(X_K^{(\tag)}\)};
}
\draw[draw=red!75!black,dashed,line width=.5pt,rounded corners=1pt]
  (11.75,{\sB-.24}) rectangle (13.25,{\sB+.50});
\node[font=\tiny\bfseries,text=red!80!black,anchor=north,inner sep=1pt]
  at (12.50,{\sB-.26}) {pick!};
\foreach \y in {\sA,\sN}{
  \draw[draw=black!45,line width=.9pt] (14.62,{\y-.20}) -- (14.62,{\y+.20});
  \node[font=\tiny,text=black!55,anchor=west,inner sep=1.5pt] at (14.68,\y) {Stop};
}
\node[frame,draw=drSelect,line width=.7pt] (output) at (16.45,\sB)
  {\includegraphics[width=17.5mm]{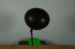}};
\draw[keep] (xk2.east) -- node[midway,above=1pt,align=center,font=\tiny,text=drSelect]
  {denoising\\\(D-K\)} (output.west);
\node[font=\tiny,anchor=north,inner sep=2pt] at (output.south) {Completed video};
\end{tikzpicture}}
\end{figure}
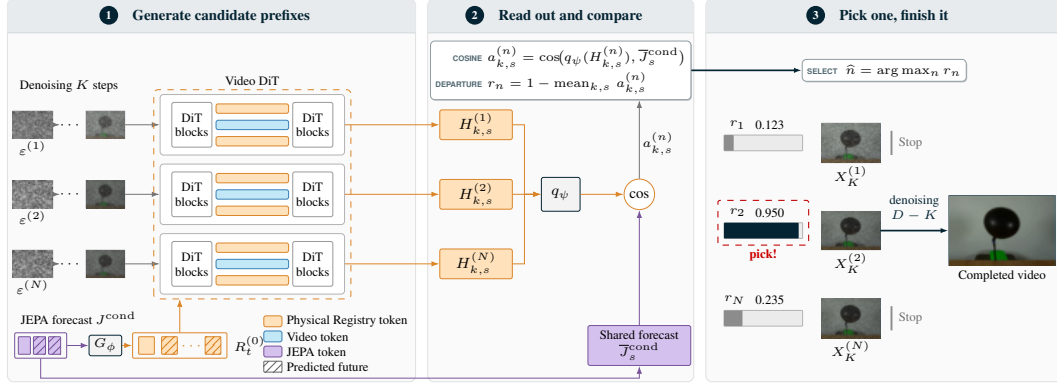
\endgroup

%% file: sections/main/04_experiments.tex
\section{Experiments}
\label{sec:experiments}

\postsub{We first show that \method{} is a more effective interface for predictive
representations than conditioning and joint generation
(Sec.~\ref{sec:results}), and that its gains generalize across video backbones
and world-model benchmarks (Sec.~\ref{sec:main-results}). We then show that its
registry states support candidate selection during generation
(Sec.~\ref{sec:selector-analysis}), before analyzing the sources of its gains
and its responses to physical changes
(Secs.~\ref{sec:registry-ablation} and~\ref{sec:analysis}).}

\input{sections/main/04_experimental_setup}
\input{sections/main/05_results}

%% file: sections/main/04_experimental_setup.tex
\subsection{Experimental Details}

\paragraph{Benchmarks and tasks.}
We evaluate on Physics-IQ Verified \citep{motamed2026physicsiqverified},
which scores physical plausibility against recorded futures, and on
WorldModelBench \citep{li2025worldmodelbench}, which scores instruction
following, physical adherence, and common sense over its full 350-prompt set.
Metric definitions are in Appendix~\ref{app:setup-detail}.

\paragraph{Multi-backbone evaluation.}
\leavevmode\rev{
We use Cosmos and WAN 2.2 as two backbone families: Cosmos is a widely used world-model backbone that supports both I2V and V2V continuation, while WAN 2.2 is a widely adopted video-generation backbone in recent world-model research. Since WAN does not provide a V2V setting, we evaluate it on I2V only. Models are fine-tuned on the Enlarge dataset, which combines physics-oriented, web, simulation, and robot videos
\citep{wang2025wisa,wang2023internvid,nvidia2026physxsimdataset,nvidia2026robotsimdataset} while excluding Physics-IQ videos
(Appendix~\ref{app:training-data}). Each backbone--\method{} pair shares the task, conditioning input, sampling protocol, and evaluator.}

\paragraph{Candidate selection.}
\leavevmode\rev{
\postsub{All methods use} the same ten-candidate \postsub{pool per input.} Plain reports the \postsub{pool
average,} while \postsub{WMReward and PRS select one candidate using completed-video
scores and intermediate registry readouts, respectively. Published leaderboard
values are used for Plain where available.}}
Additional details are in Appendix~\ref{app:selector-protocol}.

%% file: sections/main/05_results.tex
\subsection{Controlled Comparison of Predictive-State Interfaces}
\label{sec:results}

We \postsub{compare alternative interfaces for incorporating} predictive representations
into the video generator \postsub{under} a controlled \postsub{held-view setting.}
Each variant is adapted on the 66 center-view \postsub{Physics-IQ} videos and evaluated on
the 132 held-out left and right views using Physics-IQ Verified\postsub{, with} the same
pretrained backbone, training budget, and frozen JEPA source. \rev{Table~\ref{tab:conditioning-ablation} \postsub{compares conditioning} and \postsub{modulation,}
\postsub{video-stream alignment, joint generation,} and \postsub{dedicated} registry states
(Appendix~\ref{app:setup-detail}).}

\input{tables/main/conditioning_ablation}

\paragraph{\rev{\postsub{Dedicated predictive states benefit} most \postsub{from predictive supervision.}}}
\rev{Table~\ref{tab:conditioning-ablation} \postsub{shows that a dedicated registry state is
already beneficial without predictive supervision: Registry only improves
vanilla fine-tuning by \(+3.7/+3.9\) on I2V/V2V. In contrast, applying
representation alignment directly to} the \postsub{video stream with} REPA \postsub{yields} only
\(+0.5/+0.5\)\postsub{. Adding predictive} supervision \postsub{to the registry further raises the
scores to \(51.2/55.1\), the best among the compared interfaces. These results
suggest that explicitly allocating} a \postsub{dedicated internal state for predictive
representations} is \postsub{beneficial, and that predictive supervision is important for
assigning this state a meaningful role in representing future dynamics.}}

\subsection{Fine-Tuning on \texorpdfstring{\rev{the Enlarge Dataset}}{the Enlarge Dataset} across Video Backbones}
\label{sec:main-results}

\input{tables/main/cosmos_results}

\paragraph{\method{} improves \postsub{physical plausibility} across backbones.}
\leavevmode\rev{
\postsub{\method{} improves Physics-IQ Verified over every evaluated backbone
(}Table~\ref{tab:main-results}\postsub{).}
Under the expanded-prompt \postsub{setting (}Appendix~\ref{app:prompt-expansion}\postsub{), it raises} Cosmos 3 Nano 16B from
\(37.3/46.45\) to \(43.56/49.54\) on I2V/V2V, \postsub{surpassing} the Cosmos 3
Super 64B reference (\(42.7/48.02\)) \postsub{with} only \(0.45\%\) \postsub{additional}
parameters (Table~\ref{tab:training-configuration}).
\postsub{The} WAN TI2V-5B matched-budget control \postsub{further shows that the gain is not
explained by fine-tuning alone:} vanilla fine-tuning on the Enlarge dataset
\postsub{remains} near pretrained performance\postsub{, whereas} \method{} improves
(Appendix~\ref{app:wan5b-matched-budget}).
Figure~\ref{fig:qual-physicsiq} \postsub{further} shows \postsub{more faithful realization of
the intended physical events} across five physics categories.}

\input{figures/main/qualitative_physicsiq}

\paragraph{\method{} also improves on other physics-related benchmarks.}
\leavevmode\rev{
On WorldModelBench, \method{} improves physical adherence and the total score
\postsub{across} all four backbones\postsub{. Common} sense \postsub{is maintained} or \postsub{improved, while}
instruction following decreases on Cosmos v2.5 and WAN TI2V-5B
(Table~\ref{tab:worldmodelbench}; full leaderboard in
Table~\ref{tab:worldmodelbench-leaderboard}, \postsub{qualitative} examples in
Appendix~\ref{app:worldmodelbench-qualitative}).
On EWMBench \citep{hu2025ewmbench}, \method{} \postsub{also} improves Cosmos 3 Nano's
scene consistency and all three motion metrics, \postsub{with the largest gain} in
trajectory agreement (Appendix~\ref{app:ewmbench}).}
\input{tables/main/worldmodelbench}

\subsection{Selection from Registry Readouts}
\label{sec:selector-analysis}
\label{sec:selector-robustness}
\input{figures/main/selection_comparison}

\rev{Figure~\ref{fig:selection-comparison} compares PRS with Plain generation and
completed-video WMReward. \postsub{Under the same WMReward selector,} \method{} improves
over its \postsub{corresponding} backbone \postsub{across} all \postsub{evaluated I2V} and \postsub{V2V settings.}
PRS \postsub{improves} I2V over Plain by \(5.6\%\)--\(15.9\%\) across the four default
backbone configurations, \postsub{using only} intermediate registry readouts without
completed-video external scoring.}

On Cosmos v2.5, PRS \postsub{selects} after 10 of 35 denoising steps \postsub{and} reaches
\(38.36/46.68\) on I2V/V2V, \postsub{slightly exceeding} WMReward's
\(37.74/46.59\)\postsub{.} Continuing only the selected trajectory \postsub{requires 125
candidate-steps instead of 350, reducing} the serial denoising budget \postsub{by
\(64\%\). Together, these results show that registry readouts provide} a \postsub{useful
early signal for candidate selection before generation is complete.}
\rev{Appendices~\ref{app:selection-analysis} and~\ref{app:prs-early-selection}
\postsub{provide} the full results and early-selection analysis.}

\input{sections/main/05_ablation}

%% file: tables/main/conditioning_ablation.tex
\begingroup
\newcommand{\yesmark}{\textcolor{green!45!black}{\(\checkmark\)}}
\newcommand{\nomark}{\textcolor{red!70!black}{\(\boldsymbol{\times}\)}}
\begin{table}[!htb]
\centering
\caption{\textbf{Controlled comparison of predictive-state interfaces.}
\rev{Interface mechanisms and Physics-IQ Verified scores
after center-view training and left/right evaluation, with std over three
generation seeds and \(\Delta\) relative to vanilla fine-tuning.
Registry only sets \(\lambda_{\mathrm{align}}=0\) in Eq.~(\ref{eq:total-loss}).}}
\label{tab:conditioning-ablation}
\scriptsize
\setlength{\tabcolsep}{2pt}
\renewcommand{\arraystretch}{1.0}
\begin{tabular*}{\linewidth}{@{\extracolsep{\fill}}llcccrrrr@{}}
\toprule
Variant & Mechanism &
Registry &
Alignment &
Readout &
I2V & \(\Delta\) & V2V & \(\Delta\) \\
\midrule
Vanilla fine-tuning & -- & \nomark & \nomark & \nomark & \scorestd{41.3}{2.1} & -- & \scorestd{46.4}{1.0} & -- \\
REPA \citep{yu2025repa} & Feature alignment & \nomark & \yesmark & \nomark & \scorestd{41.8}{1.7} & +0.5 & \scorestd{46.9}{0.9} & +0.5 \\
Cross-attention \citep{peebles2023dit} & External context & \nomark & \nomark & \nomark & \scorestd{43.1}{1.9} & +1.8 & \scorestd{49.2}{1.2} & +2.8 \\
AdaLN \citep{peebles2023dit} & Feature modulation & \nomark & \nomark & \nomark & \scorestd{45.1}{0.8} & +3.8 & \scorestd{48.3}{0.5} & +1.9 \\
Joint generation \citep{kouzelis2025redi} & Joint denoising & \nomark & \yesmark & \nomark & \scorestd{45.1}{1.8} & +3.8 & \scorestd{48.0}{0.4} & +1.6 \\
\midrule
Registry only & \textbf{Dedicated predictive state} & \yesmark & \nomark & \yesmark & \scorestd{45.0}{1.3} & +3.7 & \scorestd{50.3}{0.9} & +3.9 \\
\prrow\textbf{\methodfull{}} & \textbf{Dedicated predictive state} & \yesmark & \yesmark & \yesmark & \scorestd{\textbf{51.2}}{1.5} & \textbf{+9.9} & \scorestd{\textbf{55.1}}{1.1} & \textbf{+8.7} \\
\bottomrule
\end{tabular*}

\end{table}
\endgroup

%% file: tables/main/cosmos_results.tex
\begin{table}[!tbp]
\centering
\caption{\textbf{Physics-IQ Verified.}
Plain averages our candidate pool or \postsub{reports} the published value
(\textsuperscript{*}).
\(\Delta\) is the gain over the backbone above, green parentheses the PRS gain
over Plain.
Bold and underline mark the best and second-best \postsub{per} task.
Gray marks unavailable entries, dashes unreported.}
\label{tab:main-results}
\begingroup
\fontsize{7.6}{9.0}\selectfont
\renewcommand{\scorestd}[2]{#1\,\textpm\,#2}
\setlength{\tabcolsep}{3pt}
\renewcommand{\arraystretch}{1.02}
\begin{tabular*}{\linewidth}{@{\extracolsep{\fill}}lrrrrrr@{}}
\toprule
Model &
\multicolumn{3}{c}{I2V} &
\multicolumn{3}{c}{V2V} \\
\cmidrule(lr){2-4}\cmidrule(lr){5-7}
& Plain & \(\Delta\) & PRS & Plain & \(\Delta\) & PRS \\
\midrule
Cosmos 3 Nano 16B \citep{nvidia2026cosmos3}
& \scorestd{30.18}{0.84} & -- & \DRSUnavailable & \scorestd{38.78}{0.78} & -- & \DRSUnavailable \\
\prrow\quad + \methodfull{}
& \scorestd{36.01}{0.95} & +5.83 & \prsscore{40.27}{+4.26} & \scorestd{45.12}{0.94} & +6.34 & \prsscore{46.78}{+1.66} \\
\midrule
Cosmos 3 Nano 16B\textsuperscript{\textdagger} \citep{nvidia2026cosmos3}
& \scorestd{37.3}{0.9}\textsuperscript{*} & -- & \DRSUnavailable & \scorestd{46.45}{0.76} & -- & \DRSUnavailable \\
\prrow\quad + \methodfull{}\textsuperscript{\textdagger}
& \scorestd{\textbf{43.56}}{0.95} & +6.26 & \prsscore{45.81}{+2.25} & \scorestd{\textbf{49.54}}{0.63} & +3.09 & \prsscore{52.06}{+2.52} \\
\midrule
Cosmos v2.5 2B \citep{nvidia2025cosmospredict25}
& \scorestd{30.65}{0.37} & -- & \DRSUnavailable & \scorestd{37.40}{0.53} & -- & \DRSUnavailable \\
\prrow\quad + \methodfull{}
& \scorestd{33.71}{1.11} & +3.06 & \prsscore{39.08}{+5.37} & \scorestd{42.33}{1.90} & +4.93 & \prsscore{47.67}{+5.34} \\
\midrule
WAN 2.2 I2V-A14B \citep{wanteam2025wan22}
& \scorestd{32.2}{0.6}\textsuperscript{*} & -- & \DRSUnavailable & \TaskUnavailable & \TaskUnavailable & \TaskUnavailable \\
\prrow\quad + \methodfull{}
& \scorestd{38.44}{1.06} & +6.24 & \prsscore{40.58}{+2.14} & \TaskUnavailable & \TaskUnavailable & \TaskUnavailable \\
\midrule
WAN 2.2 TI2V-5B \citep{wanteam2025wan22}
& \scorestd{27.7}{0.9}\textsuperscript{*} & -- & \DRSUnavailable & \TaskUnavailable & \TaskUnavailable & \TaskUnavailable \\
\prrow\quad + \methodfull{}
& \scorestd{29.24}{1.30} & +1.54 & \prsscore{32.03}{+2.79} & \TaskUnavailable & \TaskUnavailable & \TaskUnavailable \\
\midrule
Cosmos 3 Super 64B\textsuperscript{\textdagger} \citep{nvidia2026cosmos3}
& \scorestd{\underline{42.7}}{0.8}\textsuperscript{*} & -- & \DRSUnavailable & \scorestd{48.02}{0.75} & -- & \DRSUnavailable \\
MiniMax H3 \citep{minimax2026h3}
& \scorestd{39.8}{0.3}\textsuperscript{*} & -- & \DRSUnavailable & \scorestd{37.71}{1.75} & -- & \DRSUnavailable \\
Grok Imagine Video \citep{anates2026physicsiqleaderboard}
& \scorestd{34.8}{0.6}\textsuperscript{*} & -- & \DRSUnavailable & \TaskUnavailable & \TaskUnavailable & \TaskUnavailable \\
HunyuanVideo 1.5 \citep{motamed2026physicsiqverified}
& \scorestd{33.4}{0.8}\textsuperscript{*} & -- & \DRSUnavailable & \TaskUnavailable & \TaskUnavailable & \TaskUnavailable \\
Cosmos 3 Edge\textsuperscript{\textdagger} \citep{nvidia2026cosmos3}
& \scorestd{32.7}{1.1}\textsuperscript{*} & -- & \DRSUnavailable & \scorestd{33.29}{0.78} & -- & \DRSUnavailable \\
CogVideoX-5B \citep{yang2025cogvideox}
& \scorestd{31.8}{1.5}\textsuperscript{*} & -- & \DRSUnavailable & \TaskUnavailable & \TaskUnavailable & \TaskUnavailable \\
Kandinsky-WM 1.0 \citep{anates2026physicsiqleaderboard}
& \scorestd{30.8}{0.9}\textsuperscript{*} & -- & \DRSUnavailable & \TaskUnavailable & \TaskUnavailable & \TaskUnavailable \\
Magi-1 (op) \citep{physicsiq2026githubleaderboard}
& \scorestd{30.2}{1.1}\textsuperscript{*} & -- & \DRSUnavailable & \scorestd{\underline{48.4}}{1.1}\textsuperscript{*} & -- & \DRSUnavailable \\
Sora 2 \citep{motamed2026physicsiqverified}
& \scorestd{26.5}{0.8}\textsuperscript{*} & -- & \DRSUnavailable & \TaskUnavailable & \TaskUnavailable & \TaskUnavailable \\
P-Video \citep{motamed2026physicsiqverified}
& \scorestd{25.3}{1.8}\textsuperscript{*} & -- & \DRSUnavailable & \TaskUnavailable & \TaskUnavailable & \TaskUnavailable \\
\bottomrule
\end{tabular*}
\par\smallskip
\parbox{\linewidth}{\fontsize{7}{8.2}\selectfont\raggedright
\textsuperscript{\textdagger} \rev{Expanded prompt: Claude Opus 5 converts the image-conditioned OP
description to \href{https://github.com/NVIDIA/cosmos-framework/blob/3845f4ca34dec8561c120a52c895c08781ec806b/cosmos_framework/inference/prompting_templates/reasoner/reasoner_i2v_json_schema.json}{NVIDIA's official I2V JSON format}; 50 steps, CFG 3.0, shift 4.75
(Appendix~\ref{app:prompt-expansion}).}
\textsuperscript{*} Published Physics-IQ Verified reference; unmarked estimates are our runs.}
\endgroup

\end{table}

%% file: figures/main/qualitative_physicsiq.tex
\begin{figure}[!t]
\centering
\caption{\textbf{\rev{Qualitative comparisons across five Physics-IQ categories.}}
\rev{Top: Cosmos 3 Nano 16B backbone; bottom: \methodfull{}, with matched
conditioning and eight equally spaced frames. \methodfull{} better realizes the
intended physical event:} ball separation after collision, liquid motion in the inverted
jar, persistence of the ball and its reflection, preservation of the paper under
rising smoke, and \rev{wrench attraction toward} the magnet. More examples: Figures~\ref{fig:qual-physicsiq-cosmos25}
to~\ref{fig:qual-physicsiq-more-b}.}
\label{fig:qual-physicsiq}
\includegraphics[width=0.9\linewidth]{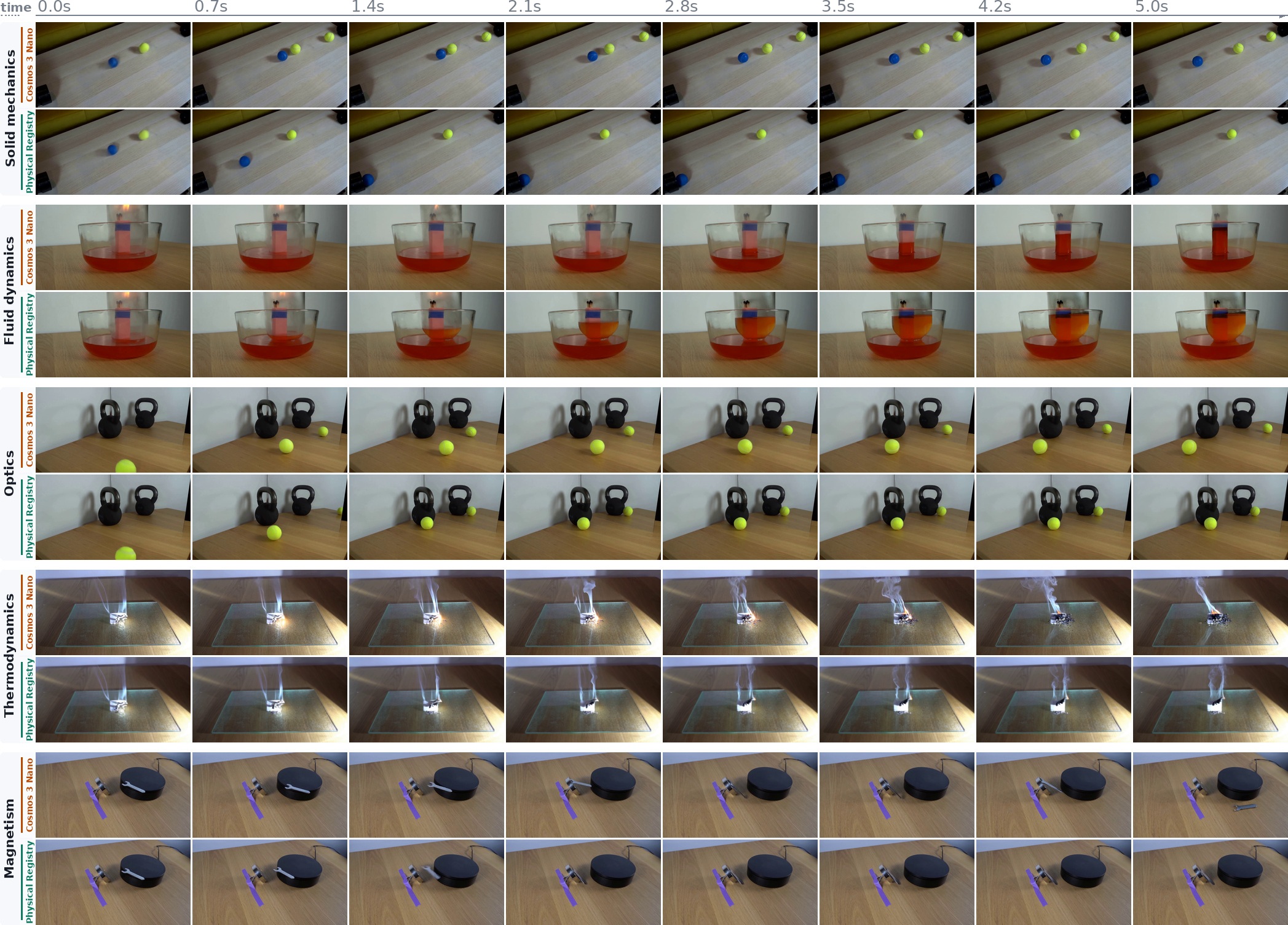}
\end{figure}

%% file: tables/main/worldmodelbench.tex
\begin{table}[!htb]
\centering
\caption{\textbf{WorldModelBench results.}
\postsub{Values in parentheses indicate the change from the corresponding backbone.}
\textsuperscript{*} \postsub{denotes} published leaderboard scores
\citep{li2025worldmodelbench}. Table~\ref{tab:worldmodelbench-leaderboard} \postsub{for the full results.}}
\label{tab:worldmodelbench}
\begingroup
\fontsize{8}{9.4}\selectfont
\renewcommand{\scorestd}[2]{#1\,\textpm\,#2}
\setlength{\tabcolsep}{2pt}
\renewcommand{\arraystretch}{1.04}
\begin{tabular*}{\linewidth}{@{\extracolsep{\fill}}lrrrr@{}}
\toprule
Model
& \shortstack[r]{Instruction\\following (0--3)}
& \shortstack[r]{Physical\\adherence (0--5)}
& \shortstack[r]{Common\\sense (0--2)}
& \shortstack[r]{Total\\(0--10)} \\
\midrule

Cosmos 3 Nano 16B \citep{nvidia2026cosmos3}
& \scorestd{2.20}{0.02}
& \scorestd{4.50}{0.08}
& \scorestd{1.43}{0.06}
& \scorestd{8.13}{0.13} \\
\prrow\quad + \methodfull{}
& \scorestd{2.20}{0.002}\scorechange{black}{--}
& \scorestd{\textbf{4.82}}{0.03}\scorechange{green!45!black}{+0.32}
& \scorestd{\underline{1.81}}{0.02}\scorechange{green!45!black}{+0.38}
& \scorestd{\underline{8.83}}{0.04}\scorechange{green!45!black}{+0.69} \\

\midrule

Cosmos v2.5 2B \citep{nvidia2025cosmospredict25}
& \scorestd{2.14}{0.03}
& \scorestd{4.43}{0.07}
& \scorestd{1.55}{0.05}
& \scorestd{8.12}{0.14} \\
\prrow\quad + \methodfull{}
& \scorestd{2.07}{0.01}\scorechange{red!70!black}{-0.07}
& \scorestd{4.54}{0.03}\scorechange{green!45!black}{+0.11}
& \scorestd{1.58}{0.04}\scorechange{green!45!black}{+0.03}
& \scorestd{8.19}{0.01}\scorechange{green!45!black}{+0.07} \\

\midrule

WAN 2.2 I2V-A14B \citep{wanteam2025wan22}
& \scorestd{2.23}{0.03}
& \scorestd{4.45}{0.04}
& \scorestd{1.60}{0.03}
& \scorestd{8.28}{0.02} \\
\prrow\quad + \methodfull{}
& \scorestd{2.30}{0.04}\scorechange{green!45!black}{+0.07}
& \scorestd{4.53}{0.04}\scorechange{green!45!black}{+0.07}
& \scorestd{1.60}{0.02}\scorechange{green!45!black}{+0.004}
& \scorestd{8.43}{0.05}\scorechange{green!45!black}{+0.15} \\

\midrule

WAN 2.2 TI2V-5B \citep{wanteam2025wan22}
& \scorestd{\textbf{2.33}}{0.01}
& \scorestd{4.35}{0.10}
& \scorestd{1.57}{0.04}
& \scorestd{8.24}{0.14} \\
\prrow\quad + \methodfull{}
& \scorestd{2.24}{0.05}\scorechange{red!70!black}{-0.09}
& \scorestd{4.50}{0.05}\scorechange{green!45!black}{+0.14}
& \scorestd{1.68}{0.06}\scorechange{green!45!black}{+0.11}
& \scorestd{8.41}{0.11}\scorechange{green!45!black}{+0.17} \\

\midrule

KLING \citep{kuaishou2024kling}
& \underline{2.32}\textsuperscript{*}
& \underline{4.81}\textsuperscript{*}
& \textbf{1.95}\textsuperscript{*}
& \textbf{9.10}\textsuperscript{*} \\

Luma \citep{lumalabs2024dreammachine}
& 1.98\textsuperscript{*}
& 4.49\textsuperscript{*}
& 1.76\textsuperscript{*}
& 8.24\textsuperscript{*} \\

OpenSoraPlan-T2V \citep{lin2024opensoraplan}
& 1.72\textsuperscript{*}
& 4.64\textsuperscript{*}
& 1.68\textsuperscript{*}
& 8.04\textsuperscript{*} \\

Mochi 1 \citep{genmo2024mochi}
& 2.06\textsuperscript{*}
& 4.39\textsuperscript{*}
& 1.45\textsuperscript{*}
& 7.91\textsuperscript{*} \\

CogVideoX-I2V \citep{yang2025cogvideox}
& 1.78\textsuperscript{*}
& 4.17\textsuperscript{*}
& 1.12\textsuperscript{*}
& 7.08\textsuperscript{*} \\

\bottomrule
\end{tabular*}
\endgroup

\end{table}

%% file: figures/main/selection_comparison.tex
\begin{figure}[!t]
\centering
\caption{\textbf{Selection comparison on Physics-IQ Verified.}
(left) I2V and (right) V2V.
Dashed caps mark Oracle scores and hatched bars denote published scores.}
\label{fig:selection-comparison}
\includegraphics[width=0.92\linewidth]{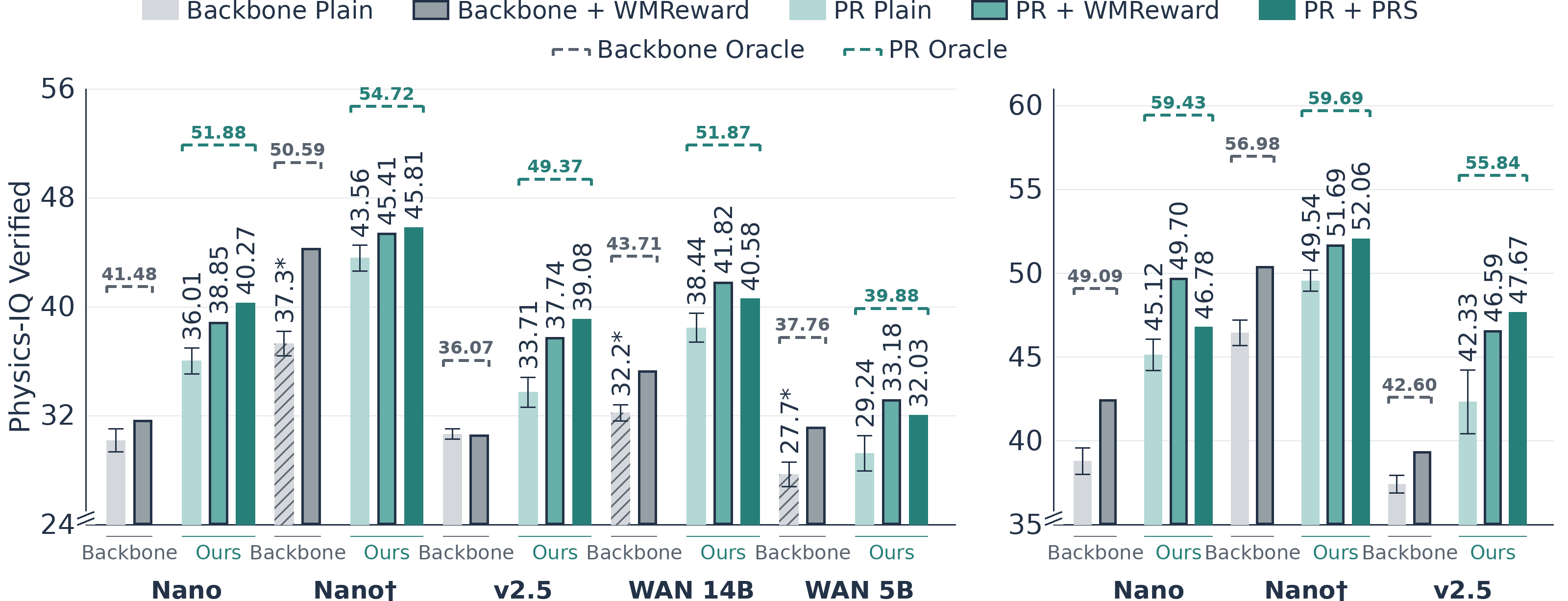}
\end{figure}

%% file: sections/main/05_ablation.tex
\subsection{Ablation of the Physical Registry}
\label{sec:registry-ablation}

\rev{\postsub{We examine whether} the \postsub{gains} of \postsub{\method{} come from the} forecast \postsub{itself,
additional registry capacity, or} predictive supervision\postsub{.}
Table~\ref{tab:registry-ablation}(a) \postsub{tests} the forecast \postsub{content, while}
Table~\ref{tab:registry-ablation}(b) \postsub{isolates the contributions of registry
capacity, forecast initialization,} and \postsub{predictive supervision.}}

\input{tables/main/registry_ablation}

\paragraph{Generation depends on the forecast content.}
\leavevmode\rev{
In Table~\ref{tab:registry-ablation}(a), \postsub{removing} the projected forecast drops
Nano from \(36.01\) to \(16.78\) on I2V, whereas \postsub{moderate} Gaussian perturbation
(\(\sigma=0.5\)) \postsub{has} only \postsub{a small effect;} WAN 14B shows the same trend.
The registry \postsub{therefore depends strongly on} the forecast \postsub{content while remaining}
robust to moderate perturbation. In \postsub{the} motion-transfer experiment
(Appendix~\ref{app:dynamics-transfer}), replacing the forecast with JEPA
features \postsub{from a source video shifts the generated motion toward that source
while keeping} the initial frame, prompt, and noise fixed.}

\paragraph{\rev{Learnable registers alone are not sufficient.}}
\leavevmode\rev{
In Table~\ref{tab:registry-ablation}(b), input-independent \postsub{learnable registers
underperform} vanilla fine-tuning, \postsub{while} forecast initialization improves over
\postsub{them} and predictive supervision improves further.
The gains therefore do not come from additional \postsub{registry} capacity alone:
\postsub{forecast content in the registry helps, and predictive supervision further
strengthens its role in representing future dynamics.}}

\subsection{Registry Attention and Activation}
\label{app:primary-registry-responses}
\label{sec:analysis}

\ifdefined\omitfigurethree
\begin{center}
\includegraphics[width=\linewidth]{figures/fig_attention_activation_placeholder.png}
\end{center}
\else
\input{figures/main/registry_hidvar}
\fi

\rev{
Figure~\ref{fig:registry-attention} \postsub{highlights how \method{} responds differently
from text cross-attention and video-token attention} during balloon
inflation. \postsub{While these attention maps broadly follow} the balloon \postsub{and its}
interaction region, attention to the aligned registry slots and registry
activation concentrate \postsub{near} the expanding boundary, \postsub{where the dominant physical
change occurs. This suggests} that \postsub{the} registry \postsub{provides a more localized representation of
regions where the intended scene dynamics unfold.}}

%% file: tables/main/registry_ablation.tex
\begingroup
\renewcommand{\scorestd}[2]{#1\,\textpm\,#2}
\newsavebox{\registryContentBox}
\newsavebox{\registryVariantsBox}
\newlength{\registryPanelWidth}
\newlength{\registryScoreWidth}
\newlength{\registryContentPadding}
\newlength{\registryVariantsPadding}
\setlength{\registryPanelWidth}{0.49\linewidth}
\newcommand{\RegistryNotApplicable}{%
  \begingroup
  \setlength{\fboxsep}{0pt}%
  \colorbox{black!12}{\makebox[\registryScoreWidth]{\strut}}%
  \endgroup}
\newcommand{\registrytableformat}{%
  \scriptsize
  \settowidth{\registryScoreWidth}{\textbf{00.0}}%
  \setlength{\tabcolsep}{3pt}%
  \renewcommand{\arraystretch}{1.12}}
\newcommand{\registrycontenttable}[1]{%
\registrytableformat
\begin{tabular*}{\registryPanelWidth}{@{\extracolsep{\fill}}lrr@{}}
\toprule
Registry content & I2V & V2V \\
\midrule
\shortstack[l]{\textbf{\methodfull{}}\\\textbf{(Cosmos 3 Nano 16B)}} & \scorestd{\textbf{36.01}}{0.95} & \scorestd{\textbf{45.12}}{0.94} \\
\noalign{\vskip#1}
Zero mask & \scorestd{16.78}{0.01} & \scorestd{24.15}{0.32} \\
\noalign{\vskip#1}
Perturbed mask (Noise \(\sigma=0.5\)) & \scorestd{35.84}{0.11} & \scorestd{44.35}{0.75} \\
\noalign{\vskip#1}
\midrule
\shortstack[l]{\textbf{\methodfull{}}\\\textbf{(WAN 2.2 I2V-A14B)}} & \scorestd{\textbf{38.44}}{1.06} & \RegistryNotApplicable \\
\noalign{\vskip#1}
Zero mask & \scorestd{35.88}{0.22} & \RegistryNotApplicable \\
\noalign{\vskip#1}
Perturbed mask (Noise \(\sigma=0.5\)) & \scorestd{36.40}{0.39} & \RegistryNotApplicable \\
\noalign{\vskip#1}
\bottomrule
\end{tabular*}}
\newcommand{\registryvarianttable}[1]{%
\registrytableformat
\begin{tabular*}{\registryPanelWidth}{@{\extracolsep{\fill}}lrrrr@{}}
\toprule
Variant & I2V & \(\Delta\) & V2V & \(\Delta\) \\
\midrule
Vanilla fine-tuning & \scorestd{41.3}{2.1} & -- & \scorestd{46.4}{1.0} & -- \\
\noalign{\vskip#1}
\shortstack[l]{ViT Needs Registers\\\citep{darcet2024registers}} & \scorestd{39.0}{1.1} & \(-2.3\) & \scorestd{44.6}{0.9} & \(-1.8\) \\
\noalign{\vskip#1}
Joint generation & \scorestd{45.1}{1.8} & +3.8 & \scorestd{48.0}{0.4} & +1.6 \\
\noalign{\vskip#1}
Registry only & \scorestd{45.0}{1.3} & +3.7 & \scorestd{50.3}{0.9} & +3.9 \\
\noalign{\vskip#1}
\midrule
\prrow[\registryPanelWidth]\textbf{\methodfull{}} & \scorestd{\textbf{51.2}}{1.5} & \textbf{+9.9} & \scorestd{\textbf{55.1}}{1.1} & \textbf{+8.7} \\
\noalign{\vskip#1}
\bottomrule
\end{tabular*}}
\begin{table}[!htb]
\centering
\caption{\textbf{Registry ablations on Physics-IQ Verified.}
(a) Forecast-content controls on Cosmos 3 Nano 16B and WAN 2.2 I2V-A14B.
(b) Comparison of \rev{learnable registers}, joint generation, forecast-initialized
registry states, and full \methodfull{} on Cosmos v2.5 2B.}
\label{tab:registry-ablation}

\sbox{\registryContentBox}{\registrycontenttable{0pt}}
\sbox{\registryVariantsBox}{\registryvarianttable{0pt}}
\ifdim\dimexpr\ht\registryContentBox+\dp\registryContentBox\relax
  >\dimexpr\ht\registryVariantsBox+\dp\registryVariantsBox\relax
  \setlength{\registryVariantsPadding}{%
    \dimexpr(\ht\registryContentBox+\dp\registryContentBox
      -\ht\registryVariantsBox-\dp\registryVariantsBox)/5\relax}
\else
  \setlength{\registryContentPadding}{%
    \dimexpr(\ht\registryVariantsBox+\dp\registryVariantsBox
      -\ht\registryContentBox-\dp\registryContentBox)/6\relax}
\fi
\sbox{\registryContentBox}{\registrycontenttable{\registryContentPadding}}
\sbox{\registryVariantsBox}{\registryvarianttable{\registryVariantsPadding}}
\begin{minipage}[t]{\registryPanelWidth}
\centering
\vspace{0pt}
\usebox{\registryContentBox}
\end{minipage}
\hfill
\begin{minipage}[t]{\registryPanelWidth}
\centering
\vspace{0pt}
\usebox{\registryVariantsBox}
\end{minipage}

\end{table}
\endgroup

%% file: figures/main/registry_hidvar.tex
\begin{figure}[!t]
\centering
\caption{\textbf{Registry attention and activation during balloon inflation.}
\rev{Foreground queries are video tokens inside the Physics-IQ motion mask of the
generated video. Labels give query\(\rightarrow\)key pairs (e.g., q3\(\rightarrow\)r1: queries
in latent frame 3 attending to its time-aligned registry slot r1); r1 alone marks
that slot's activation. Protocol: Appendix~\ref{app:implementation}; more examples:
Appendices~\ref{app:registry-visualization} and~\ref{app:additional-attention}.}}
\label{fig:registry-attention}
\includegraphics[width=0.9\linewidth]{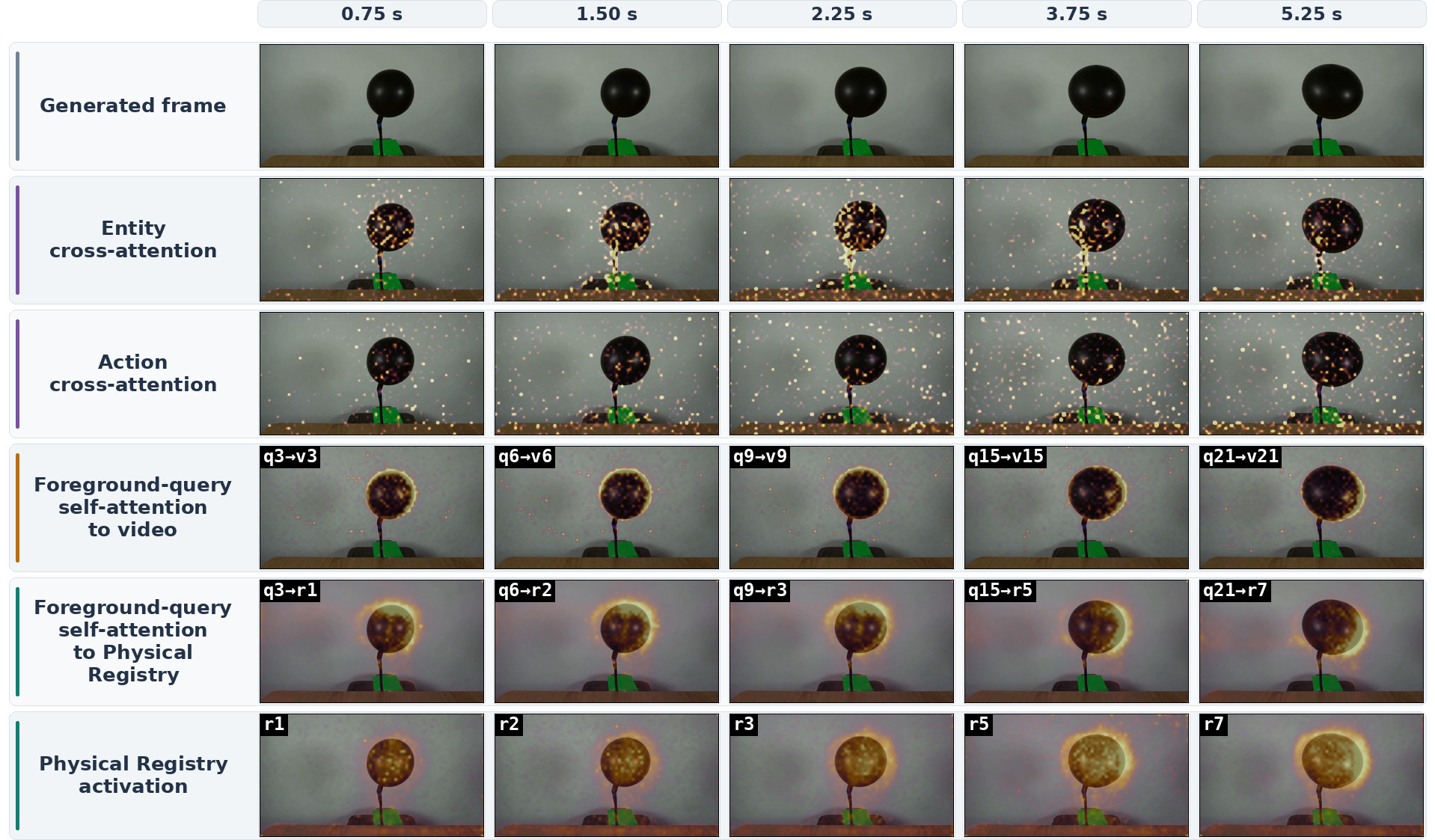}
\end{figure}

%% file: sections/main/06_discussion_conclusion.tex
\section{Discussion and Conclusion}
\label{sec:conclusion}

\method{} gives predictive representations \postsub{an explicit,} temporally grounded
state within the video backbone \postsub{rather than leaving them entangled with the
video representation. Our results suggest} that this dedicated state \postsub{can be
explicitly supervised toward future dynamics} while \postsub{remaining readable} during
\postsub{inference. More broadly,} explicit predictive \postsub{states provide} a useful
\textbf{interface for \postsub{representing} and reading physically plausible futures}
\postsub{in world models.}

%% file: sections/appendix/c_ethics_llm_disclosure.tex
\section*{AI Use Statement}
\label{sec:ai-use}

\paragraph{Writing and language polishing.}
Generative AI tools were used to refine wording, shorten figure captions, and
improve consistency in terminology and presentation.

\paragraph{Retrieval and discovery.}
Generative AI tools were used to assist in identifying relevant prior methods
and related literature. Papers considered or cited in this work were
subsequently checked against their original sources by the authors.

\paragraph{Research execution.}
Generative AI tools were used for routine operational assistance during
experimental execution, including preparing and submitting experiment jobs to
cloud computing environments, formatting execution commands and configuration
files, and assisting with minor scripting or runtime issues.

\paragraph{Manuscript drafting.}
Generative AI tools were used to help convert author-provided technical notes
and descriptions into preliminary prose, reorganize existing text for clarity,
and suggest alternative wording. All AI-assisted material was reviewed by the
authors.

%% file: sections/appendix/a_technical_appendix.tex
\section{Implementation and Experimental Details}
\label{app:implementation}
\label{app:training-data}
\label{app:setup-detail}
\label{sec:projection-architecture}
\label{app:prompt-expansion}

\subsection{\postsub{Registry Architecture and Supervision}}

\paragraph{\postsub{Registry representation.}}
\leavevmode\rev{
The frozen JEPA encoder is V-JEPA 2.1
\citep{murlabadia2026vjepa21} ViT-g at \(384\) px, loaded from the pretrained
target-encoder branch. The registry uses eight slots, one context feature and
seven predictor rollout steps, each with \(196\) spatial tokens of \(5{,}632\)
channels. Full-video JEPA features provide the alignment targets, mean-pooled
over the \(196\) spatial tokens. Video and registry tokens share dense
self-attention, while text remains a separate cross-attention condition.}

\paragraph{\postsub{Predictive supervision.}}
We use \(\lambda_{\mathrm{align}}=0.5\) in
Equation~\eqref{eq:total-loss}, following the default REPA projection-loss
weight \citep{yu2025repa}. Registry only retains the forecast-initialized
registry and removes the alignment term.
\leavevmode\rev{
Following prior representation-alignment work
\citep{yu2025repa,zhang2025videorepa}, we supervise an early-to-middle
generation block. The block used for each backbone is listed below. For WAN I2V-A14B, the block count and alignment index apply to each noise
expert.}

\begin{center}
\footnotesize
\begin{tabular}{lrr}
\toprule
Mean-pooled implementation & Generation blocks & Alignment block index \\
\midrule
Cosmos 3 Nano 16B & 36 & 10 \\
Cosmos v2.5 2B & 28 & 14 \\
WAN 2.2 I2V-A14B & 40 & 11 \\
WAN 2.2 TI2V-5B & 30 & 15 \\
\bottomrule
\end{tabular}
\end{center}

\input{sections/appendix/g_projection_layer}

\subsection{\postsub{Training and Data}}

\input{sections/appendix/e_training_dataset}

\paragraph{Fine-tuning settings.}
Table~\ref{tab:training-configuration} reports model sizes, effective batch,
and GPU resources for \method{} fine-tuning on the Enlarge dataset.
The matched-budget WAN control in
Appendix~\ref{app:wan5b-matched-budget} uses 8,000 iterations and effective
batch 16.

\input{tables/appendix/training_configuration}

\subsection{\postsub{Inference and Sampling}}

\paragraph{\rev{Sampling settings.}}
\leavevmode\rev{
Apart from the expanded-prompt setting
(\textsuperscript{\textdagger}, Table~\ref{tab:main-results}), guidance,
shift, sampler, and step count follow each model's official inference defaults
(Table~\ref{tab:inference-configuration}). Each backbone and its \method{}
counterpart use the same settings. Following the Physics-IQ protocol, the first
five seconds after conditioning are evaluated.}

\input{tables/appendix/inference_configuration}

\paragraph{Sampling cost.}
\leavevmode\rev{
Table~\ref{tab:computational-cost} reports the median per-step time over
three Physics-IQ scenarios with three repeats; for Cosmos 3 Nano 16B, whose
sampler caches activations, it is computed from total sampling time. The registry increases per-step cost
to \(1.67\times\) on Cosmos v2.5 2B and \(1.55\times\) on Cosmos 3 Nano 16B.
Times are compared only within each backbone.}

\input{tables/appendix/computational_cost}

\input{sections/appendix/j_prompt_expansion}

\subsection{\postsub{Evaluation and Visualization Details}}

\input{sections/appendix/e_experimental_setup_detail}

\input{figures/appendix/jepa_cross_view}

\paragraph{\rev{Visualization protocol.}}
\leavevmode\rev{
Registry activation is the hidden-state variance over denoising steps 0--10 at
each aligned registry slot and spatial position, averaged over feature channels.

For Figure~\ref{fig:registry-attention}, foreground queries use the Physics-IQ
generated-foreground mask \citep{motamed2026physicsiq}, downsampled by area
averaging to the \(44\times80\) video-latent grid. Attention maps average native
QK-softmax responses over steps \(\{0,5,11,16,22,28,34\}\).

For Appendix~\ref{app:additional-attention}, text-attention maps aggregate the
corresponding lexical token groups and normalize by valid-text attention at each
spatial location. Text and video self-attention use steps 0--10, while registry
self-attention uses steps 11--22.}

%% file: sections/appendix/g_projection_layer.tex
\paragraph{Registry projection architecture.}

The adapter \(G_\phi\) of Equation~\eqref{eq:registry-projection} combines established
video-modeling components: local spatio-temporal convolutions
\citep{tran2015c3d} and per-position temporal attention, a factorization used
in video diffusion architectures \citep{ho2022videodiffusion}.
The convolutions mix neighboring locations and adjacent forecast features\rev{. Temporal} attention then integrates the slot sequence at each spatial position.
Spatial resizing and channel projection match the backbone's latent geometry
while retaining the forecast's temporal indexing.
Figure~\ref{fig:projection-layer} shows this \rev{adapter}.

\input{figures/appendix/projection_layer}

\paragraph{Projection input and output.}
For the Cosmos v2.5 2B configuration, the input is the forecast
\(J^{\mathrm{cond}}\), a sequence of
\(T_r{=}8\) slots, each a \(14\times14\) grid of \(5632\)-dimensional features
(the encoder embedding width of \(1408\) concatenated over four blocks). The
output is \(R_t^{(0)}\) in the video-latent format of the backbone:
\(16\) channels over \(8\) registry slots at the \(88\times160\) latent
resolution.

\paragraph{Projection stages.}
\leavevmode\rev{Figure~\ref{fig:projection-layer} gives the operations and shapes of each stage.
Upsampling acts on the two spatial axes only, and the temporal Transformer is
applied independently at each spatial position with the eight registry slots as
its sequence, so spatial mixing comes from the convolutions and, after
concatenation, from the backbone's self-attention. The adapter is initialized
randomly and trained jointly with the backbone.}

\paragraph{Projection placement.}
The module emits the backbone's native latent channel count.
Its output is normalized with the
tokenizer's own per-frame statistics for the registry slots, which places the
registry on the scale of the video latent\rev{. The normalized output is then
concatenated} along the temporal axis to give \(T_v{+}T_r=24+8=32\) latent
frames. Registry slot \(i\) receives the video-latent RoPE coordinate \(3i\),
the temporal alignment of Equation~\eqref{eq:registry-rope}.
With \(2\times2\) spatial patchification, each registry slot contains
\(44\times80\) tokens, giving \(8\times44\times80=28{,}160\) registry
tokens in this configuration.

\paragraph{Registry readout modules.}
The readout operator \(q_\psi\) in
Equations~\eqref{eq:registry-readout} and~\eqref{eq:selector-main} includes spatial mean-pooling
within each registry slot, before its projection.
Registry slots remain separate until loss or score aggregation.
For training, \(q_\psi\) \rev{uses} a three-layer MLP with output
widths \(2048\), \(2048\), and \(5632\), and \textsc{SiLU} activations
(\(19.9\)M parameters). It maps the spatially averaged registry hidden state
to the encoder feature dimension.

The slot-level target \(\overline Z_s^{\mathrm{tgt}}\) and reference
\(\overline J_s^{\mathrm{cond}}\) are spatial means of the frozen encoder's
target grid and the unprojected shared-forecast grid, respectively.
These reductions preserve the temporal indexing of their source features.
The alignment loss trains a single readout, which the selection policy reuses
frozen. No additional module is trained for selection.
The conditioning adapter \(G_\phi\) operates in the reverse direction,
mapping the forecast into the video-latent geometry.

%% file: figures/appendix/projection_layer.tex
\begingroup
\definecolor{drForecast}{HTML}{78549C}
\definecolor{drForecastFill}{HTML}{F0EAF7}
\definecolor{drRegistry}{HTML}{177C72}
\definecolor{drRegistryFill}{HTML}{E3F4EF}
\definecolor{drVideo}{HTML}{B76D19}
\definecolor{drVideoFill}{HTML}{FFF0DA}
\begin{figure}[!htbp]
\centering
\caption{\textbf{The registry projection layer \(G_\phi\).}
The three bands show spatial lifting with 3D convolutions
\citep{tran2015c3d}, per-position temporal attention
\citep{ho2022videodiffusion} and channel projection, and backbone input
assembly. Shapes are for the Cosmos v2.5 2B configuration.}
\label{fig:projection-layer}
\resizebox{\linewidth}{!}{%
\begin{tikzpicture}[
  x=1cm,y=1cm,font=\scriptsize,
  band/.style={draw=black!18,fill=black!1,rounded corners=3pt,line width=.45pt},
  box/.style={draw=black!35,fill=white,rounded corners=2pt,align=center,
    inner sep=2pt,minimum height=8.8mm},
  forecaststate/.style={box,draw=drForecast,fill=drForecastFill},
  reg/.style={box,draw=drRegistry,fill=drRegistryFill},
  videostate/.style={box,draw=drVideo,fill=drVideoFill},
  dims/.style={font=\tiny,text=black!60,inner sep=1pt},
  badge/.style={circle,fill=drRegistry,text=white,inner sep=1.7pt,font=\tiny\bfseries},
  ar/.style={-{Latex[length=1.5mm,width=1.1mm]},line width=.55pt},
  farrow/.style={ar,draw=drForecast},
  rarrow/.style={ar,draw=drRegistry},
  varrow/.style={ar,draw=drVideo},
]
\path[band] (0,4.30) rectangle (13.6,6.10);
\path[band] (0,2.15) rectangle (13.6,4.00);
\path[band] (0,.12) rectangle (13.6,1.85);
\foreach \stage/\y/\heading in {
  1/5.79/{Spatial lifting and refinement},
  2/3.69/{Temporal mixing and channel head},
  3/1.57/{Backbone input assembly}}{
  \node[badge] at (.40,\y) {\stage};
  \node[anchor=west,font=\scriptsize\bfseries] at (.70,\y) {\heading};
}
\node[dims,anchor=east] at (13.20,5.79) {8 registry slots throughout};
\node[dims,anchor=east] at (13.20,3.69) {independently at each spatial position};

\node[forecaststate,minimum width=15mm] (jc) at (1.10,5.12)
  {\(J^{\mathrm{cond}}\)\\forecast};
\node[box,minimum width=17mm] (rs) at (3.30,5.12)
  {\textbf{Reshape}\\\(14\times14\) grid};
\node[box,minimum width=18mm] (c1) at (5.50,5.12)
  {\textbf{Conv3D} \(3^3\)\\\(5632\to96\)};
\node[box,minimum width=24mm] (up) at (8.30,5.12)
  {\textbf{Trilinear resize}\\\(14\times14\to88\times160\)\\
    {\tiny spatial only; \(T\) fixed}};
\node[box,minimum width=30mm] (rc) at (11.55,5.12)
  {\textbf{Residual block}\\
    \(2\times(\mathrm{GN}_{32}\!-\!\mathrm{SiLU}\!-\!3^3)\)\\
    {\tiny one skip around the pair}};
\foreach \a/\b in {jc/rs,rs/c1}{\draw[farrow] (\a.east) -- (\b.west);}
\foreach \a/\b in {c1/up,up/rc}{\draw[rarrow] (\a.east) -- (\b.west);}
\node[dims] at (1.10,4.51) {\((B,8,196,5632)\)};
\node[dims] at (5.50,4.51) {\((B,96,8,14,14)\)};
\node[dims] at (11.55,4.51) {\((B,96,8,88,160)\)};

\node[box,minimum width=23mm] (seq) at (1.65,2.99)
  {\textbf{Slot sequence}\\\(8\times96\)\\\(+E_{\mathrm{pos}}\)};
\node[reg,minimum width=44.5mm] (tt) at (6.05,2.99)
  {\textbf{Temporal Transformer} \(\times8\)\\
    \(d=96\), \(8\) heads, FFN \(384\)\\
    {\tiny post-norm; ReLU; dropout \(0.1\)}};
\node[box,minimum width=19mm] (cmu) at (9.70,2.99)
  {\textbf{Conv3D} \(1^3\)\\\(96\to16\)};
\node[reg,minimum width=18.5mm] (out) at (12.00,2.99)
  {\(R_t^{(0)}\)\\registry init};
\draw[rarrow] (rc.east) -- (13.40,5.12) -- (13.40,4.15)
  -- (.16,4.15) |- (seq.west);
\foreach \a/\b in {seq/tt,tt/cmu,cmu/out}{\draw[rarrow] (\a.east) -- (\b.west);}
\node[dims] at (1.65,2.34) {learned positions: \((1,8,96)\)};
\node[dims] at (12.00,2.34) {\((B,16,8,88,160)\)};

\node[box,minimum width=33.5mm,minimum height=7mm] (wn) at (2.25,.83)
  {\textbf{Tokenizer normalization}\\
    {\tiny \(\mu,\sigma\) of slots \(T_v{:}T_v{+}T_r\)}};
\node[box,minimum width=32mm,minimum height=7mm] (cat) at (6.55,.83)
  {\textbf{Temporal concat}\\\([X_t;R_t^{(0)}]\)};
\node[box,minimum width=36mm,minimum height=7mm] (rope) at (11.20,.83)
  {\textbf{Temporal RoPE}\\registry slot \(i\mapsto3i\)};
\node[videostate,font=\tiny,inner sep=1pt,minimum width=22mm,minimum height=3mm]
  (video) at (6.55,1.58) {\(X_t\): noisy video};
\draw[rarrow] (out.east) -- (13.40,2.99) -- (13.40,2.00)
  -- (.16,2.00) |- (wn.west);
\foreach \a/\b in {wn/cat,cat/rope}{\draw[rarrow] (\a.east) -- (\b.west);}
\draw[varrow] (video.south) -- (cat.north);
\node[dims] at (6.55,.26) {\(24+8=32\) latent frames};
\node[dims] at (11.20,.26) {to the shared video DiT};

\end{tikzpicture}}
\end{figure}
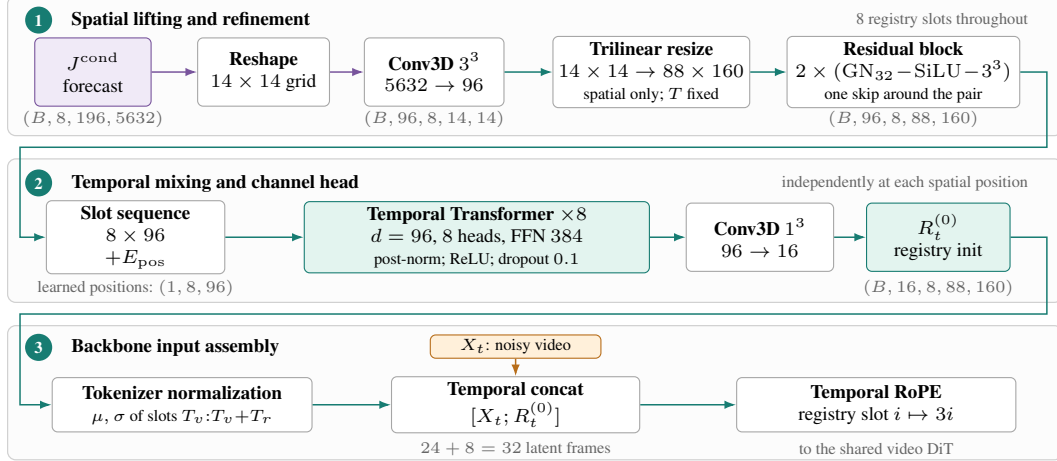
\endgroup

%% file: sections/appendix/e_training_dataset.tex
\paragraph{The Enlarge dataset.}
The Enlarge dataset combines physics-focused data with broad
visual coverage. It includes physics-oriented video, rigid-body simulation,
general web video, and embodied robot simulation
\citep{wang2025wisa,wang2023internvid,nvidia2026physxsimdataset,
nvidia2026robotsimdataset}. We fine-tune \method{} using only its RGB videos and
corresponding captions, without additional physical
annotations or simulator states. \rev{The Enlarge dataset} excludes Physics-IQ videos.
Table~\ref{tab:training-data} and
Figure~\ref{fig:training-data-distribution} report the exact number of clips and
video hours used from each source.

\input{tables/main/training_data}

%% file: tables/main/training_data.tex
\begingroup
\hyphenpenalty=10000\relax
\exhyphenpenalty=10000\relax
\newsavebox{\trainingDataTableBox}
\newsavebox{\trainingDataChartBox}
\newlength{\trainingDataWidth}
\newlength{\trainingDataRowPadding}
\setlength{\trainingDataWidth}{0.49\linewidth}
\newcommand{\trainingdatatable}[1]{%
\footnotesize%
\setlength{\tabcolsep}{3.1pt}%
\renewcommand{\arraystretch}{1.1}%
\begin{tabular*}{\trainingDataWidth}{@{\extracolsep{\fill}}lrrr@{}}
\toprule
Dataset & Clips & Mean (s) & Hours \\
\midrule
\noalign{\vskip#1}
WISA & 74,370 & 7.1 & 146.7 \\
\noalign{\vskip\dimexpr#1+#1\relax}
InternVid & 8,115 & 7.5 & 16.9 \\
\noalign{\vskip\dimexpr#1+#1\relax}
SDG-PhysXSim & 101,422 & 5.0 & 140.9 \\
\noalign{\vskip\dimexpr#1+#1\relax}
SDG-RobotSim & 28,193 & 9.0 & 70.6 \\
\noalign{\vskip#1}
\midrule
\textbf{Total} & \textbf{212,100} & \textbf{6.4} & \textbf{375.2} \\
\bottomrule
\end{tabular*}}
\begin{figure}[!htbp]
\centering
\sbox{\trainingDataChartBox}{%
\includegraphics[width=\trainingDataWidth]{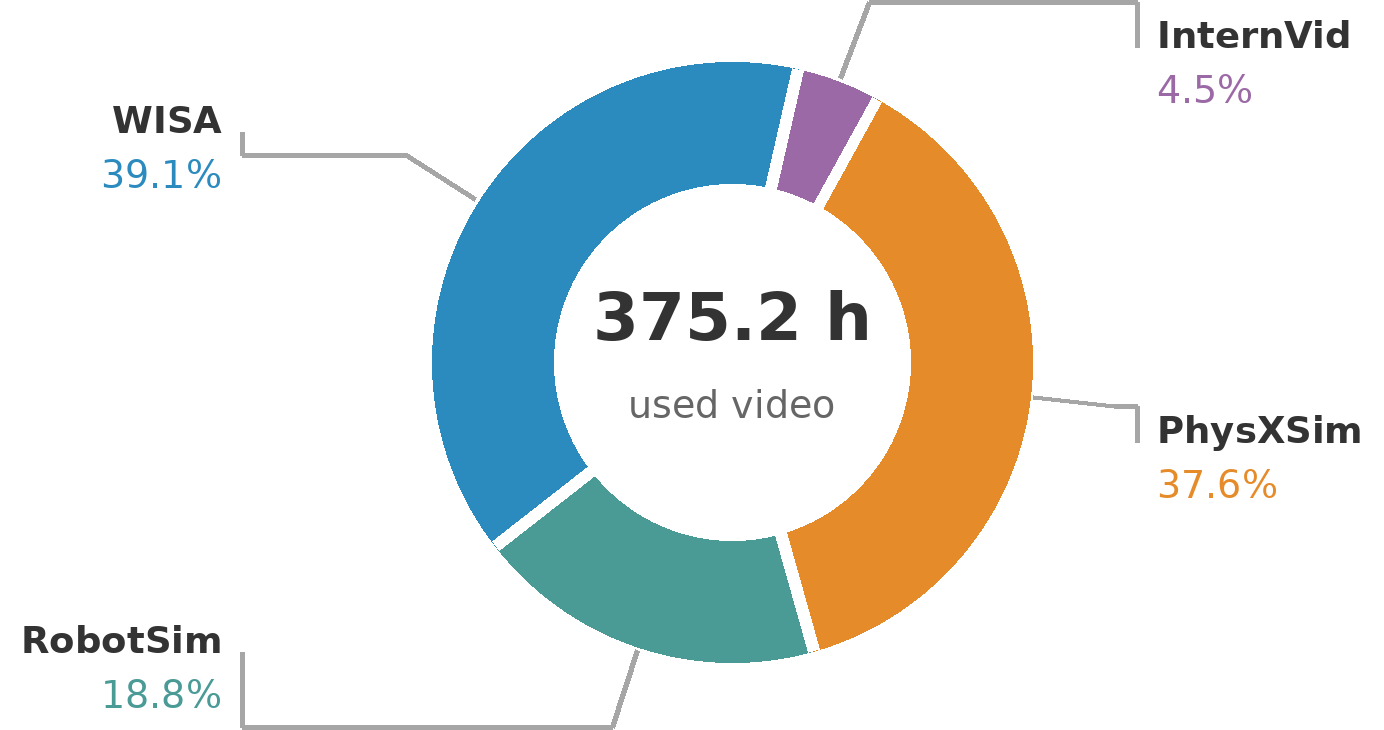}}
\sbox{\trainingDataTableBox}{\trainingdatatable{0pt}}
\setlength{\trainingDataRowPadding}{%
  \dimexpr(\ht\trainingDataChartBox+\dp\trainingDataChartBox
    -\ht\trainingDataTableBox-\dp\trainingDataTableBox)/8\relax}
\ifdim\trainingDataRowPadding<0pt
  \PackageError{training-data}{Composition chart is shorter than the training table}{Increase the shared panel width or reduce table padding.}
\fi
\sbox{\trainingDataTableBox}{\trainingdatatable{\trainingDataRowPadding}}
\begin{minipage}[t]{\trainingDataWidth}
\vspace{0pt}
\centering
\captionof{table}{\textbf{The Enlarge dataset.} Exact clips used by the evaluated \methodfull{} models.}
\label{tab:training-data}
\raisebox{\dp\trainingDataTableBox}{\usebox{\trainingDataTableBox}}
\end{minipage}
\hfill
\begin{minipage}[t]{\trainingDataWidth}
\vspace{0pt}
\centering
\captionof{figure}{\textbf{Used-hour composition.} Projected hours for the exact used clips.}
\label{fig:training-data-distribution}
\raisebox{\dp\trainingDataChartBox}{\usebox{\trainingDataChartBox}}
\end{minipage}
\end{figure}
\endgroup

%% file: tables/appendix/training_configuration.tex
\begin{table}[!htbp]
\centering
\caption{\textbf{Model sizes, fine-tuning resources, and optimization settings.}
Registry modules denote the added projection and readout parameters
(e.g., 16.0M + 19.9M for Cosmos v2.5 2B).}
\label{tab:training-configuration}
\footnotesize
\setlength{\tabcolsep}{3pt}
\renewcommand{\arraystretch}{1.0}

\begin{tabular*}{\linewidth}{@{\extracolsep{\fill}}lrrrl@{}}
\toprule
Model & Pretrained & Registry modules & Batch/update & GPUs \\
\midrule
Cosmos 3 Nano 16B & 16B & 72.63M & 16 & \(8\times\) H100 \\
Cosmos v2.5 2B & 2B & 35.93M & 16 & \(8\times\) A100 40 GB \\
WAN 2.2 I2V-A14B & \(2\times14\)B & 129.39M & 16 & \(8\times\) H100 \\
WAN 2.2 TI2V-5B & 5B & 105.02M & 16 & \(8\times\) A100 40 GB \\
\bottomrule
\end{tabular*}

\par\smallskip

\begin{tabular*}{\linewidth}{@{\extracolsep{\fill}}lrlrc@{}}
\toprule
Model & Learning rate & Schedule & Weight decay & AdamW \((\beta_1,\beta_2)\) \\
\midrule
Cosmos 3 Nano 16B
& \(1\times10^{-5}\)
& constant
& 0
& \((0.9,0.95)\) \\

Cosmos v2.5 2B
& \(3\times10^{-5}\)
& 100-step warm-up, then decay
& \(1\times10^{-3}\)
& \((0.9,0.999)\) \\

WAN 2.2 I2V-A14B
& \(1\times10^{-5}\)
& constant
& \(1\times10^{-2}\)
& \((0.9,0.95)\) \\

WAN 2.2 TI2V-5B
& \(1\times10^{-5}\)
& constant
& \(1\times10^{-2}\)
& \((0.9,0.999)\) \\
\bottomrule
\end{tabular*}

\end{table}

%% file: tables/appendix/inference_configuration.tex
\begin{table}[!htbp]
\centering
\caption{\rev{\textbf{Sampling settings.} Steps are denoising steps and guidance is the
classifier-free guidance scale (WAN I2V-A14B: high-/low-noise experts).
Generated frames exclude conditioning frames (Cosmos v2.5: I2V / V2V).
\textsuperscript{\textdagger} Expanded-prompt setting of Table~\ref{tab:main-results}.}}
\label{tab:inference-configuration}
\footnotesize
\setlength{\tabcolsep}{4pt}
\renewcommand{\arraystretch}{1.0}
\rev{%
\begin{tabular*}{\linewidth}{@{\extracolsep{\fill}}lrrrlrr@{}}
\toprule
Model & Steps & Guidance & Shift & Sampler & Resolution & Gen.\ frames @ fps \\
\midrule
Cosmos 3 Nano 16B & 35 & 6.0 & 10.0 & UniPC & \(1280\times720\) & 120 @ 24 \\
Cosmos 3 Nano 16B\textsuperscript{\textdagger} & 50 & 3.0 & 4.75 & UniPC & \(1280\times720\) & 120 @ 24 \\
Cosmos v2.5 2B & 35 & 7.0 & 5.0 & UniPC & \(1280\times704\) & 92 / 80 @ 16 \\
WAN 2.2 I2V-A14B & 40 & 3.5 / 3.5 & 5.0 & UniPC & \(1280\times720\) & 80 @ 16 \\
WAN 2.2 TI2V-5B & 50 & 5.0 & 5.0 & UniPC & \(1280\times704\) & 120 @ 24 \\
\bottomrule
\end{tabular*}}
\end{table}

%% file: tables/appendix/computational_cost.tex
\begin{table}[!htbp]
\centering
\caption{\textbf{Sampling cost of the registry.}
Both arms of a backbone are measured in one session on a single H100 80\,GB.
Times are per sampler step; peak memory is the generation peak above the weights
resident before sampling. Two sequence lengths are given for Cosmos 3 Nano 16B
because its two classifier-free guidance branches carry different text lengths.}
\label{tab:computational-cost}
\footnotesize
\setlength{\tabcolsep}{4pt}
\renewcommand{\arraystretch}{1.0}
\begin{tabular*}{\linewidth}{@{\extracolsep{\fill}}llrrr@{}}
\toprule
Model & Arm & Sequence (tokens) & s / sampler step & Peak (GiB) \\
\midrule
Cosmos v2.5 2B & Backbone & 84,480 & 6.514 & 6.20 \\
 & \(+\)\methodfull{} & 112,640 & 10.909 & 7.00 \\
\addlinespace[2pt]
Cosmos 3 Nano 16B & Backbone & 28,607 / 31,364 & 1.384 & 11.44 \\
 & \(+\)\methodfull{} & 35,967 / 38,724 & 2.148 & 11.94 \\
\bottomrule
\end{tabular*}
\end{table}

%% file: sections/appendix/j_prompt_expansion.tex
\begingroup
\newcommand{\promptblock}[2]{%
  \par\smallskip\noindent
  \fcolorbox{black!18}{black!3}{%
    \begin{minipage}{\dimexpr\linewidth-2\fboxsep-2\fboxrule\relax}
    {\normalfont\footnotesize\bfseries #1\par}
    \smallskip\footnotesize\ttfamily\raggedright
    #2
    \end{minipage}}%
  \par}
\paragraph{Prompt expansion.}
For the expanded-prompt Nano setting in
Tables~\ref{tab:main-results} and~\ref{tab:selection-results},
Claude Opus 5 (model id \texttt{claude-opus-5}) expands the
Physics-IQ original description (OP) together with the conditioning image into
\href{https://github.com/NVIDIA/cosmos-framework/blob/3845f4ca34dec8561c120a52c895c08781ec806b/cosmos_framework/inference/prompting_templates/reasoner/reasoner_i2v_json_schema.json}{NVIDIA's Cosmos I2V structured prompt format}.
The image specifies appearance and layout\rev{, whereas} OP specifies temporal dynamics.
\rev{The expansions are generated once and reused in every run.}
The balloon-inflation example contrasts OP with the benchmark's best-practice
prompt (BPP), and shows \rev{the full} expanded prompt produced from OP and
the image.

\input{sections/appendix/j_prompt_inputs}

\promptblock{Expanded prompt (OP + image)}{%
\input{sections/appendix/j_prompt_output}}
\endgroup

%% file: sections/appendix/j_prompt_inputs.tex
\promptblock{Original description (OP)}{%
A black balloon is attached to a fixed electric air pump hose on a table with a plain wall in the background. Air is being pumped in the balloon. Static shot with no camera movement.
}
\promptblock{Benchmark best-practice prompt (BPP)}{%
Static locked-off single-shot with fixed frame throughout filmed with constant framerate in real-time. The scene shows a realistic scientific demonstration. The scene only contains the described setup and actions. The balloon is attached to a black hose that points upwards from behind a wooden table. Air is being pumped into the black balloon.
}

%% file: sections/appendix/j_prompt_output.tex
\par\begingroup\footnotesize\ttfamily\setlength{\parskip}{3pt}
\noindent{\normalfont\bfseries Temporal dynamics.} Air is pumped continuously into the black balloon so that it swells steadily larger over the five seconds, its rubber stretching thinner and its highlights spreading while the neck stiffens and lifts the balloon more upright on the hose.\par
\noindent{\normalfont\bfseries Subject.} A matte black rubber balloon inflated to a firm round sphere, its neck pulled down over the black nozzle of a hose and clamped there with a band of bright blue tape.\par
\noindent{\normalfont\bfseries Setting.} A bare indoor corner used as a test bay: a plain pale grey-green painted wall fills the frame, and a honey-toned oak tabletop runs straight across the bottom as a level horizontal band. A broad soft shadow of the rig falls on the wall to the left. No other furniture, decoration or clutter is present.\par
\noindent{\normalfont\bfseries Camera.} Static\par
\noindent{\normalfont\bfseries 0:00-0:01.} The pump runs on and the black balloon, already a firm centred sphere, grows visibly wider; its paired highlights drift apart as the surface curvature flattens.\par
\noindent{\normalfont\bfseries 0:01-0:02.} The balloon continues to swell, its lower bulge rounding out and the pear-shaped taper above the blue tape shortening as rubber is drawn up into the body.\par
\noindent{\normalfont\bfseries 0:02-0:03.} Pressure rises enough to stiffen the braided hose; the balloon climbs a little higher on the vertical midline and its slight lean straightens out entirely.\par
\noindent{\normalfont\bfseries 0:03-0:04.} Expansion carries on at the same steady rate, the rubber now thinner and its blacks a shade lighter where it is most stretched, with a small wobble passing through the balloon on each pump surge.\par
\noindent{\normalfont\bfseries 0:04-0:05.} The balloon reaches its largest size in the clip, taut and nearly upright on the stiff hose, still growing and still intact as the pump keeps running.\par
\endgroup

%% file: sections/appendix/e_experimental_setup_detail.tex
\paragraph{Benchmark metrics.}
Physics-IQ Verified measures physical plausibility against a recorded
future, with higher scores indicating better results
\citep{motamed2026physicsiqverified}.
It normalizes four subscores per sample and averages equally across
subscores and inputs, replacing dataset-level normalization.
The input-level scores support within-input candidate comparisons.
WorldModelBench \citep{li2025worldmodelbench} supplies a first-frame image
and text instruction for 350 prompts spanning 7 application domains and
56 subdomains. Its human-aligned multimodal judge scores instruction
following, physical adherence over five physical laws, and common sense
over frame-wise and temporal quality. The total is the sum of these scores.
\rev{EWMBench \citep{hu2025ewmbench} evaluates embodied world models on real robot
manipulation episodes, scoring visual scene consistency, motion correctness against
the recorded trajectory, and semantic alignment with the task instruction
(Appendix~\ref{app:ewmbench}).}
Each backbone and \method{} pair uses the same conditioning
inputs, sampling protocol, and judge on the full set.

\paragraph{Tasks and candidate pools.}
The evaluation covers image-to-video and video-to-video generation on Physics-IQ \citep{motamed2026physicsiq}. Candidates are compared within the same scene, view, and conditioning input. Within matched pools of ten candidates per input, Plain scores average the pool, while WMReward and PRS select one candidate from it.
\rev{The content ablations in Table~\ref{tab:registry-ablation}(a) use Cosmos 3 Nano
16B without prompt expansion and WAN 2.2 I2V-A14B. They either replace the
projected registry features with zeros (zero mask) or apply Gaussian noise with
\(\sigma=0.5\) (perturbed mask). Their full-model references repeat the Plain
values in Table~\ref{tab:main-results}.}

\paragraph{Controlled comparison (Table~\ref{tab:conditioning-ablation}).}
All interfaces share the pretrained backbone, center-view training data,
training budget, JEPA source, and held-out left/right-view evaluation. REPA
aligns ordinary denoising hidden states to JEPA features; joint generation
adapts ReDi \citep{kouzelis2025redi} to denoise video and future JEPA tokens in
one sequence; Registry only keeps the forecast-initialized registry but removes
its alignment loss. Training uses only center views, while evaluation uses the left and right
views. This protocol is meaningful only if the JEPA condition describes the
same scenario across viewpoints. Figure~\ref{fig:jepa-cross-view} confirms this:
JEPA features from different views of the same scenario are closer than those
of different scenarios in both L2 distance and cosine similarity. A held-out
view therefore supplies a JEPA condition consistent with its scenario, and the
comparison tests how each interface uses this condition under a new viewpoint.

%% file: figures/appendix/jepa_cross_view.tex
\begin{figure}[H]
\centering
\caption{\textbf{Cross-view JEPA feature correspondence.}
Views of the same scene (blue) are closer than different scenes (red) in L2
distance (left; means 15.7 versus 28.0) and cosine similarity (right; 0.9887
versus 0.9649).}
\label{fig:jepa-cross-view}
\includegraphics[width=0.49\linewidth]{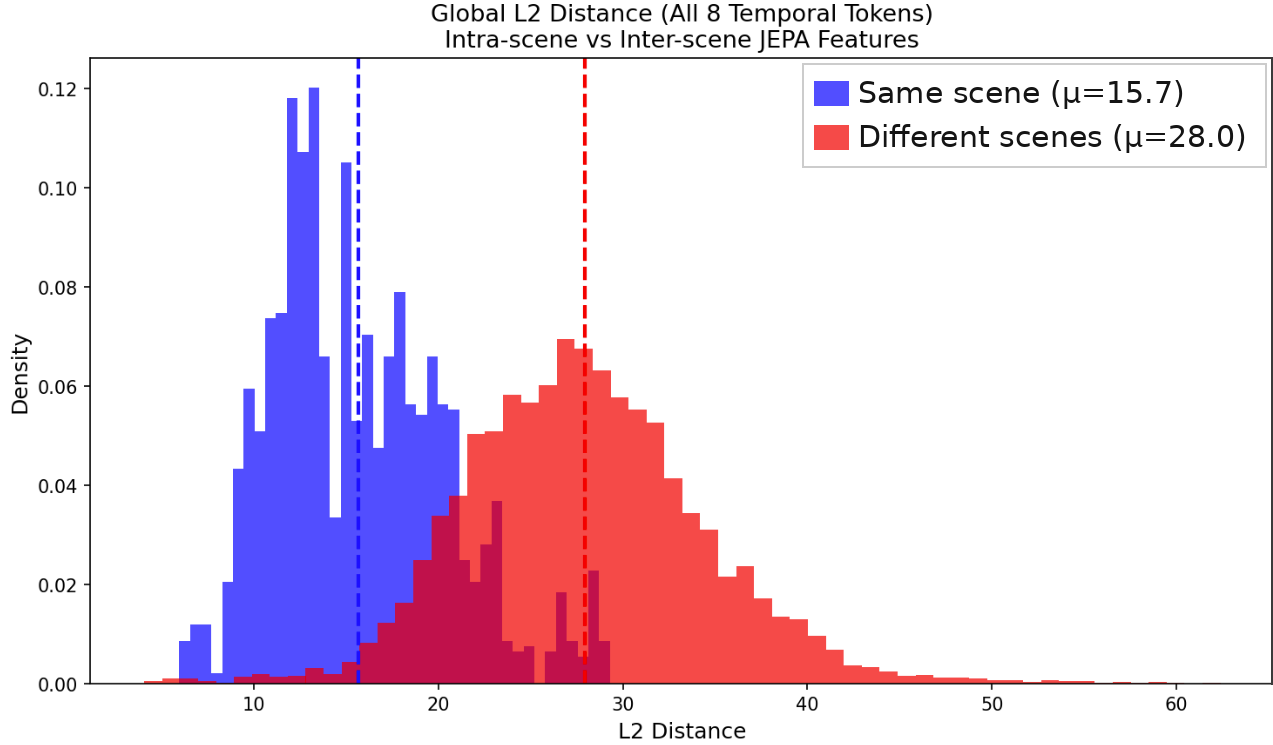}\hfill
\includegraphics[width=0.49\linewidth]{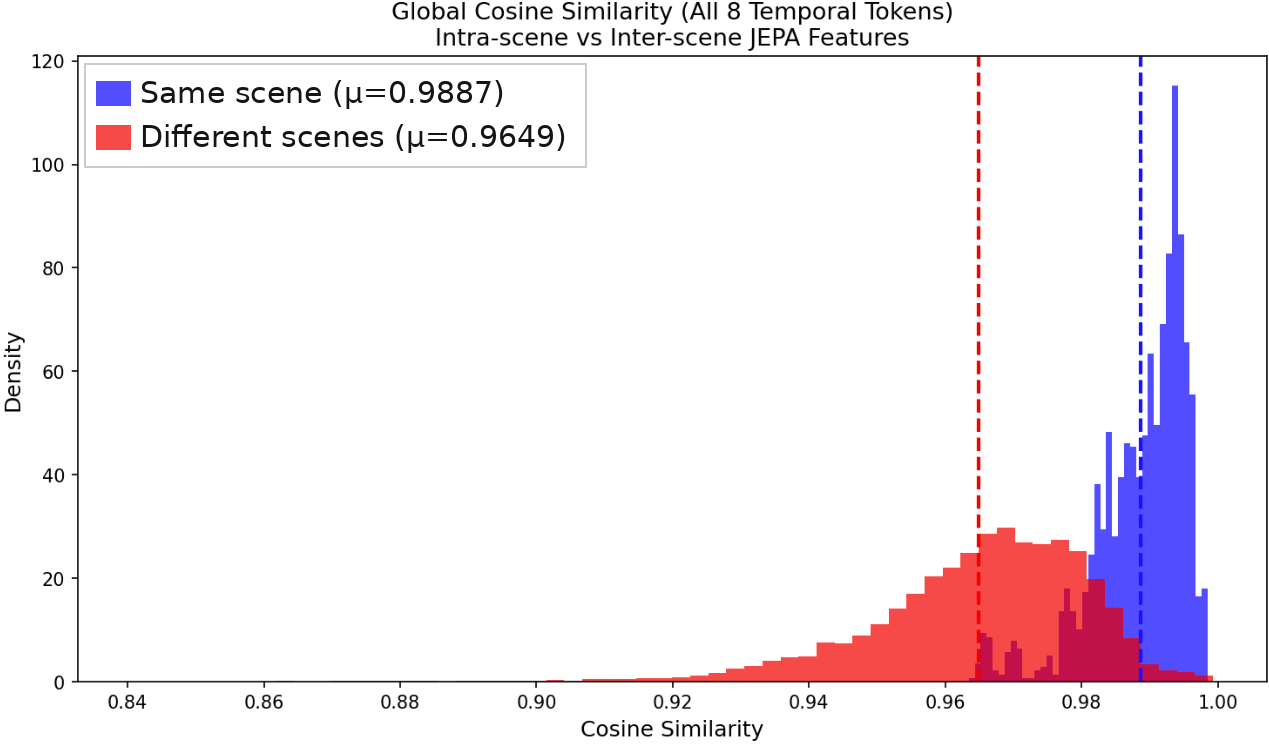}
\end{figure}

%% file: sections/appendix/k_wan5b_matched_budget.tex
\section{Evaluation Curves: Vanilla Fine-Tuning versus Physical Registry}
\label{app:wan5b-matched-budget}

\noindent\begin{minipage}{\linewidth}
\small
\setlength{\intextsep}{4pt}
The controlled comparisons in \rev{Table~\ref{tab:conditioning-ablation} and Table~\ref{tab:registry-ablation}(b)}
use Cosmos v2.5 2B\rev{. To} separate fine-tuning on \rev{the Enlarge dataset} from \rev{the} Physical Registry in
a second backbone family, we track WAN TI2V-5B for 8,000 iterations with
batch size 16, evaluating Physics-IQ Verified
every 1,000 iterations.
Vanilla fine-tuning on \rev{the Enlarge dataset} uses the same data, schedule, and optimizer as
\method{}, differing only in the absence of registry tokens and predictive alignment.

\input{figures/appendix/wan5b_matched_budget}

\rev{The vanilla fine-tuning} stays within 1.3 points of the pretrained score throughout and
ends at \(+0.15\) after 8,000 iterations.
\method{} exceeds vanilla at all eight evaluation points.
Even our smallest gain (\(+1.98\)) exceeds vanilla's largest (\(+1.25\)). \rev{Under the same data and update budget, vanilla fine-tuning produces little change
from pretrained performance, whereas \method{} improves consistently across checkpoints.
This control supports the interpretation that the WAN TI2V-5B gain in
Table~\ref{tab:main-results} arises from the registry interface rather than from
additional training data alone.}
\end{minipage}

%% file: figures/appendix/wan5b_matched_budget.tex
\begin{figure}[H]
\centering
\caption{Physics-IQ Verified gains over pretrained WAN TI2V-5B.
Shading marks one pretrained standard deviation (\(\pm0.8\)) around zero\rev{.}}
\label{fig:wan5b-matched-budget}
\includegraphics[width=0.96\linewidth]{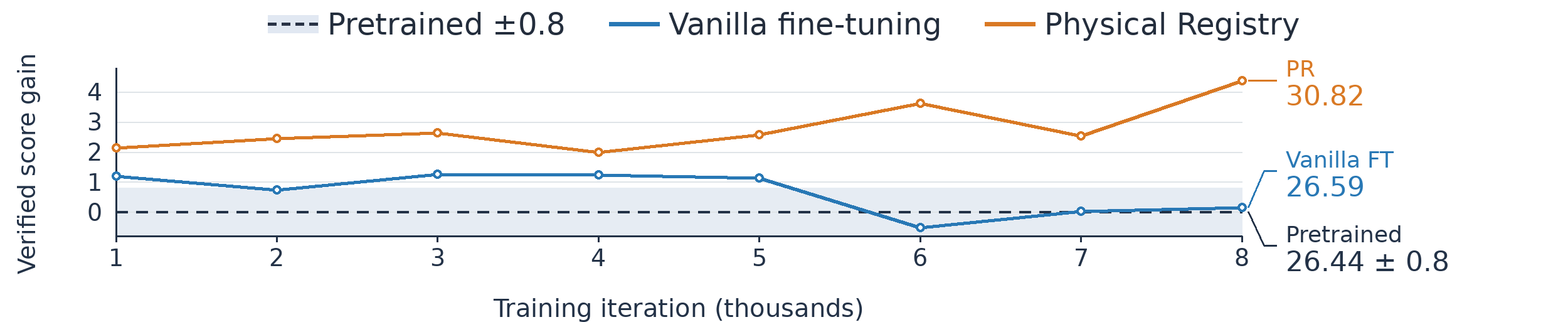}
\end{figure}

%% file: sections/appendix/c_physicsiq_components.tex
\section{Physics-IQ Components and Original Physics-IQ Score}
\label{app:physicsiq-components}

Table~\ref{tab:physicsiq-components} reports \rev{the four Physics-IQ Verified
components of the \method{} models fine-tuned on the Enlarge dataset in
Tables~\ref{tab:main-results} and~\ref{tab:selection-results}}, retaining both Nano prompt settings.
\rev{The Original column} reports the original Physics-IQ score, recomputed from the same
evaluation metrics \citep{motamed2026physicsiq,motamed2026physicsiqverified}.
\rev{Across both tasks, the model ordering under the original score is
unchanged from Verified.}

\input{tables/appendix/physicsiq_components}

%% file: tables/appendix/physicsiq_components.tex
\begin{table}[!htb]
\centering
\caption{\textbf{Physics-IQ components and original Physics-IQ score.}
Component scores, the Verified score, and the original Physics-IQ score for our
\methodfull{} \rev{(PR)} models, each normalized to 0--100.
\textbf{Bold}/\underline{underline} mark the highest/second-highest value per
column within a task. \rev{\textsuperscript{\textdagger} Expanded-prompt setting of Table~\ref{tab:main-results}.}}
\label{tab:physicsiq-components}
\begingroup
\footnotesize
\setlength{\tabcolsep}{4pt}
\begin{tabular*}{\linewidth}{@{\extracolsep{\fill}}lrrrrrr@{}}
\toprule
\textcolor{blue!55!black}{\textbf{I2V}} & Spatial & Spatiotemp. & Weighted & Inv.\ MSE & Verified & Original \\
\midrule
Nano 16B \rev{+ PR} & \underline{46.63} & 31.01 & \underline{32.76} & 33.63 & 36.01 & 35.00 \\
\addlinespace[2pt]
Nano 16B\rev{\textsuperscript{\textdagger} + PR} & \textbf{60.70} & 33.56 & \textbf{42.46} & \textbf{37.53} & \textbf{43.56} & \textbf{42.95} \\
\addlinespace[2pt]
Cosmos v2.5 2B \rev{+ PR} & 44.46 & \underline{34.44} & 30.29 & 25.65 & 33.71 & 34.00 \\
\addlinespace[2pt]
WAN I2V-A14B \rev{+ PR} & 44.77 & \textbf{42.06} & 32.67 & \underline{34.19} & \underline{38.44} & \underline{38.90} \\
\addlinespace[2pt]
WAN TI2V-5B \rev{+ PR} & 36.82 & 30.81 & 22.33 & 26.99 & 29.24 & 27.87 \\
\midrule
\textcolor{blue!55!black}{\textbf{V2V}} & Spatial & Spatiotemp. & Weighted & Inv.\ MSE & Verified & Original \\
\midrule
Nano 16B \rev{+ PR} & 57.72 & \underline{38.31} & \underline{43.43} & \underline{41.02} & \underline{45.12} & \underline{45.60} \\
\addlinespace[2pt]
Nano 16B\rev{\textsuperscript{\textdagger} + PR} & \textbf{68.05} & \textbf{38.37} & \textbf{49.90} & \textbf{41.85} & \textbf{49.54} & \textbf{50.54} \\
\addlinespace[2pt]
Cosmos v2.5 2B \rev{+ PR} & \underline{57.98} & 37.43 & 43.37 & 30.56 & 42.33 & 44.39 \\
\bottomrule
\end{tabular*}
\endgroup
\end{table}

%% file: sections/appendix/d_worldmodelbench_leaderboard.tex
\clearpage
\section{WorldModelBench Reference Leaderboard}
\label{app:worldmodelbench-leaderboard}

Table~\ref{tab:worldmodelbench-leaderboard} collects published WorldModelBench
reference scores \citep{li2025worldmodelbench} together with the backbone
comparisons in Table~\ref{tab:worldmodelbench}. All four
backbone-versus-\method{} pairs use the matched protocol described in
Appendix~\ref{app:setup-detail}.

\input{tables/appendix/worldmodelbench_leaderboard}

%% file: tables/appendix/worldmodelbench_leaderboard.tex
\begin{table}[!htbp]
\centering
\caption{\textbf{WorldModelBench reference scores.}
Backbone pairs report mean $\pm$ std, rounded to two decimals.
\textbf{Bold}/\underline{underline} mark the highest/second-highest mean in each displayed column.
Green/red parentheses show positive/negative \methodfull{}-minus-backbone changes before rounding; (--) marks a tie.
Rows from KLING onward are published references under their own generation settings \citep{li2025worldmodelbench}.
Main results: Table~\ref{tab:worldmodelbench}.}
\label{tab:worldmodelbench-leaderboard}
\begingroup
\fontsize{8}{9.4}\selectfont
\renewcommand{\scorestd}[2]{#1\,\textpm\,#2}
\setlength{\tabcolsep}{2pt}
\renewcommand{\arraystretch}{1.04}
\begin{tabular*}{\linewidth}{@{\extracolsep{\fill}}lrrrr@{}}
\toprule
Model & \shortstack[r]{Instruction\\following (0--3)} & \shortstack[r]{Physical\\adherence (0--5)} & \shortstack[r]{Common\\sense (0--2)} & \shortstack[r]{Total\\(0--10)} \\
\midrule
Cosmos 3 Nano 16B \citep{nvidia2026cosmos3} & \scorestd{2.20}{0.02} & \scorestd{4.50}{0.08} & \scorestd{1.43}{0.06} & \scorestd{8.13}{0.13} \\
\prrow\quad + \methodfull{} & \scorestd{2.20}{0.002}\scorechange{black}{--} & \scorestd{\textbf{4.82}}{0.03}\scorechange{green!45!black}{+0.32} & \scorestd{1.81}{0.02}\scorechange{green!45!black}{+0.38} & \scorestd{8.83}{0.04}\scorechange{green!45!black}{+0.69} \\
\midrule
Cosmos v2.5 2B \citep{nvidia2025cosmospredict25} & \scorestd{2.14}{0.03} & \scorestd{4.43}{0.07} & \scorestd{1.55}{0.05} & \scorestd{8.12}{0.14} \\
\prrow\quad + \methodfull{} & \scorestd{2.07}{0.01}\scorechange{red!70!black}{-0.07} & \scorestd{4.54}{0.03}\scorechange{green!45!black}{+0.11} & \scorestd{1.58}{0.04}\scorechange{green!45!black}{+0.03} & \scorestd{8.19}{0.01}\scorechange{green!45!black}{+0.07} \\
\midrule
WAN 2.2 I2V-A14B \citep{wanteam2025wan22} & \scorestd{2.23}{0.03} & \scorestd{4.45}{0.04} & \scorestd{1.60}{0.03} & \scorestd{8.28}{0.02} \\
\prrow\quad + \methodfull{} & \scorestd{2.30}{0.04}\scorechange{green!45!black}{+0.07} & \scorestd{4.53}{0.04}\scorechange{green!45!black}{+0.07} & \scorestd{1.60}{0.02}\scorechange{green!45!black}{+0.004} & \scorestd{8.43}{0.05}\scorechange{green!45!black}{+0.15} \\
\midrule
WAN 2.2 TI2V-5B \citep{wanteam2025wan22} & \scorestd{\textbf{2.33}}{0.01} & \scorestd{4.35}{0.10} & \scorestd{1.57}{0.04} & \scorestd{8.24}{0.14} \\
\prrow\quad + \methodfull{} & \scorestd{2.24}{0.05}\scorechange{red!70!black}{-0.09} & \scorestd{4.50}{0.05}\scorechange{green!45!black}{+0.14} & \scorestd{1.68}{0.06}\scorechange{green!45!black}{+0.11} & \scorestd{8.41}{0.11}\scorechange{green!45!black}{+0.17} \\
\midrule
KLING \citep{kuaishou2024kling} & \underline{2.32} & \underline{4.81} & \textbf{1.95} & \textbf{9.10} \\
Minimax \citep{minimax2024hailuo} & 2.28 & 4.72 & \underline{1.92} & \underline{8.92} \\
Mochi 1-official \citep{genmo2024mochi} & 2.00 & 4.80 & 1.86 & 8.66 \\
Runway \citep{runway2024gen3} & 2.17 & 4.60 & 1.86 & 8.64 \\
Luma \citep{lumalabs2024dreammachine} & 1.98 & 4.49 & 1.76 & 8.24 \\
OpenSoraPlan-T2V \citep{lin2024opensoraplan} & 1.72 & 4.64 & 1.68 & 8.04 \\
Mochi 1 \citep{genmo2024mochi} & 2.06 & 4.39 & 1.45 & 7.91 \\
CogVideoX-T2V \citep{yang2025cogvideox} & 2.03 & 4.27 & 1.35 & 7.65 \\
CogVideoX-I2V \citep{yang2025cogvideox} & 1.78 & 4.17 & 1.12 & 7.08 \\
Pandora \citep{xiang2024pandora} & 1.56 & 4.31 & 1.03 & 6.90 \\
T2V-Turbo \citep{li2024t2vturbo} & 1.37 & 4.10 & 1.08 & 6.56 \\
OpenSora-T2V \citep{zheng2024opensora} & 1.61 & 3.88 & 0.69 & 6.17 \\
OpenSora-I2V \citep{zheng2024opensora} & 1.42 & 3.85 & 0.55 & 5.82 \\
\bottomrule
\end{tabular*}
\endgroup
\end{table}

%% file: sections/appendix/m_ewmbench.tex
\clearpage
\section{Embodied World Model Benchmark}
\label{app:ewmbench}

EWMBench \citep{hu2025ewmbench} evaluates embodied world models on real robot
manipulation episodes along three axes: visual scene consistency, motion
correctness against the recorded gripper trajectory, and semantic alignment
with the task instruction. It complements Physics-IQ, which scores physical
plausibility on passive scenes, by asking whether a generated rollout follows
an embodiment through a task.

\paragraph{Protocol.}
We run the official evaluator on the full public ground-truth split, 21
episodes over seven tasks. Each episode is generated three times from its
initial frame and instruction, giving 126 videos across the two arms.
Our clips are 121 frames at 24\,fps\rev{, whereas} the recorded episodes run at
30\,fps. Both arms share the backbone, sampling recipe, and conditioning
inputs, and differ only in the registry interface.

\input{tables/appendix/ewmbench}

\paragraph{Results.}
\methodfull{} improves scene consistency and all three motion metrics
(Table~\ref{tab:ewmbench}). The largest gains are in trajectory agreement, where
NDTW rises by 0.073 and DYN by 0.076.
Semantic metrics change by at most 0.013.
\rev{Diversity, the spread across the three seeds, decreases from 0.029 to 0.018.}

%% file: tables/appendix/ewmbench.tex
\begin{table}[!htb]
\centering
\caption{\textbf{EWMBench \citep{hu2025ewmbench}.}
All metrics are higher-is-better and shown on a 0--1 scale.
Bold/underline mark the highest/second-highest value in each column across all
rows, and green parentheses give the \methodfull{} change over the backbone
above it.}
\label{tab:ewmbench}
\begingroup
\fontsize{7.4}{8.8}\selectfont
\newcommand{\dgain}[1]{{\fontsize{6.3}{7.2}\selectfont\textcolor{green!45!black}{(#1)}}}%
\newcommand{\dloss}[1]{{\fontsize{6.3}{7.2}\selectfont\textcolor{red!70!black}{(#1)}}}%
\setlength{\tabcolsep}{2.5pt}
\renewcommand{\arraystretch}{1.08}
\begin{tabular*}{\linewidth}{@{\extracolsep{\fill}}lrrrrrrrr@{}}
\toprule
& \multicolumn{1}{c}{Scene} & \multicolumn{3}{c}{Motion} &
\multicolumn{3}{c}{Semantics} & \multicolumn{1}{c}{Diversity} \\
\cmidrule(lr){2-2}\cmidrule(lr){3-5}\cmidrule(lr){6-8}\cmidrule(lr){9-9}
Model & Consist. & HSD & NDTW & DYN & BLEU & CLIP & Logic & Div. \\
\midrule
Cosmos 3 Nano 16B \citep{nvidia2026cosmos3}
& 0.855 & 0.328 & 0.305 & 0.383 & \underline{0.242} & 0.884 & 0.921 & 0.029 \\
\prrow\quad \(+\)\methodfull{}
& 0.895 & 0.351 & 0.378 & 0.459
& \textbf{0.244} & \textbf{0.897} & 0.921 & 0.018 \\
\prrow& \dgain{+0.040} & \dgain{+0.023} & \dgain{+0.073} & \dgain{+0.076}
& \dgain{+0.002} & \dgain{+0.013} & & \dloss{-0.011} \\
\midrule
EnerVerse-FT\textsuperscript{*} \citep{huang2025enerverse}
& \underline{0.943} & \textbf{0.536} & \textbf{0.596} & \underline{0.536} & 0.180 & 0.864 & \textbf{0.978} & \underline{0.069} \\
LTX-FT\textsuperscript{*} \citep{hacohen2025ltxvideo}
& \textbf{0.944} & \underline{0.476} & \underline{0.521} & \textbf{0.620} & 0.174 & 0.855 & \underline{0.967} & 0.016 \\
Kling\textsuperscript{*} \citep{kuaishou2024kling}
& 0.889 & 0.323 & 0.316 & 0.305 & 0.168 & 0.854 & \underline{0.967} & 0.049 \\
Hailuo\textsuperscript{*} \citep{minimax2024hailuo}
& 0.858 & 0.223 & 0.179 & 0.134 & 0.185 & \underline{0.886} & 0.911 & 0.037 \\
COSMOS\textsuperscript{*} \citep{nvidia2025cosmos}
& 0.796 & 0.250 & 0.253 & 0.205 & 0.123 & 0.846 & 0.733 & \textbf{0.080} \\
OpenSora\textsuperscript{*} \citep{zheng2025opensora2}
& 0.921 & 0.155 & 0.142 & 0.047 & 0.160 & 0.851 & 0.822 & 0.042 \\
LTX\textsuperscript{*} \citep{hacohen2025ltxvideo}
& 0.916 & 0.158 & 0.143 & 0.100 & 0.069 & 0.832 & 0.733 & 0.017 \\
\bottomrule
\end{tabular*}
\par\smallskip
\parbox{\linewidth}{\fontsize{7}{8.2}\selectfont\raggedright
\textsuperscript{*} Published in the EWMBench evaluation table
\citep[Table 2]{hu2025ewmbench}, evaluated on ten tasks with ten ground-truth
episodes each; the public ground-truth release used here covers 21 episodes
across seven tasks. EnerVerse-FT and LTX-FT are adapted to the benchmark's
robot domain; our models are not.}
\endgroup
\end{table}

%% file: sections/appendix/e_selector_detail.tex
\section{The Physical Registry Selector}
\label{app:selector-detail}

\subsection{Aggregation and protocol}
\label{app:selector-protocol}
\label{app:selection-analysis}

We apply the PRS rule in Equation~\eqref{eq:selector-main} to candidates generated
from the same conditioning input.
Each cosine is evaluated before slot--step averaging.
Prefix selection follows the PRS settings in Table~\ref{tab:selection-results}.
The registry subset \(\mathcal S\) is the contiguous range of registry slot
positions with the largest departure from the shared forecast.
\rev{Nano\textsuperscript{\textdagger}} averages over the latest \(\min(K,10)\)
evaluations.
Prefix-selected quality is measured on the chosen candidate's completed
video under its original 35- or 50-step schedule.

\rev{With \(N\) candidates and a full trajectory of \(D\) steps, continuing one
candidate after \(K\) prefix steps gives a serial denoising budget of
\(NK+(D-K)\) candidate-steps, compared with \(ND\) for completing all
candidates. All selection budgets are reported in these serial denoising
candidate-steps.}
In the completed-video \rev{best-of-\(N\) (BoN)} comparison, WMReward scores decoded candidates
with its external world model \citep{yuan2026wmreward}.
\rev{Relative gains of PRS over Plain are computed as}
\(100(\mathrm{PRS}-\mathrm{Plain})/\mathrm{Plain}\).
All reported selection policies and Oracles use ten candidates per input.
For backbone rows marked \textsuperscript{*}, Plain is the published
reference, while WMReward and Oracle select from our own ten-candidate pool of
that backbone; \methodfull{} rows use their own pool for every column.

\input{tables/main/selection_results}
\input{figures/main/selector_mechanism}

\subsection{Direction and consistency}
\label{app:prs-direction}
\label{app:prs-seed-sensitivity}

\input{sections/appendix/i_prs_conditional_analysis}

\begin{samepage}
\subsection{Early selection}
\label{app:prs-early-selection}

Figure~\ref{fig:selector-results} compares table-aligned prefix selection for
Cosmos v2.5 and the two Nano configurations. Cosmos v2.5 first exceeds
WMReward after 8/10 steps on I2V/V2V.
\rev{At ten steps, Cosmos v2.5 selection already slightly exceeds completed-video WMReward
on both tasks. It agrees with the full-schedule PRS decision on 72.7\%/56.1\%
of I2V/V2V inputs, and this agreement increases as more denoising steps are
used for selection (Figure~\ref{fig:selector-results}c,d).}
\par
\end{samepage}

\input{figures/main/selector_results}

\subsection{Selection cost}
\label{app:prs-cost}

Continuing only the selected candidate \rev{saves candidate-steps as well as}
the decodes and reward evaluations that completed-video selection
needs. Table~\ref{tab:selection-cost} times both policies end to end on the
same \methodfull{} model, using \(K=10\) as an illustrative prefix length;
Table~\ref{tab:selection-results} reports quality under the PRS settings, and
Figure~\ref{fig:selector-results} shows quality as a function of the prefix
length. On Cosmos v2.5 the measured wall-clock ratio is
\(2.84\times\) against the \(2.80\times\) (\(64\%\)) candidate-step reduction reported in
Section~\ref{sec:selector-analysis}. \rev{Across backbones, measured wall-clock
speedups range from \(2.84\times\) to \(4.06\times\) and broadly track the
reduction in candidate-steps, while also reflecting backbone-specific sampling,
decoding, and reward-evaluation costs.}

\input{tables/appendix/selection_cost}

%% file: tables/main/selection_results.tex
\begin{table}[!htb]
\centering
\caption{\textbf{Selection policies on Physics-IQ Verified.}
WMReward and PRS each select one candidate from our ten-candidate pool; Oracle
is the best in that pool, an upper bound rather than a policy.
\textsuperscript{*} marks published Plain values, external to the pool, and
\textsuperscript{\textdagger} the configuration in Table~\ref{tab:main-results}.
Bold marks the strongest non-Oracle score in each row, gray unavailable PRS.}
\label{tab:selection-results}
\begingroup
\footnotesize
\renewcommand{\scorestd}[2]{#1\,\textpm\,#2}
\setlength{\tabcolsep}{4pt}
\renewcommand{\arraystretch}{1.00}
\begin{tabular*}{\linewidth}{@{\extracolsep{\fill}}lrrrr@{}}
\toprule
\textcolor{blue!55!black}{\textbf{I2V}} & Plain & WMReward & PRS & Oracle \\
\midrule
Cosmos 3 Nano 16B & \scorestd{30.18}{0.84} & \textbf{31.69} & \DRSUnavailable & 41.48 \\
\prrow\quad + \methodfull{} & \scorestd{36.01}{0.95} & 38.85 & \textbf{40.27} & 51.88 \\
\addlinespace[2pt]
Cosmos 3 Nano 16B\textsuperscript{\textdagger} & \scorestd{37.3}{0.9}\textsuperscript{*} & \textbf{44.28} & \DRSUnavailable & 50.59 \\
\prrow\quad + \methodfull{}\textsuperscript{\textdagger} & \scorestd{43.56}{0.95} & 45.41 & \textbf{45.81} & 54.72 \\
\addlinespace[2pt]
Cosmos v2.5 2B & \scorestd{\textbf{30.65}}{0.37} & 30.61 & \DRSUnavailable & 36.07 \\
\prrow\quad + \methodfull{} & \scorestd{33.71}{1.11} & 37.74 & \textbf{39.08} & 49.37 \\
\addlinespace[2pt]
WAN 2.2 I2V-A14B & \scorestd{32.2}{0.6}\textsuperscript{*} & \textbf{35.31} & \DRSUnavailable & 43.71 \\
\prrow\quad + \methodfull{} & \scorestd{38.44}{1.06} & \textbf{41.82} & 40.58 & 51.87 \\
\addlinespace[2pt]
WAN 2.2 TI2V-5B & \scorestd{27.7}{0.9}\textsuperscript{*} & \textbf{31.19} & \DRSUnavailable & 37.76 \\
\prrow\quad + \methodfull{} & \scorestd{29.24}{1.30} & \textbf{33.18} & 32.03 & 39.88 \\
\midrule
\textcolor{blue!55!black}{\textbf{V2V}} & Plain & WMReward & PRS & Oracle \\
\midrule
Cosmos 3 Nano 16B & \scorestd{38.78}{0.78} & \textbf{42.46} & \DRSUnavailable & 49.09 \\
\prrow\quad + \methodfull{} & \scorestd{45.12}{0.94} & \textbf{49.70} & 46.78 & 59.43 \\
\addlinespace[2pt]
Cosmos 3 Nano 16B\textsuperscript{\textdagger} & \scorestd{46.45}{0.76} & \textbf{50.41} & \DRSUnavailable & 56.98 \\
\prrow\quad + \methodfull{}\textsuperscript{\textdagger} & \scorestd{49.54}{0.63} & 51.69 & \textbf{52.06} & 59.69 \\
\addlinespace[2pt]
Cosmos v2.5 2B & \scorestd{37.40}{0.53} & \textbf{39.38} & \DRSUnavailable & 42.60 \\
\prrow\quad + \methodfull{} & \scorestd{42.33}{1.90} & 46.59 & \textbf{47.67} & 55.84 \\
\bottomrule
\end{tabular*}
\endgroup
\end{table}

%% file: figures/main/selector_mechanism.tex
\begin{figure}[!htbp]
\centering
\caption{\textbf{Candidate-conditioned registry readouts and selection direction.}
(a) Mean cosine to the generated-video reference: the terminal readout tracks
the candidate more closely than the shared forecast (0.974 versus 0.764).
(b) Within-scene readout--generated-video agreement.
(c) \rev{PRS against controls per backbone and task: Random is} uniform selection in the same pool and
Inverse the reversed ranking.}
\label{fig:selector-mechanism}
\begingroup
\begin{minipage}[t]{0.32\linewidth}
\vspace{0pt}
\centering
\includegraphics[width=\linewidth]{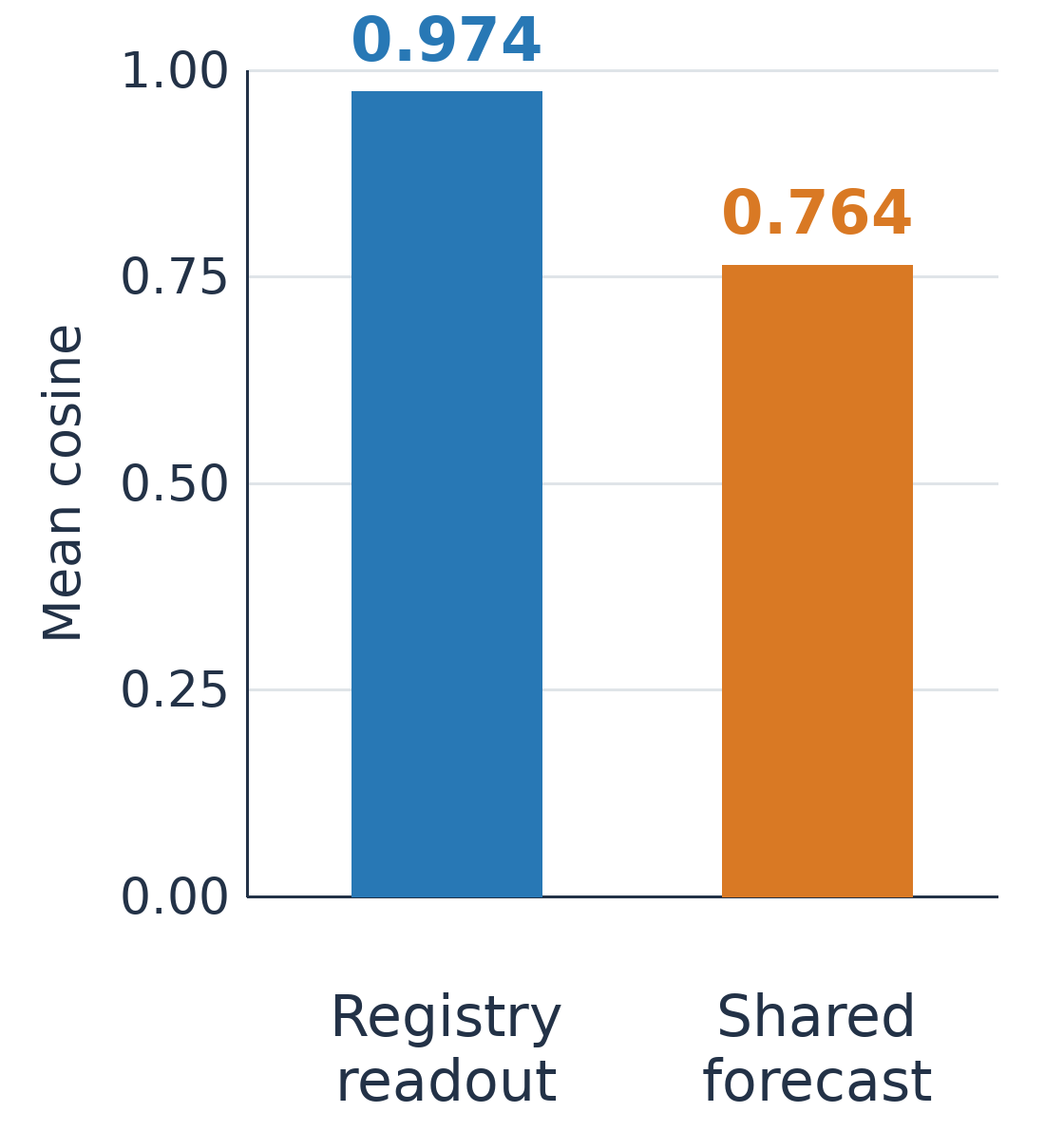}
\end{minipage}\hfill
\begin{minipage}[t]{0.32\linewidth}
\vspace{0pt}
\centering
\includegraphics[width=\linewidth]{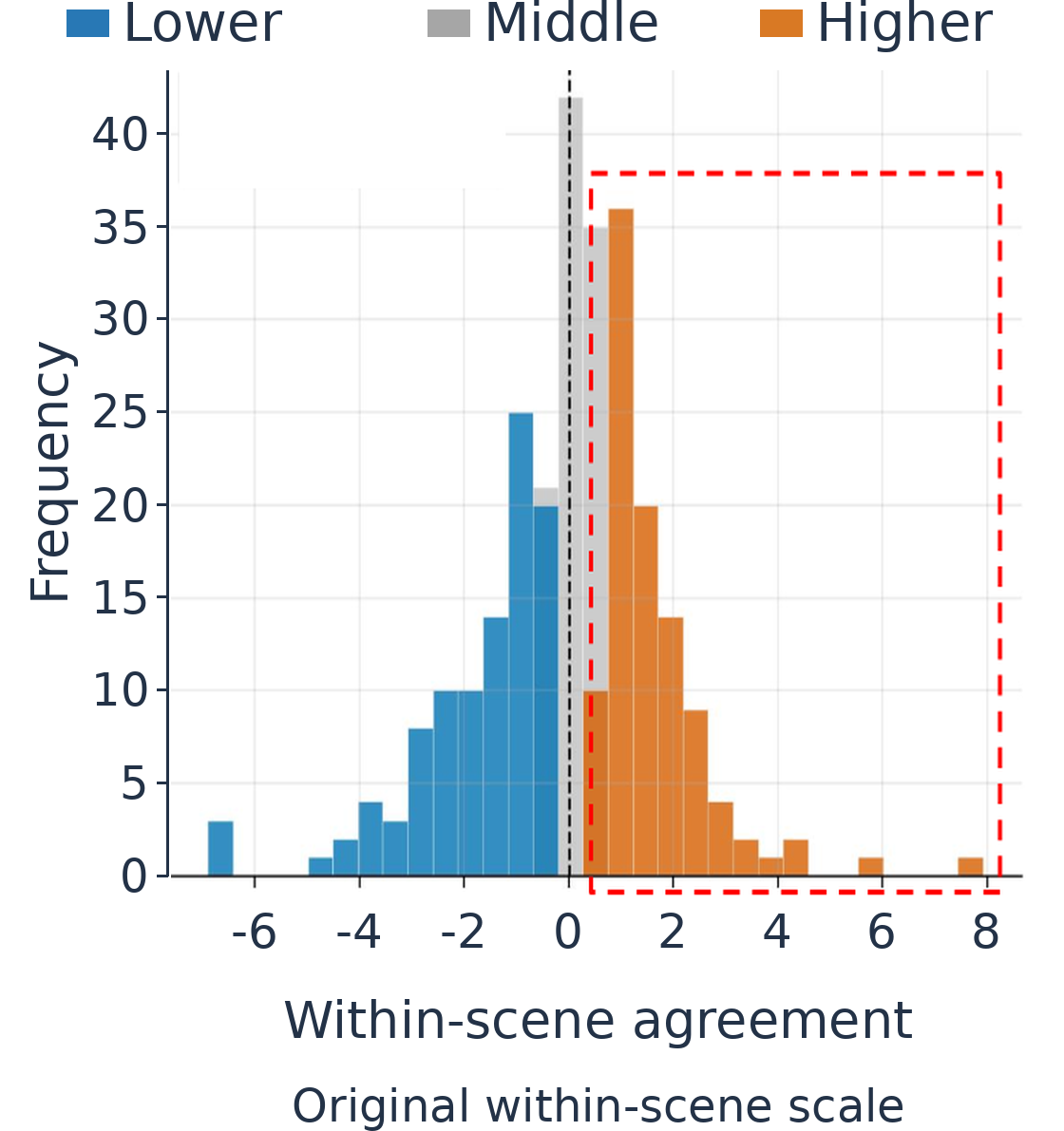}
\end{minipage}\hfill
\begin{minipage}[t]{0.32\linewidth}
\vspace{0pt}
\centering
\edef\ReadoutBodyHeight{\the\dimexpr1.178571\linewidth\relax}
\begin{minipage}[t][\ReadoutBodyHeight][c]{\linewidth}
\centering
\fontsize{7.4}{8.6}\selectfont
\setlength{\tabcolsep}{1.05pt}
\renewcommand{\arraystretch}{1.0}
\input{tables/main/selection_family}
\end{minipage}
\end{minipage}
\endgroup
\end{figure}

%% file: tables/main/selection_family.tex
\begin{tabular}{@{}llrrr@{}}
\toprule
& & \multicolumn{3}{c}{Selection} \\
\cmidrule(lr){3-5}
Model & Task & PRS & Inverse & Random \\
\midrule
\shortstack[l]{Cosmos\\v2.5} & I2V & \textbf{39.08} & 25.60 & 33.71 \\
 & V2V & \textbf{47.67} & 33.76 & 42.33 \\
\midrule
\shortstack[l]{Cosmos\\Nano} & I2V & \textbf{40.27} & 33.12 & 36.01 \\
 & V2V & \textbf{46.78} & 42.62 & 45.12 \\
\midrule
WAN 14B & I2V & \textbf{40.58} & 34.73 & 38.44 \\
\midrule
WAN 5B & I2V & \textbf{32.03} & 26.91 & 29.24 \\
\bottomrule
\end{tabular}

%% file: sections/appendix/i_prs_conditional_analysis.tex
\paragraph{The registry readout tracks the realized candidate.}
For each completed candidate \(v_{\mathrm{gen}}^{(n)}\), we encode the
generated video with the same frozen target encoder to obtain a diagnostic
reference \(Z_{\mathrm{gen}}^{(n)}\).
Figure~\ref{fig:selector-mechanism}(a) compares this reference with the
terminal registry readout and with the shared observation-conditioned
forecast.
The terminal readout has \rev{a} higher mean cosine similarity to the
generated-video reference than the shared forecast
(\(0.974\) versus \(0.764\)).
This shows that, after interacting with the evolving video tokens, the registry
more closely reflects the future realized by the candidate than the common
forecast from which it was initialized.

\paragraph{This candidate dependence is visible within scenes.}
Figure~\ref{fig:selector-mechanism}(b) resolves the same comparison within
individual scenes, showing variation in readout--generated-video agreement
across candidates that share the same observation-conditioned forecast.
Thus, the terminal registry is not merely preserving the common initialization:
its readout varies with the trajectory produced by each candidate.

\paragraph{Selection direction.}
PRS outperforms both the reversed ranking (Inverse) and uniform selection
(Random) in every configuration (Figure~\ref{fig:selector-mechanism}(c)),
supporting the direction in Equation~\eqref{eq:selector-main}. The direction
also holds for smaller pools: averaged over all pools of four to six candidates
drawn from the ten, PRS keeps positive gains over both controls, and paired
scenario-level bootstrap intervals exclude zero. Against completed-video
WMReward, PRS is comparable while selecting before the candidates are fully
denoised. Within inputs, departure is also positively rank-correlated with
Verified quality (e.g., Spearman \(0.297\) \([0.209, 0.383]\) for Cosmos v2.5
I2V at pool size four).
Entries below are PRS-minus-control gains in Physics-IQ Verified points with
95\% intervals.

\begin{center}
\footnotesize
\setlength{\tabcolsep}{3pt}
\begin{tabular*}{\linewidth}{@{\extracolsep{\fill}}lrlrrr@{}}
\toprule
Model & Pool & Task & Inverse & Random & WMReward \\
\midrule
Cosmos v2.5 2B & 6 & I2V & \(+12.58\,[9.19,16.08]\) & \(+5.07\,[3.01,7.23]\) & \(+1.33\,[-0.45,3.33]\) \\
 & 6 & V2V & \(+12.69\,[8.60,17.04]\) & \(+4.90\,[2.85,7.09]\) & \(+0.60\,[-0.64,1.92]\) \\
 & 5 & I2V & \(+11.46\,[8.39,14.66]\) & \(+4.73\,[2.94,6.59]\) & \(+1.24\,[-0.24,2.98]\) \\
 & 5 & V2V & \(+11.37\,[7.70,15.34]\) & \(+4.50\,[2.65,6.48]\) & \(+0.22\,[-0.84,1.36]\) \\
 & 4 & I2V & \(+10.05\,[7.34,12.85]\) & \(+4.32\,[2.80,5.87]\) & \(+1.05\,[-0.16,2.49]\) \\
 & 4 & V2V & \(+9.86\,[6.62,13.30]\) & \(+4.08\,[2.48,5.81]\) & \(-0.05\,[-0.91,0.87]\) \\
\midrule
WAN 2.2 I2V-A14B & 6 & I2V & \(+4.35\,[2.53,6.28]\) & \(+1.82\,[0.92,2.74]\) & \(-0.51\,[-1.31,0.30]\) \\
 & 5 & I2V & \(+3.99\,[2.32,5.76]\) & \(+1.71\,[0.88,2.55]\) & \(-0.47\,[-1.19,0.25]\) \\
 & 4 & I2V & \(+3.53\,[2.04,5.10]\) & \(+1.56\,[0.83,2.31]\) & \(-0.42\,[-1.05,0.20]\) \\
\bottomrule
\end{tabular*}
\end{center}

%% file: figures/main/selector_results.tex
\begin{figure}[!htbp]
\centering
\caption{
\textbf{Early selection and decision agreement.}
(a,b) Prefix-selected Physics-IQ Verified gain over WMReward using the
PRS settings in Table~\ref{tab:selection-results}; zero marks the WMReward reference.
(c,d) Agreement with the full-schedule PRS decision.
Horizontal axes show the fraction of denoising steps used for selection.
\textsuperscript{\textdagger} denotes the \rev{expanded-prompt} Nano configuration in Table~\ref{tab:main-results}.
Protocol: Appendix~\ref{app:selector-protocol}.
}
\label{fig:selector-results}
\includegraphics[width=\linewidth]{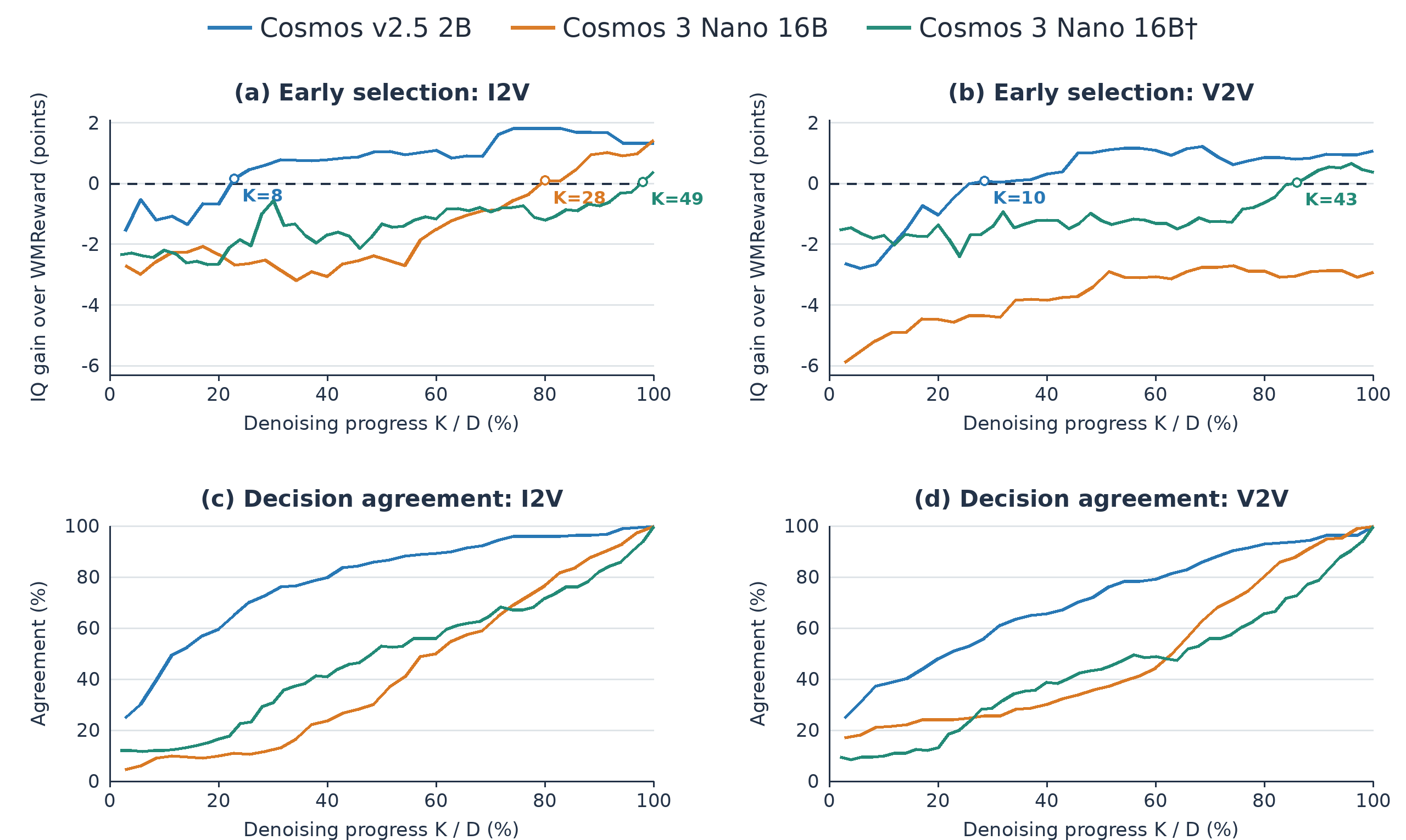}
\end{figure}

%% file: tables/appendix/selection_cost.tex
\begin{table}[!htbp]
\centering
\caption{\textbf{End-to-end cost of selection.}
One input, \(N=10\) candidates, both policies run on the same \methodfull{}
model inside one job on a single H100 80\,GB; model loading is excluded.
Complete BoN times sampling, all ten VAE decodes, and the ten WMReward
evaluations; PRS times the ten prefixes, the continuation of the selected
candidate, and its single decode. \(K=10\) is an illustrative prefix length.
\rev{\textsuperscript{\textdagger} Expanded-prompt setting (Table~\ref{tab:inference-configuration}).}}
\label{tab:selection-cost}
\footnotesize
\setlength{\tabcolsep}{3pt}
\renewcommand{\arraystretch}{1.0}
\begin{tabular*}{\linewidth}{@{\extracolsep{\fill}}llrrrr@{}}
\toprule
Backbone & Policy & Candidate-steps & Decodes & Wall-clock (s) & Speed-up \\
\midrule
Cosmos v2.5 2B & Complete BoN \(+\) WMReward & 350 & 10 & 4,031 & 1.00\(\times\) \\
 & PRS (\(K=10\) of \(D=35\)) & 125 & 1 & 1,420 & \(2.84\times\) \\
\addlinespace[2pt]
Cosmos 3 Nano 16B\rev{\textsuperscript{\textdagger}} & Complete BoN \(+\) WMReward & 500 & 10 & 1,985 & 1.00\(\times\) \\
 & PRS (\(K=10\) of \(D=50\)) & 140 & 1 & 614 & \(3.23\times\) \\
\addlinespace[2pt]
WAN 2.2 I2V-A14B & Complete BoN \(+\) WMReward & 400 & 10 & 10,748 & 1.00\(\times\) \\
 & PRS (\(K=10\) of \(D=40\)) & 130 & 1 & 3,289 & \(3.27\times\) \\
\addlinespace[2pt]
WAN 2.2 TI2V-5B & Complete BoN \(+\) WMReward & 500 & 10 & 2,834 & 1.00\(\times\) \\
 & PRS (\(K=10\) of \(D=50\)) & 140 & 1 & 697 & \(4.06\times\) \\
\bottomrule
\end{tabular*}
\end{table}

%% file: sections/appendix/l_dynamics_transfer.tex
\clearpage
\section{\texorpdfstring{\rev{Motion}}{Motion} Transfer}
\label{app:dynamics-transfer}

\subsection{Setup and scope}
\label{app:dynamics-transfer-setup}

To test whether registry content can influence generated motion, we replace
the native forecast \rev{of a receiver scene (the white-sphere scene)} with \rev{source}-video JEPA features on Cosmos 3 Nano 16B with
\methodfull{}.
Both configurations use the white-sphere image and general prompt shown in
Figures~\ref{fig:dynamics-transfer-method} and~\ref{fig:dynamics-transfer-results}.
The conditioning image, text, initial noise, and sampler are held fixed.
The text specifies no falling, bouncing, or settling behavior.

\paragraph{Experimental setup.}
The native-forecast configuration combines one source-context feature with seven
predicted features. The transfer configuration instead uses eight features
from a frozen JEPA video encoder applied to the \rev{source} video.
Both representations have shape \(8\times196\times5632\) and enter the same
trained projection and video-generation interface. The \rev{source} features are
full video representations, containing scene content as well as dynamics.
For this comparison, both configurations use the same factor-two amplification
of the projected registry residual in the conditional branch,
\(P(0)+2[P(J)-P(0)]\)\rev{. The} unconditional branch uses \(P(J)\) without
amplification. Here \(P\) is the trained projection and \(J\) the chosen
registry input.
Both receiver outputs use the same inference configuration:
50 denoising steps, \rev{classifier-free guidance scale} 3.0, flow shift 4.75, and
\(1280\times720\) outputs with 121 frames at 24 fps.

\input{figures/appendix/dynamics_transfer_method}

\input{figures/appendix/dynamics_transfer_results}

\subsection{\texorpdfstring{\rev{Source}}{Source} examples}
\label{app:dynamics-transfer-examples}

Each example compares two generations that share the conditioning image,
prompt, initial noise, and sampler: A uses the native forecast, and B replaces
it with the source video's JEPA features.

\paragraph{Bowling ball: bounce suppression.}
In the source video, a heavy bowling ball lands and stays on the table. With the
native forecast, the generated ball (A) bounces three times with decreasing
height before settling. With the bowling-ball features (B), it lands and
settles with almost no rebound (Figure~\ref{fig:dynamics-transfer-results}),
and the maximum upward return drops from 80.1 to 3.6 pixels
(Table~\ref{tab:dynamics-transfer-motion}).

\input{tables/appendix/dynamics_transfer_motion}

\paragraph{Balloon: slower descent.}
In the source video, a balloon descends slowly to the table, so we compare the
descent before contact. With the balloon features (B), the ball descends more
slowly than with the native forecast (A)
(Figure~\ref{fig:dynamics-transfer-balloon}). It takes 478 instead of 371\,ms
(\(1.29\times\)) to pass through the same height range, and first contact moves
from 0.375 to 0.500\,s. The rebound after contact also drops from 40.9 to 12.1
pixels. First-frame silhouette areas differ by only \(1.0\%\), so the slower
descent is not explained by a change in ball size.

\input{figures/appendix/dynamics_transfer_balloon}

Changing only the registry content changes the generated motion, indicating
that the generator uses the registry's predicted dynamics when producing motion.

%% file: figures/appendix/dynamics_transfer_method.tex
\begin{figure}[!htbp]
\centering
\caption{\textbf{Registry-source configurations for \rev{motion} transfer.}
The native \rev{configuration} uses the conditioning image's JEPA context and forecast;
the \rev{transfer configuration} uses video-encoded features from the \rev{source video}.
Both use the same conditioning frame, general text, initial noise, and projection recipe.
The exact prompt is shown in a monospace block and contains no motion description.}
\label{fig:dynamics-transfer-method}
\includegraphics[width=\linewidth]{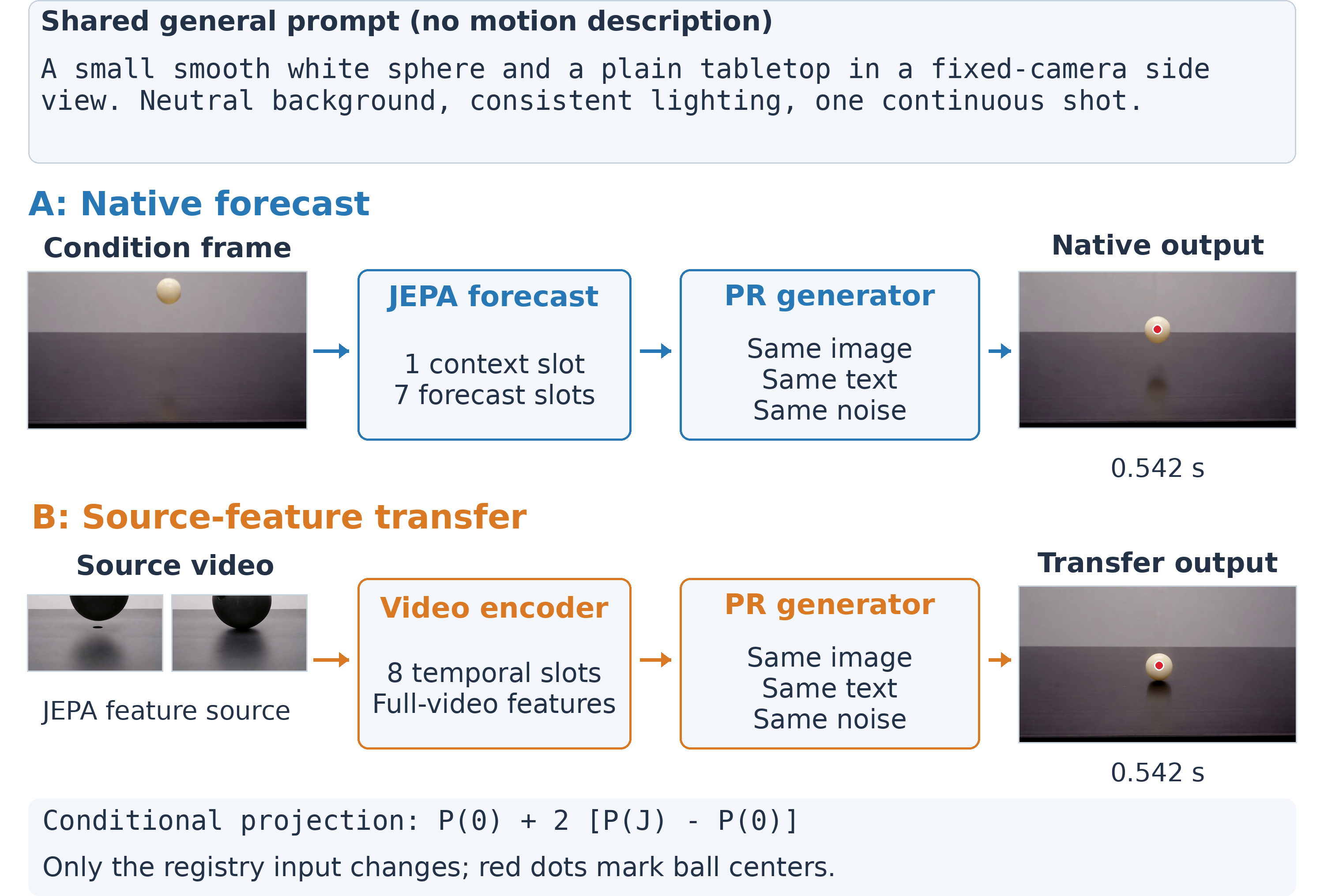}
\end{figure}

%% file: figures/appendix/dynamics_transfer_results.tex
\begin{figure}[!htbp]
\centering
\caption{\textbf{\rev{Motion} transfer with matched inputs.}
The source video supplies JEPA features; A and B are generated videos.
Red dots track ball centers and rows share timestamps.
The plot shows \rev{the} upward return \(r(t)=\max_{s\leq t}y(s)-y(t)\)\rev{, with} image \(y\)
increasing downward, over the second after A's first landing (0.333 s)\rev{.}
B first lands at 0.417 s.
Numbers mark A's three diminishing returns.}
\label{fig:dynamics-transfer-results}
\includegraphics[width=\linewidth]{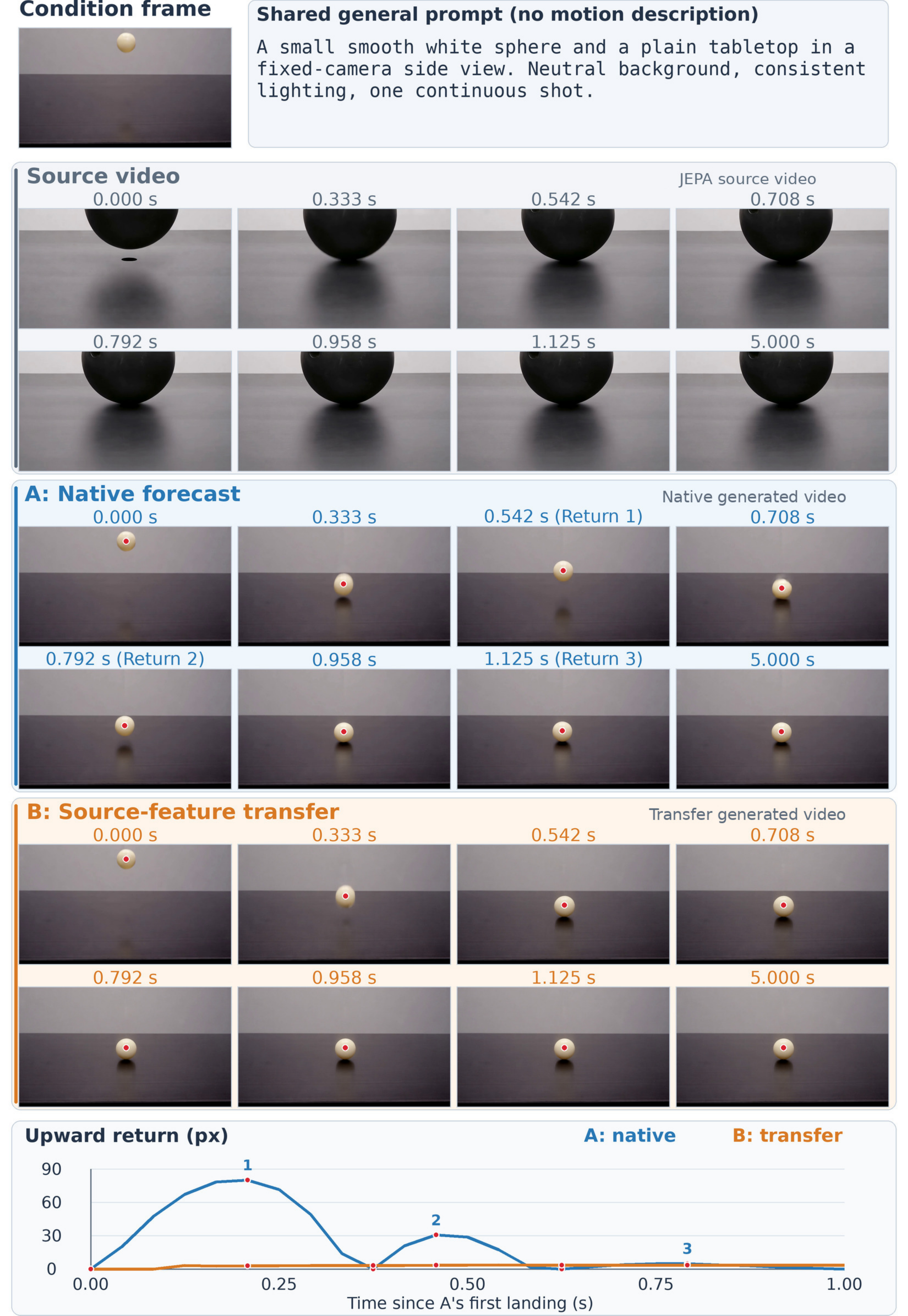}
\end{figure}

%% file: tables/appendix/dynamics_transfer_motion.tex
\begin{table}[!htbp]
\centering
\caption{\textbf{Image-plane ball-center movement in the matched pair.}
Both outputs use Cosmos 3 Nano 16B with \methodfull{}, over all 121 frames.
Image \(y\) increases downward, and distances are image-plane pixels.
Definitions in Figure~\ref{fig:dynamics-transfer-results}.}
\label{tab:dynamics-transfer-motion}
\begingroup
\footnotesize
\setlength{\tabcolsep}{4pt}
\renewcommand{\arraystretch}{1.05}
\begin{tabular*}{\linewidth}{@{\extracolsep{\fill}}lrrrr@{}}
\toprule
Configuration & Valid frames & Vertical span & Upward return & Path length \\
\midrule
A: Native forecast & 121/121 & 289.7 & 80.1 & 526.9 \\
B: Transfer & 121/121 & 281.1 & 3.6 & 290.9 \\
\bottomrule
\end{tabular*}
\endgroup
\end{table}

%% file: figures/appendix/dynamics_transfer_balloon.tex
\begin{figure}[!htbp]
\centering
\caption{\textbf{Descent transfer from a slow-descent \rev{source}.}
\rev{A and B share the prompt, initial noise, sampler, and amplification, and
differ only in the registry source.}
Red dots track ball centers and rows share timestamps.
The plot follows each center to first table contact, with image \(y\)
increasing downward and contact marked on each curve.}
\label{fig:dynamics-transfer-balloon}
\includegraphics[width=\linewidth]{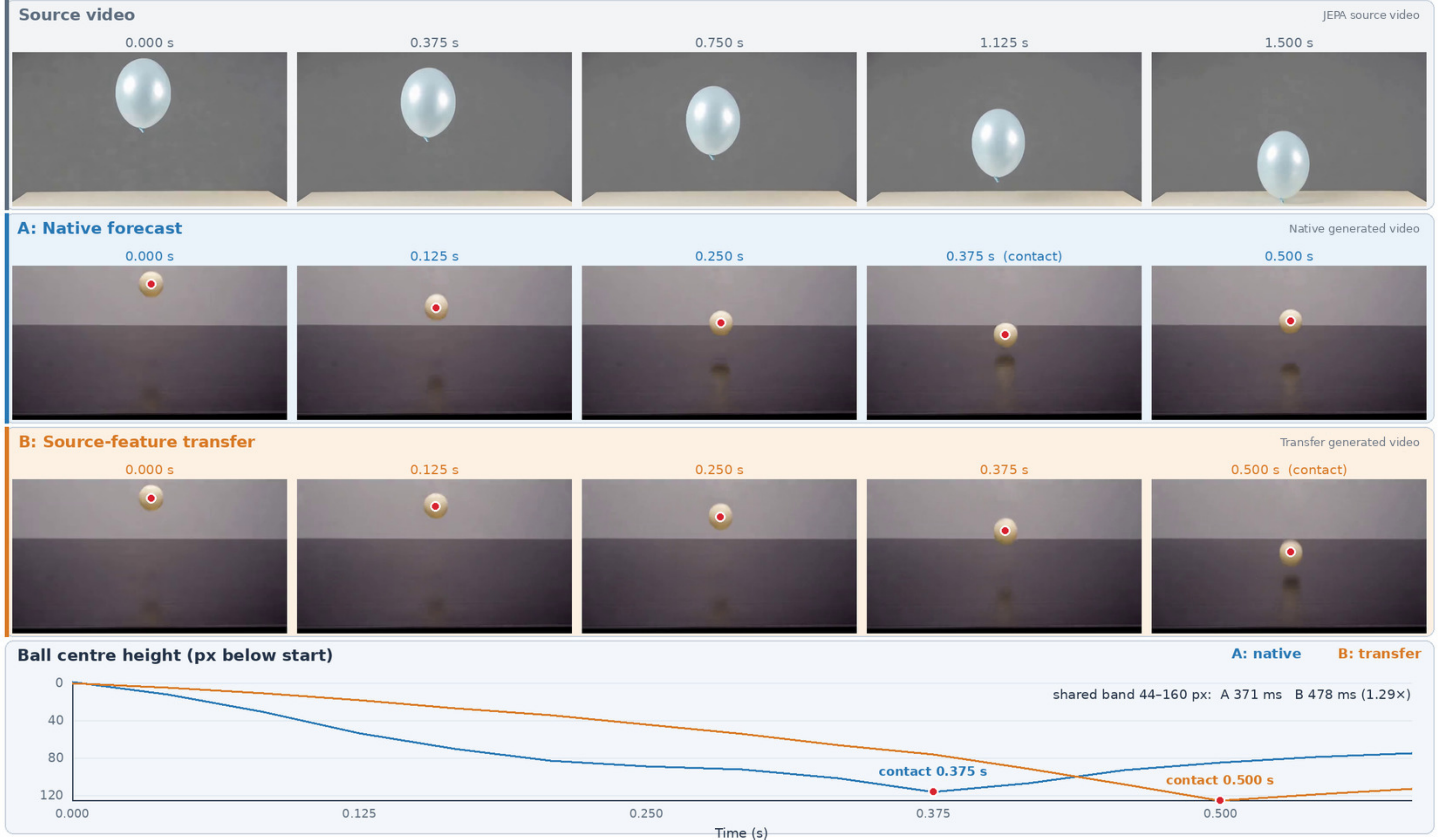}
\end{figure}

%% file: sections/appendix/f_mechanistic_diagnostics.tex
\clearpage
\section{Extended Mechanistic Diagnostics}
\label{app:mechanistic-diagnostics}

\subsection{Registry Feature Visualization}
\label{app:registry-visualization}

Registry activation maps characterize the spatial organization of each
registry slot. \rev{The main-text visualization in} Section~\ref{sec:analysis} aligns five future times from one
balloon-inflation trajectory and visualizes entity and action conditioning,
foreground-query attention to video tokens and to the aligned registry slots, and
slot-wise registry activation.

\input{figures/appendix/registry_layer_sweep}

\rev{For the depth sweep, we fix} denoising step 16 and \rev{measure} variance across hidden
feature channels at every registry token. Layers 7, 21, and 27 retain
broad scene-level activation, while the \rev{alignment-supervised} layer 14 exhibits
a response around the balloon and expanding contour. \rev{At this layer},
foreground enrichment is \(1.13{\times}\) and change-region
enrichment is \(1.05{\times}\)\rev{. This layer shows the largest measured enrichment
among these four layers}.
Enrichment is measured on the unnormalized maps.

%% file: figures/appendix/registry_layer_sweep.tex
\begin{figure}[!htbp]
\centering
\caption{\textbf{Registry feature organization across DiT layers.}
Columns align five generated frames with their registry slots. Heat rows show
hidden channel variance at denoising step 16 for layers 7, 14, 21, and 27.
Each layer uses an independent viridis log scale over all eight slots\rev{, so}
colors preserve within-layer slot magnitude.}
\label{fig:registry-layer-sweep}
\includegraphics[width=\linewidth]{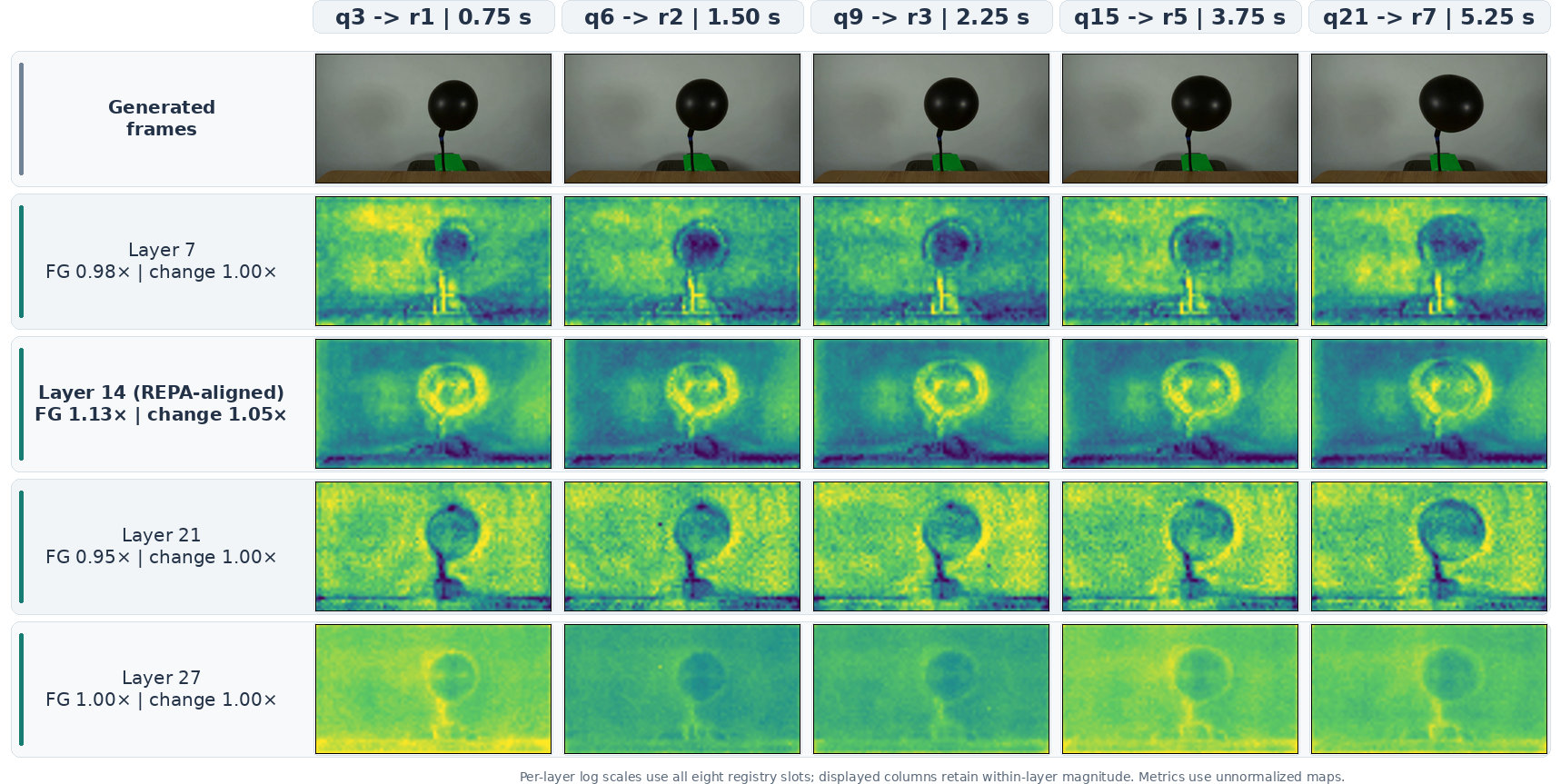}
\end{figure}

%% file: sections/appendix/e_extended_experiments.tex
\clearpage
\section{Qualitative Comparisons}
\label{app:extended-experiments}

\input{sections/appendix/e_qualitative_physicsiq}

\input{sections/appendix/h_additional_attention}

%% file: sections/appendix/e_qualitative_physicsiq.tex
\subsection{Additional Qualitative Comparisons}
\label{app:qualitative-physicsiq}

Figure~\ref{fig:qual-physicsiq-cosmos25} covers the five physics categories of
Figure~\ref{fig:qual-physicsiq} on Cosmos v2.5, and
Figures~\ref{fig:qual-physicsiq-more-a} and~\ref{fig:qual-physicsiq-more-b} provide additional
comparisons on the same backbone.
\rev{In each block, the top row shows the backbone and the bottom row shows
\method{} built on the same backbone. Both rows start from the same
conditioning input,} and frames are sampled at
eight equally spaced times of the generated sequence.

\rev{In these examples, \method{} completes several intended events that the backbone
leaves unrealized, while avoiding some duplicated or spurious objects.
On Cosmos v2.5 (Figure~\ref{fig:qual-physicsiq-cosmos25}), \method{} propagates
the domino topple after contact and extinguishes a lit match once it reaches
the water. In set A (Figure~\ref{fig:qual-physicsiq-more-a}), the backbone
duplicates the released ball and the magnet-sequence block, while \method{}
keeps each single. In set B (Figure~\ref{fig:qual-physicsiq-more-b}), \method{}
flattens the styrofoam cup under the descending kettlebell, while the backbone
leaves it standing.}

\input{figures/appendix/qualitative_physicsiq_more}

\subsection{WorldModelBench Qualitative Examples}
\label{app:worldmodelbench-qualitative}

Figures~\ref{fig:qual-worldmodelbench-robotics} and~\ref{fig:qual-worldmodelbench-other}
show eight WorldModelBench generations from Cosmos 3 Nano 16B
with \method{}, grouped into robotics and driving/industrial/natural scenes.
Each text prompt is followed by eight evenly spaced generated frames in a
large \(4\times2\) storyboard, read left to right and then top to bottom.
\rev{These examples illustrate the range of robotics, driving, industrial, and
natural-scene prompts covered by the evaluation.}

\input{figures/appendix/qualitative_worldmodelbench}

%% file: figures/appendix/qualitative_physicsiq_more.tex
\begin{figure}[!htbp]
\centering
\caption{\textbf{Qualitative comparisons on Cosmos v2.5.}
One scenario per physics category, with the layout of
Figure~\ref{fig:qual-physicsiq}; here the top row is the Cosmos v2.5 backbone.}
\label{fig:qual-physicsiq-cosmos25}
\includegraphics[width=\linewidth]{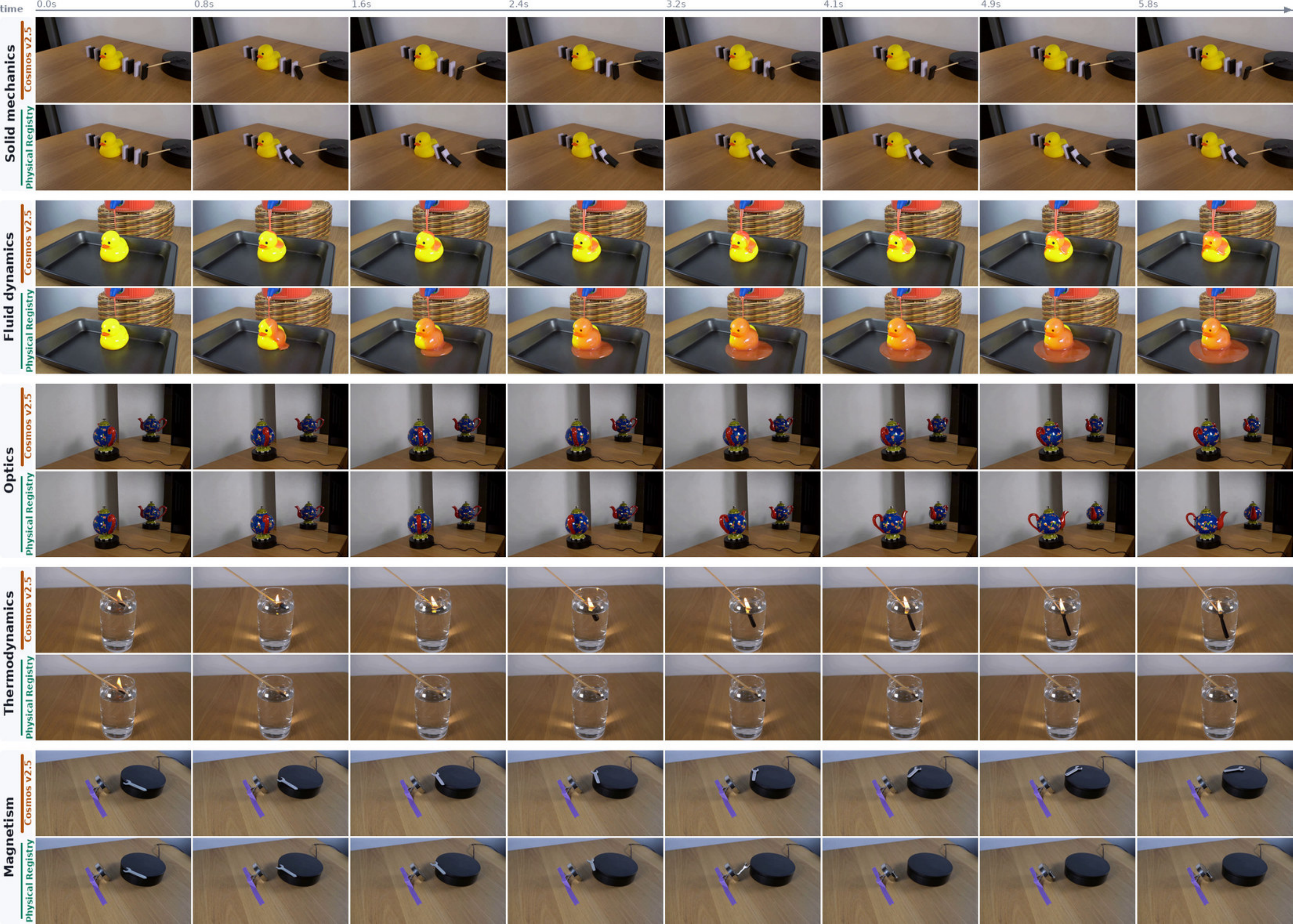}
\end{figure}

\begin{figure}[!htbp]
\centering
\caption{\textbf{Additional qualitative comparisons (set A).}
Layout as in Figure~\ref{fig:qual-physicsiq}.}
\label{fig:qual-physicsiq-more-a}
\includegraphics[width=\linewidth]{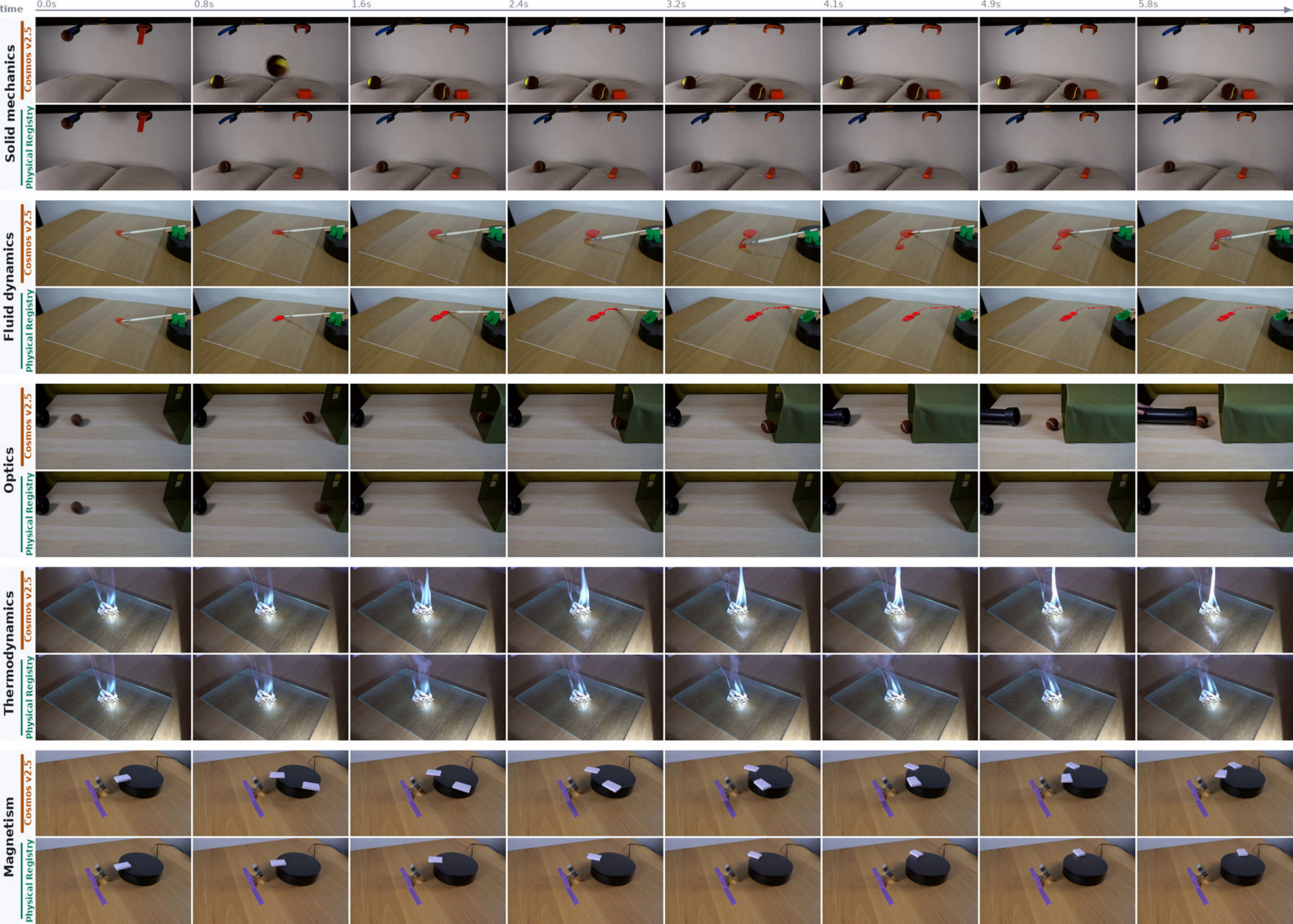}
\end{figure}

\begin{figure}[!htbp]
\centering
\caption{\textbf{Additional qualitative comparisons (set B).}
Layout as in Figure~\ref{fig:qual-physicsiq}.}
\label{fig:qual-physicsiq-more-b}
\includegraphics[width=\linewidth]{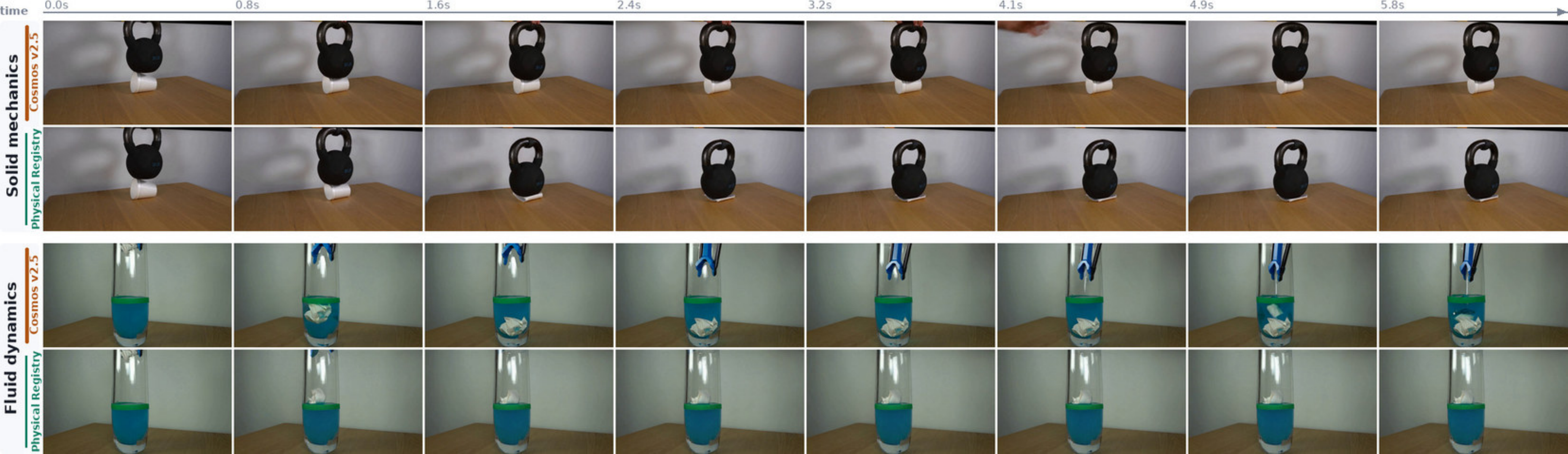}
\end{figure}

%% file: figures/appendix/qualitative_worldmodelbench.tex
\begin{figure}[!htbp]
\centering
\caption{\textbf{WorldModelBench robotics examples (Cosmos 3 Nano 16B with \methodfull{}).}
Each prompt is followed by eight frames, read left to right and top to bottom;
times are in seconds.}
\label{fig:qual-worldmodelbench-robotics}
\includegraphics[width=\linewidth]{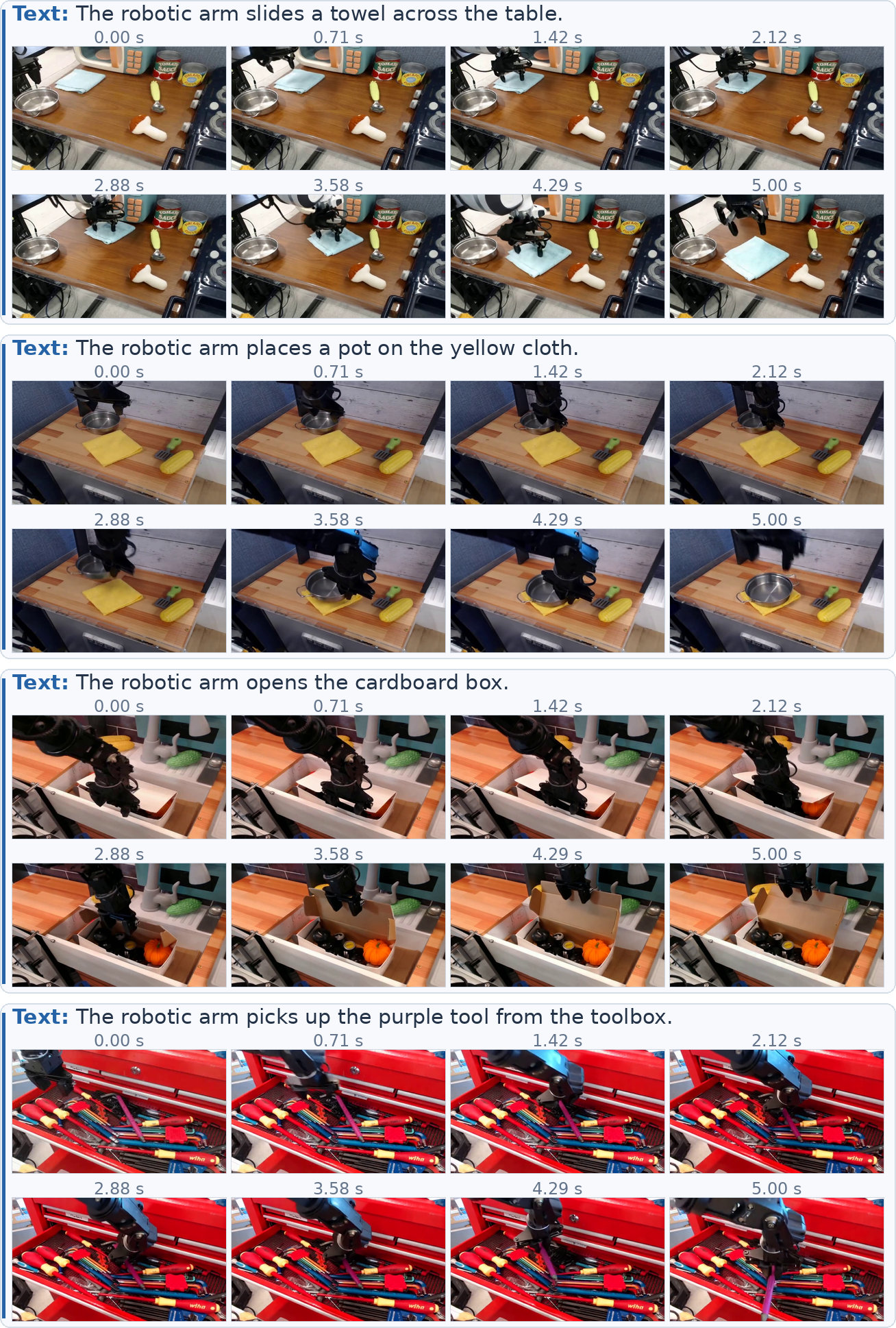}
\end{figure}

\begin{figure}[!htbp]
\centering
\caption{\textbf{WorldModelBench driving, industry, and nature examples
(Cosmos 3 Nano 16B with \methodfull{}).} Layout as in
Figure~\ref{fig:qual-worldmodelbench-robotics}.}
\label{fig:qual-worldmodelbench-other}
\includegraphics[width=\linewidth]{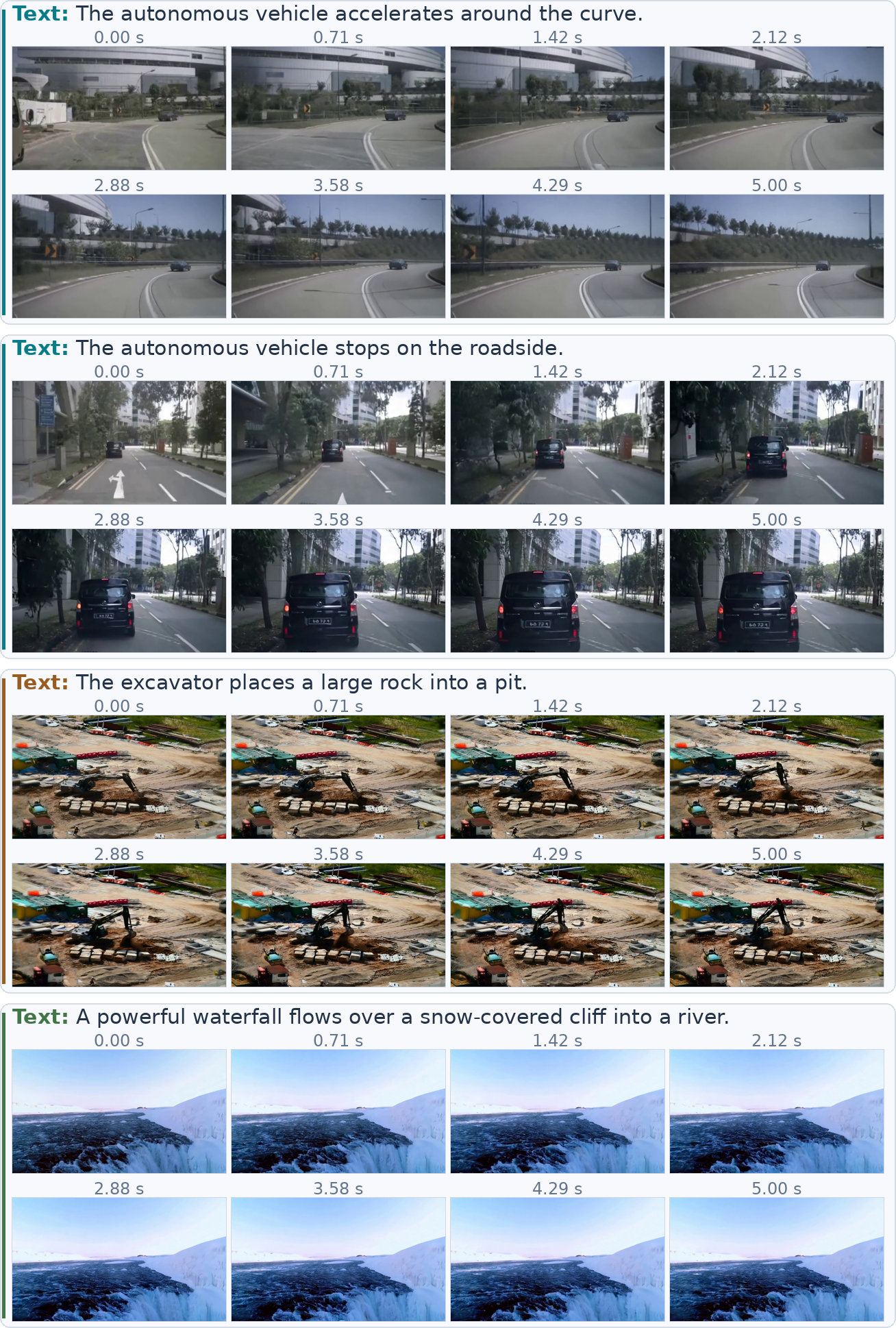}
\end{figure}

%% file: sections/appendix/h_additional_attention.tex
\subsection{Additional Attention and Registry Visualizations}
\label{app:additional-attention}

We extend Figure~\ref{fig:registry-attention} with six Physics-IQ examples spanning
solid mechanics, fluid dynamics, optics, and magnetism
\citep{motamed2026physicsiq}. All panels show the same six rows: generated
frames, entity and action cross-attention, self-attention to video and
registry slots, and registry activation. Captions identify entity words and
event verbs. Appendix~\ref{app:implementation} gives the visualization protocol.
\rev{Across these examples, registry attention and activation tend to concentrate
more locally around interacting or changing regions, whereas text cross-attention
is often broader.}

\input{figures/appendix/attention_occlusion}
\input{figures/appendix/attention_filling}

\clearpage
\input{figures/appendix/attention_shadow}
\input{figures/appendix/attention_magnet}

\clearpage
\input{figures/appendix/attention_ball_train}
\input{figures/appendix/attention_duck}

%% file: figures/appendix/attention_occlusion.tex
\begin{figure}[H]
\centering
\caption{\textbf{Solid mechanics: object occlusion.}
Entity text: \emph{ball}. Action text: \emph{lowered}.}
\label{fig:attention-occlusion}
\includegraphics[width=\linewidth]{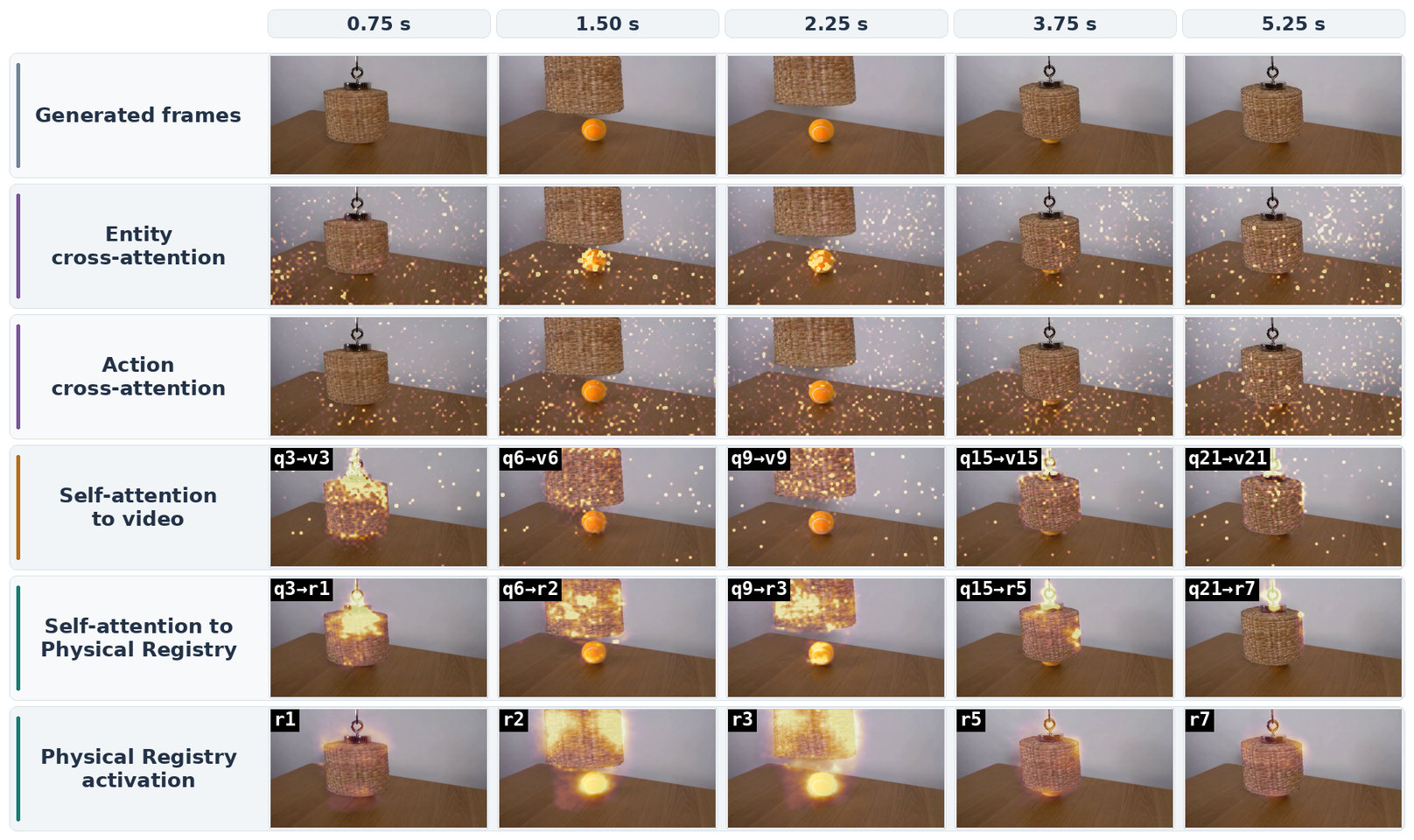}
\end{figure}

%% file: figures/appendix/attention_filling.tex
\begin{figure}[H]
\centering
\caption{\textbf{Fluid dynamics: filling a glass.}
Entity text: \emph{glass, red}. Action text: \emph{filled}.}
\label{fig:attention-filling}
\includegraphics[width=\linewidth]{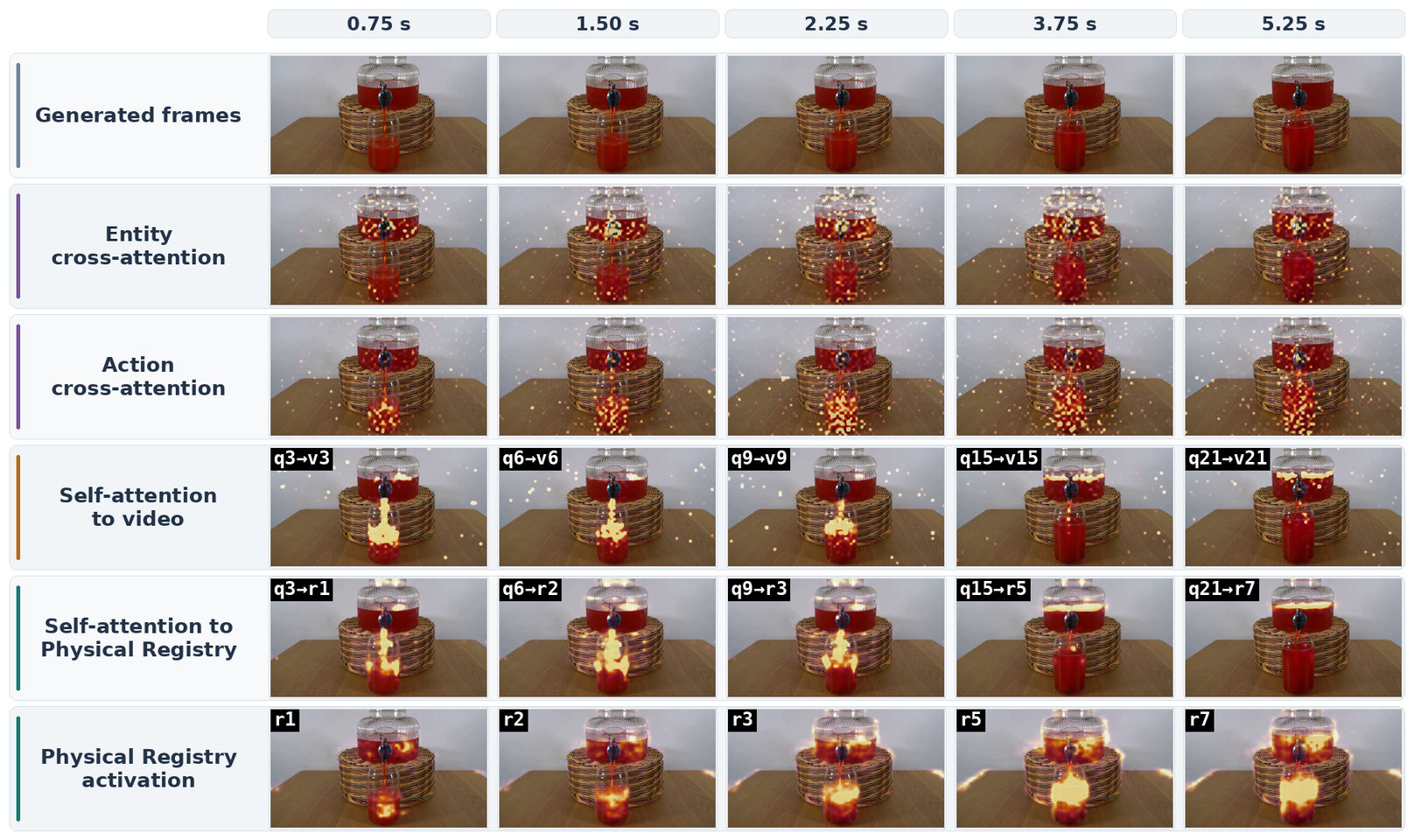}
\end{figure}

%% file: figures/appendix/attention_shadow.tex
\begin{figure}[H]
\centering
\caption{\textbf{Optics: a mug and its cast shadow.}
Entity text: \emph{mug}. Action text: \emph{placed, rotating, rotates, casting}.}
\label{fig:attention-shadow}
\includegraphics[width=\linewidth]{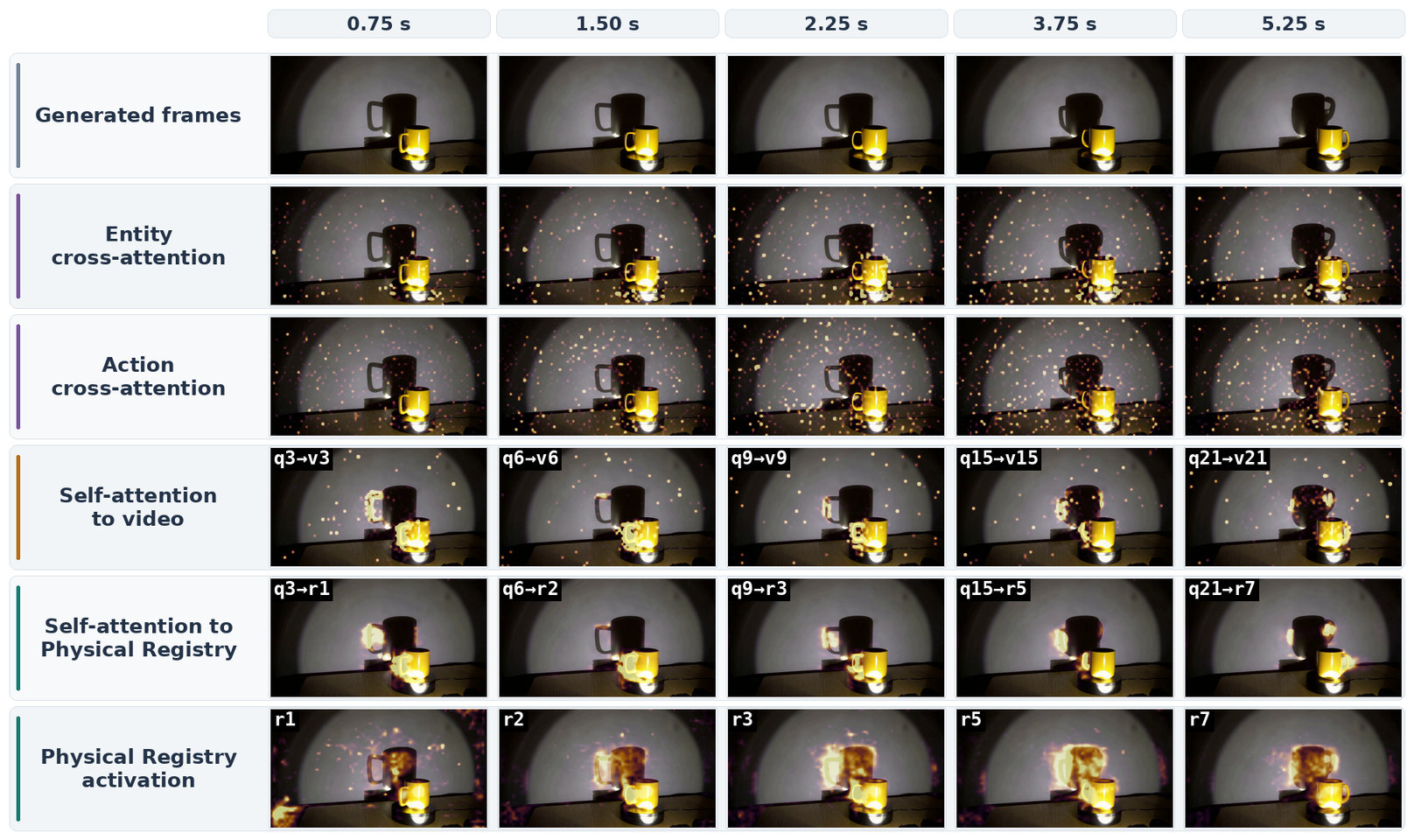}
\end{figure}

%% file: figures/appendix/attention_magnet.tex
\begin{figure}[H]
\centering
\caption{\textbf{Magnetism: a magnet--domino interaction.}
Entity text: \emph{domino, magnet}. Action text: \emph{placed, rotating}.}
\label{fig:attention-magnet}
\includegraphics[width=\linewidth]{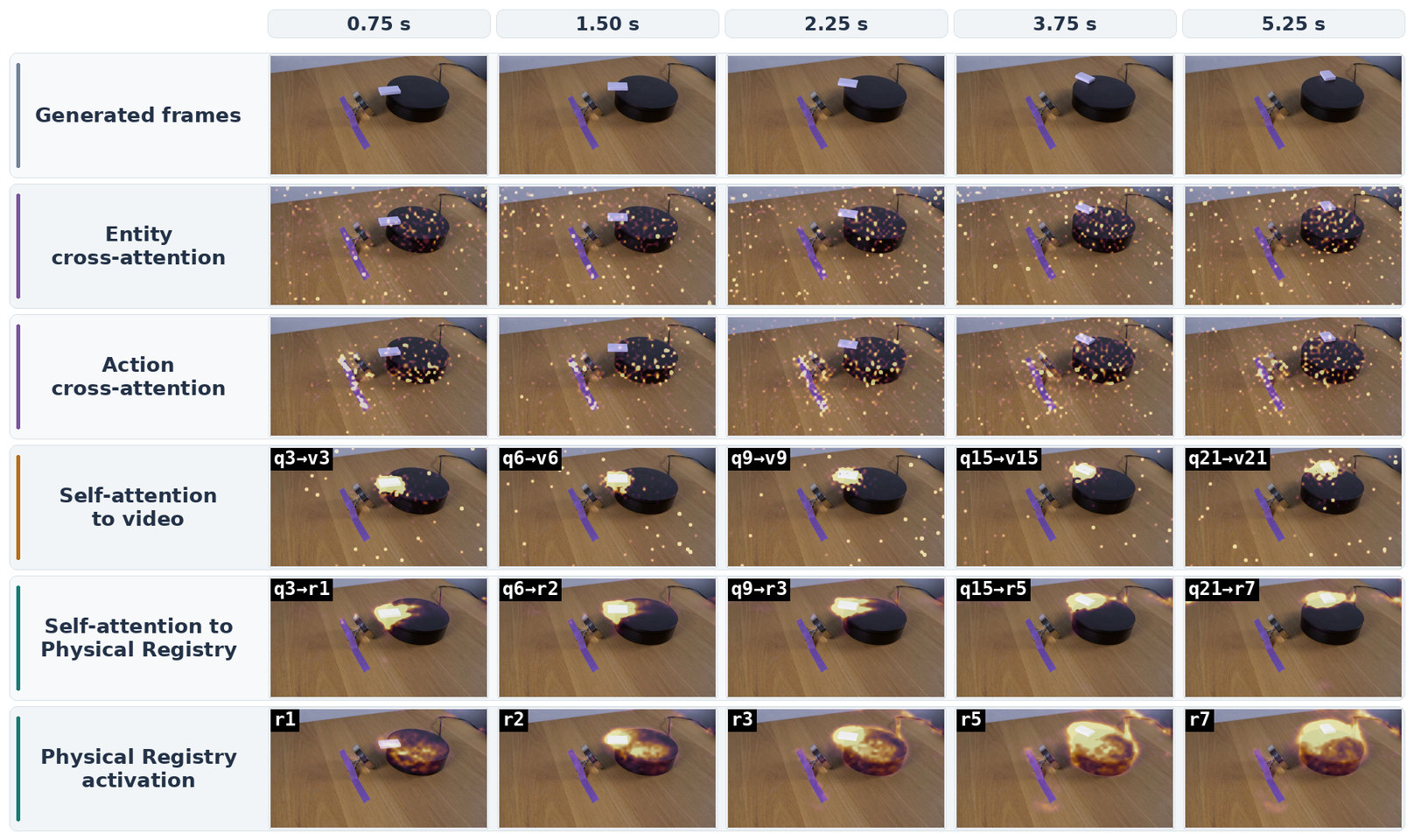}
\end{figure}

%% file: figures/appendix/attention_ball_train.tex
\begin{figure}[H]
\centering
\caption{\textbf{Solid mechanics: rolling-ball interaction.}
Entity text: \emph{ball, balls}. Action text: \emph{rolls}.}
\label{fig:attention-ball-train}
\includegraphics[width=\linewidth]{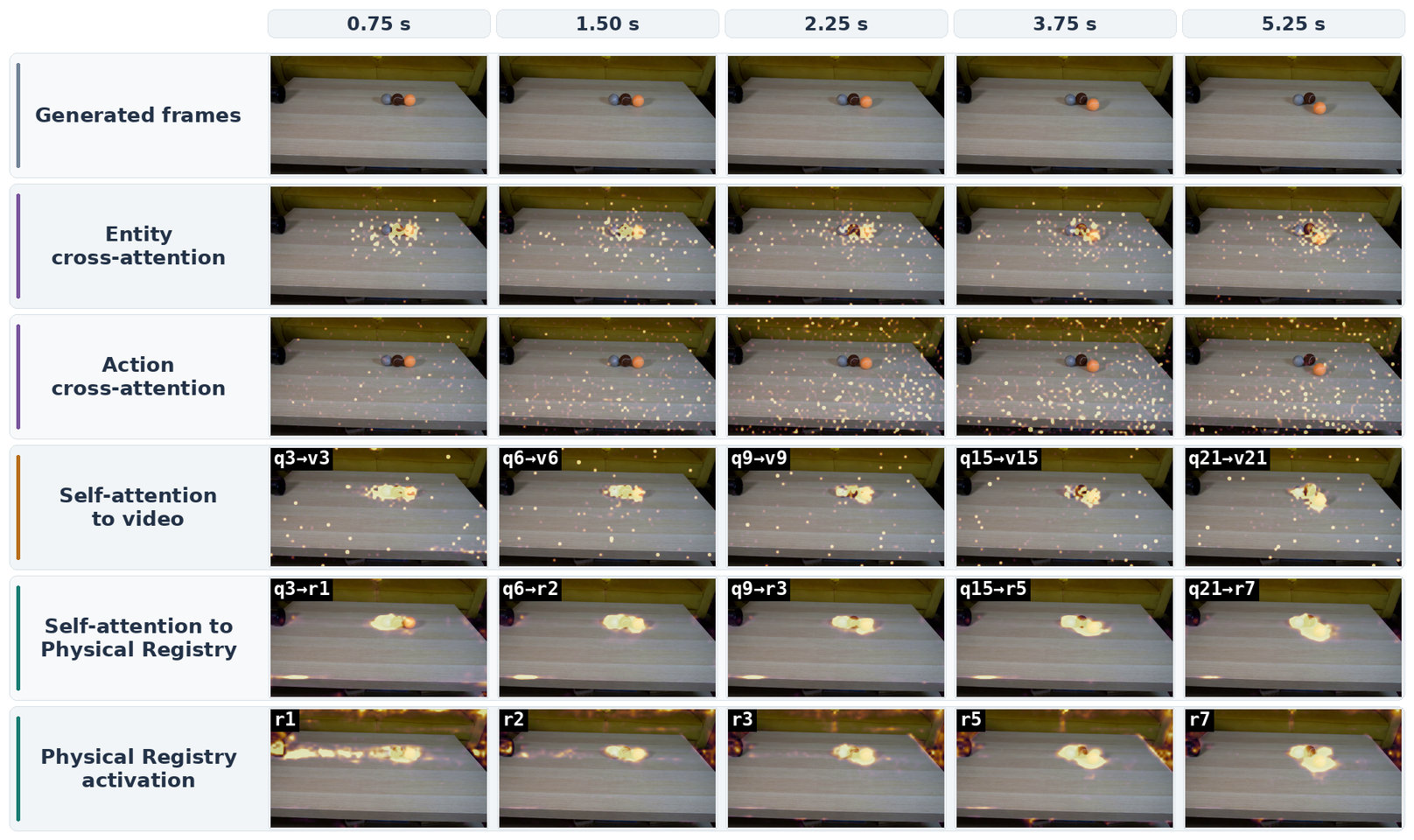}
\end{figure}

%% file: figures/appendix/attention_duck.tex
\begin{figure}[H]
\centering
\caption{\textbf{Fluid dynamics: pouring liquid over a duck.}
Entity text: \emph{liquid, duck}. Action text: \emph{pours}.}
\label{fig:attention-duck}
\includegraphics[width=\linewidth]{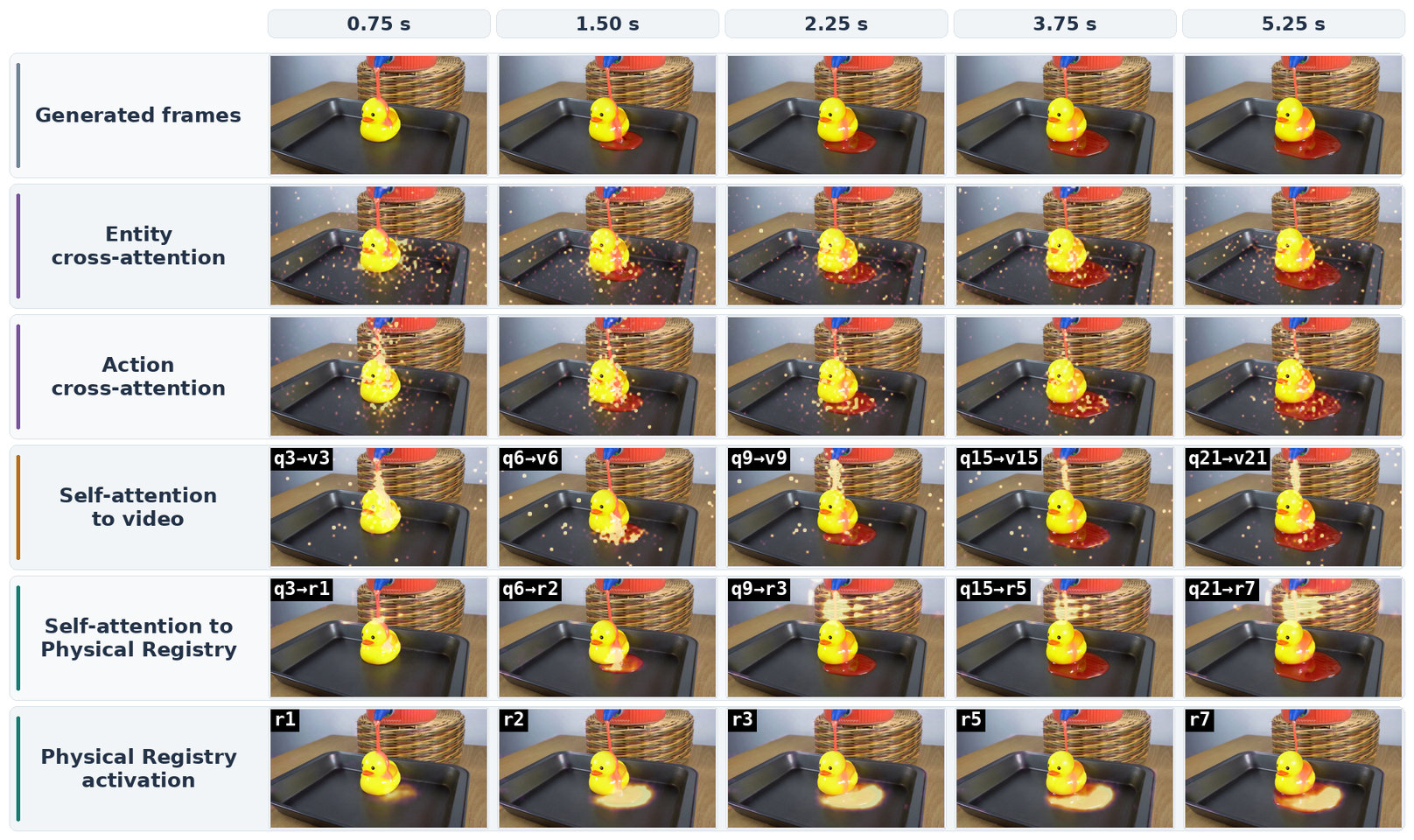}
\end{figure}